\documentclass[11pt]{article}
\usepackage[preprint]{acl}
\usepackage{times}

\usepackage[utf8]{inputenc} 
\usepackage[T1]{fontenc}    
\usepackage{hyperref}       
\usepackage{url}            
\usepackage{booktabs}       
\usepackage{amsfonts}       
\usepackage{nicefrac}       
\usepackage{microtype}      
\usepackage{xcolor}         
\usepackage{tabularx}
\usepackage{enumitem}
\usepackage{float}
\usepackage{placeins}
\usepackage{needspace}
\usepackage{listings}
\usepackage{tcolorbox}
\tcbuselibrary{breakable}
\usepackage{fvextra}
\usepackage{xspace}
\usepackage{amsmath}
\usepackage{pifont}
\usepackage{colortbl}
\usepackage{makecell}
\usepackage{caption}  
\usepackage{wrapfig}  
\usepackage{marvosym}
\usepackage{tikz}
\usepackage{graphicx}
\usepackage{fontawesome5}

\definecolor{BenchHeaderBg}{RGB}{40, 100, 175}
\definecolor{BenchAltRow}{RGB}{230, 240, 252}
\definecolor{BenchOursRow}{RGB}{195, 240, 205}
\definecolor{BenchModelBlue}{RGB}{52, 93, 126}
\definecolor{BenchModelRed}{RGB}{139, 69, 79}

\newcommand{\cmark}{\textcolor{green!60!black}{\ding{51}}}
\newcommand{\xmark}{\textcolor{red!70!black}{\ding{55}}}
\newcommand{\pmark}{\textcolor{orange!80!black}{\textbf{\textasciitilde}}}

\newcommand{\ourbench}[0]{\textsc{HEART-Bench}\xspace}

\newcommand{\hficon}{%
  \raisebox{-0.2\height}{\includegraphics[height=1em]{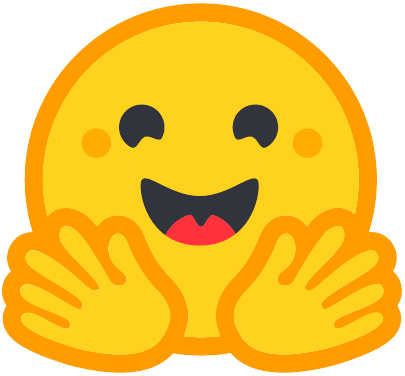}}%
}
\title{HEART-Bench: Do LLMs Make Persona-Consistent Decisions?}

\author{%
  \textbf{Weihan Peng\textsuperscript{1,*}},
  \textbf{Chenxu Zhang\textsuperscript{2,*}},
  \textbf{Qianao Wang\textsuperscript{3,4}},
  \textbf{Yuling Shi\textsuperscript{1}} \\
  \textbf{Heng Lian\textsuperscript{1}},
  \textbf{Qihong Mao\textsuperscript{3}},
  \textbf{Jiahao Pang\textsuperscript{5}},
  \textbf{Chunliang Feng\textsuperscript{5}} \\
  \textbf{Bowen Li\textsuperscript{3}},
  \textbf{Xiaodong Gu\textsuperscript{1,\Letter}} \\
  \\
  \textsuperscript{1}Shanghai Jiao Tong University,
  \textsuperscript{2}Imperial College London \\
  \textsuperscript{3}Quwan Group,
  \textsuperscript{4}University of Washington,
  \textsuperscript{5}South China Normal University \\
  \faGithub\ Code: \href{https://github.com/peng-weihan/HEART-BENCH}{\nolinkurl{peng-weihan/HEART-BENCH}} \\
  \hficon\ Dataset: \href{https://huggingface.co/datasets/HEART-BENCH/HEART-BENCH}{\nolinkurl{HEART-BENCH/HEART-BENCH}}
}

\begin{document}

\maketitle

\begingroup
\renewcommand{\thefootnote}{*}
\footnotetext[0]{Equal contribution.}
\renewcommand{\thefootnote}{\Letter}
\footnotetext[0]{Corresponding author.}
\endgroup

\begin{abstract}
LLM agents in interactive entertainment and social applications, including companions, role-play characters, and persona clones, need decisions grounded in personality and past experiences. We introduce \ourbench to evaluate memory-grounded persona-consistent response selection. The benchmark combines 11 Big Five--anchored synthetic characters, each with 1,000 autobiographical memories, and 64 DIAMONDS-guided decision scenarios into 673 multiple-choice questions. Domain experts independently compare candidate responses, with a senior expert resolving disagreements and revising responses when needed; other characters' adjudicated responses to the same scenario serve as distractors. In the main evaluation, models receive character IDs and occupations, scenarios, and options, with retrieved memories added in memory-augmented settings. Detailed personality-and-value profiles are withheld. We evaluate 16 LLMs under No-Retrieval, Naive-RAG, Mem0, and PersonaDB. Accuracy without memories stays near the 25\% chance level, and the best memory-augmented configuration reaches 64.9\%. On a matched 150-item subset, human annotators reach 80.8\%, compared with 64.0\% for the best model. Memory-content ablations show the importance of retrieving experiences that match both the target character and the decision context.
\end{abstract}

\begin{figure}[ht]
\centering
\includegraphics[width=\linewidth]{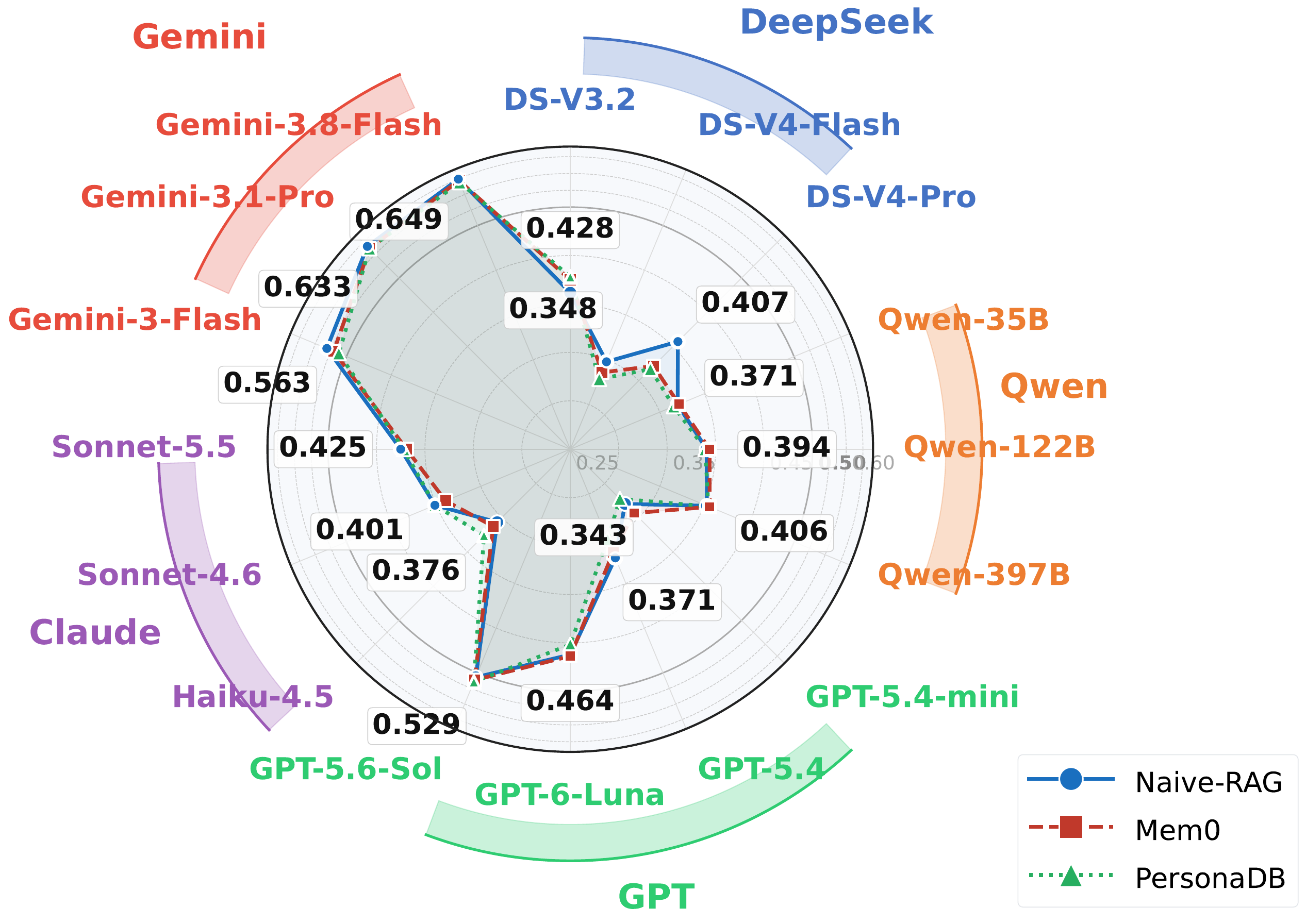}
\caption{Accuracy for all 16 models under the three memory-augmented settings. Radial scales are piecewise linear; tick labels indicate accuracy.}
\label{fig:radar_configs}
\end{figure}

\section{Introduction}
\label{sec:Introduction}

Large language model (LLM) agents are reshaping interactive entertainment and social applications, acting as AI companions~\cite{maples2024loneliness,phang2025investigating}, social and role-playing characters~\cite{chen2024persona,tseng2024two}, persona clones of specific people~\cite{wei2025ai,lee2026creating}, and non-player characters in games~\cite{park2023generative,gallotta2024large}. For such characters to feel alive, users must be able to recognize a stable personality and values in what they choose to do; consistency of personality expression is a long-standing requirement for believable agents~\cite{loyall1997believable,park2023generative}. Because personality and values are often conveyed only implicitly through what the character has experienced, acting consistently requires inferring them from these experiences and applying them to new situations. LifeChoice evaluates character decisions from narrative histories~\cite{xu2025lifechoice}, and PersonaMem-v2 studies personalized responses from implicit cues in conversational histories~\cite{jiang2025personamemv2}. We ask a complementary question: \emph{when different characters face the same external situation, can an LLM use their autobiographical memories to distinguish their decisions while explicit personality profiles remain hidden?}
\begin{table*}[t]
\centering
\footnotesize
\resizebox{\textwidth}{!}{
\begin{tabular}{lccccccc}
\toprule
\textbf{Benchmark} &
\textbf{\makecell{Big Five\\grounding}} &
\textbf{\makecell{Age-defined\\life-stage\\coverage}} &
\textbf{\makecell{Episodic\\memory}} &
\textbf{\makecell{Scenario\\evaluation}} &
\textbf{\makecell{Expert\\annotation}} &
\textbf{\makecell{In-character\\decision}} &
\textbf{\makecell{Profile-hidden\\evaluation}} \\
\midrule
\rowcolor{BenchAltRow}
CharacterEval~\cite{tu2024charactereval} & \xmark & \xmark & \xmark & \cmark & \pmark & \xmark & \xmark \\
PersonaHub~\cite{ge2024scaling}       & \xmark & \xmark & \xmark & \cmark & \xmark & \xmark & \textemdash \\
\rowcolor{BenchAltRow}
LoCoMo~\cite{maharana2024evaluating}     & \xmark & \xmark & \cmark & \cmark & \xmark & \xmark & \cmark \\
CloneMem~\cite{hu2026clonemem}           & \cmark & \pmark & \cmark & \xmark & \xmark & \xmark & \cmark \\
\rowcolor{BenchAltRow}
PersonaLens~\cite{zhao2025personalens}   & \xmark & \xmark & \pmark & \cmark & \pmark & \xmark & \cmark \\
PersonaMem~\cite{jiang2025personamem}    & \xmark & \xmark & \pmark & \cmark & \xmark & \xmark & \cmark \\
\rowcolor{BenchAltRow}
LifeChoice~\cite{xu2025lifechoice}       & \xmark & \xmark & \pmark & \cmark & \pmark & \cmark & \pmark \\
PersonaMem-v2~\cite{jiang2025personamemv2} & \xmark & \xmark & \pmark & \cmark & \xmark & \xmark & \cmark \\
\midrule
\rowcolor{BenchOursRow}
\textbf{\ourbench}       & \cmark & \cmark & \cmark & \cmark & \cmark & \cmark & \cmark \\
\bottomrule
\end{tabular}
}
\caption{Comparison of \ourbench with related benchmarks across seven dimensions (see Appendix~\ref{app:benchmark_dims} for dimension definitions).
\cmark~yes;~~\xmark~no;~~\pmark~partial;~~\textemdash~not applicable.}
\label{tab:benchmark_comparison}
\end{table*}

We introduce \ourbench for controlled comparisons of character-specific decisions under shared scenarios. Each of its 11 synthetic characters has basic information (a character ID and an occupation), a Big Five--anchored personality-and-value profile, and 1,000 synthetic autobiographical memories~\cite{singer1995seeing} spanning eight age-defined life stages. All characters face the same 64 decision scenarios designed with the DIAMONDS framework~\cite{rauthmann2014diamonds}, and three domain experts, with a senior expert resolving disagreements, adjudicate each reference response against the complete character specification and history. Each of the resulting 673 multiple-choice questions (MCQs) keys the target character's response and uses other characters' responses to the same scenario as distractors, so that all options are situationally plausible and differ mainly in the personality and values they reflect. In the main evaluation, models receive basic information, the scenario, the options, and retrieved memories when available; the detailed profile is withheld. Table~\ref{tab:benchmark_comparison} compares \ourbench with representative persona and memory benchmarks.

We evaluate 16 LLMs from five model families under No-Retrieval and three memory-augmented settings: Naive-RAG~\cite{lewis2020retrieval}, Mem0~\cite{chhikara2025mem0}, and PersonaDB~\cite{sun2025persona}. With basic information alone, accuracy remains near the 25\% chance level; retrieved memories improve every model (Table~\ref{tab:main_results}; Appendix~\ref{app:statistical_analysis}), and the best configuration reaches 64.9\% (Figure~\ref{fig:radar_configs}). On a 150-item subset, human annotators given the same information reach 80.8\%, compared with 64.0\% for the best model, a 16.8-percentage-point human--model gap on the same items. On all 673 items, Gemini-3.8-Flash~\cite{google2026gemini38flash} reaches 77.7\% when the construction profile is supplied alongside the same memories. Memory-content ablations further show that memories help most when they belong to the target character and match the scenario, whereas another character's memories for the same scenario reduce accuracy below chance.

We summarize our contributions as follows:
\begin{itemize}[leftmargin=2em]
    \item We develop a controlled evaluation design that combines Big Five--anchored character specifications, autobiographical histories, and shared scenarios to compare character-specific decisions under a fixed external context.
    \item We introduce \ourbench, 673 MCQs that pair each character's expert-adjudicated response with other characters' responses to the same scenario, testing response selection from memory with the explicit profile hidden.
    \item We benchmark 16 LLMs with and without memory retrieval and compare their performance with human judgments under matched inputs.
\end{itemize}

\begin{figure*}[t]
    \centering
    \includegraphics[width=1\linewidth]{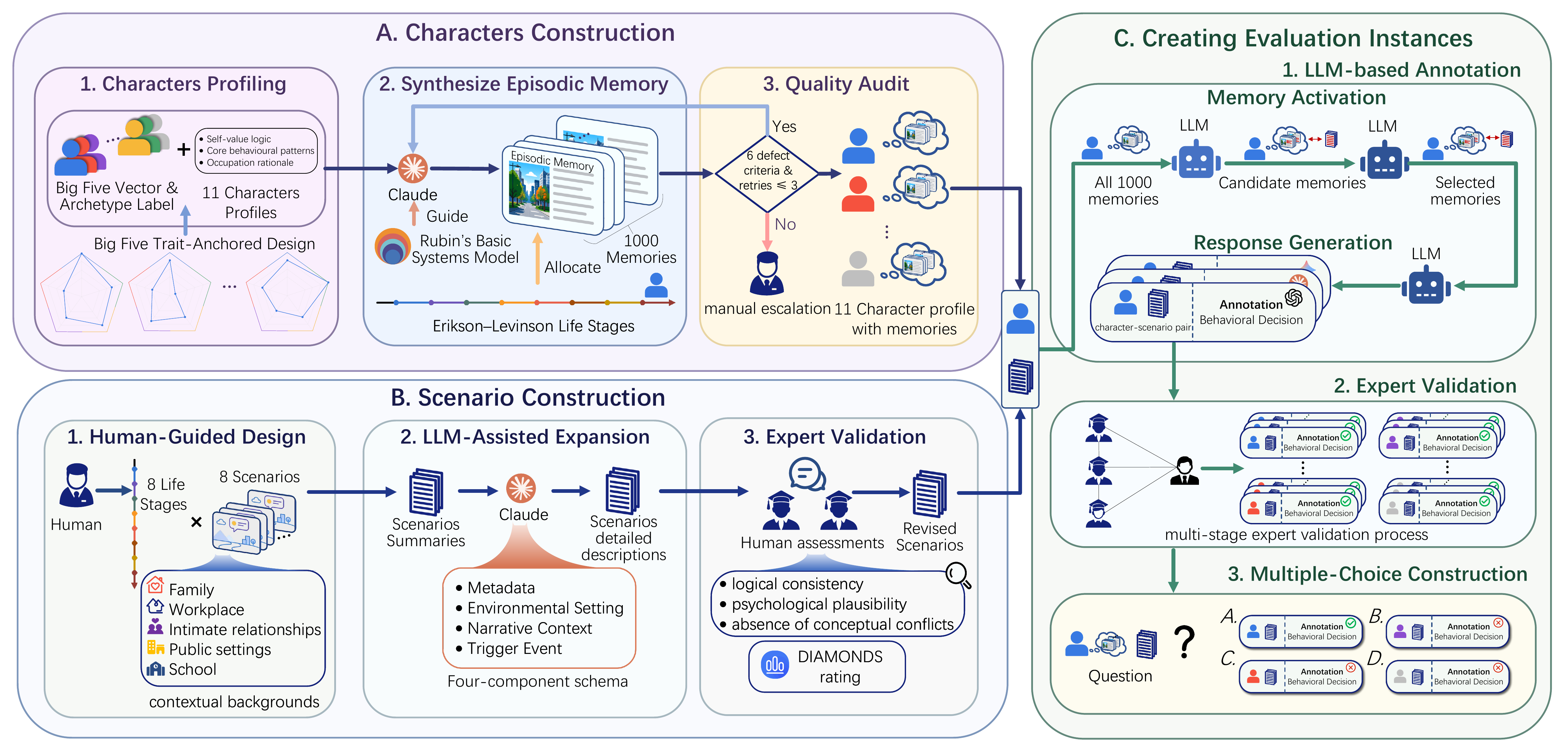}
    \caption{The pipeline has four stages: character construction, scenario construction, expert review and adjudication of reference responses, and conversion into MCQ format for model evaluation. The personality-and-value profile guides construction and adjudication. In the four main evaluation settings, it is withheld; models receive basic information, the scenario, the options, and retrieved memories when available.}
    \label{fig:pipeline}
\end{figure*}

\section{\ourbench}
\subsection{Overview}

Figure~\ref{fig:pipeline} illustrates the full pipeline of \ourbench. The benchmark comprises two components---an 11-character set informed by the Big Five~\cite{mccrae1992introduction}, with each character equipped with $1{,}000$ synthetic long-horizon memories, and a \emph{scenario suite} of 64 character-agnostic situations that span the eight DIAMONDS~\cite{rauthmann2014diamonds} dimensions---Duty, Intellect, Adversity, Mating, pOsitivity, Negativity, Deception, and Sociality---and are distributed across the eight age-defined life-stage bins. Expert review and MCQ assembly yield $673$ evaluation instances. Across the corpus, each character's $1{,}000$ memories total $\sim$1.4M tokens; the headline statistics are consolidated in Table~\ref{tab:dataset_overview} (Appendix~\ref{app:stats}).

\subsection{Character Construction}
\label{sec:character}

We define a synthetic character as \(c=(P,M)\), where $P$ is the complete construction specification and $M=\{m_i\}_{i=1}^{T}$ is a synthetic autobiographical history informed by research on narrative memory~\cite{singer1995seeing}. The specification includes the character identifier, a descriptive construction label (\texttt{name}), occupation, a predefined Big Five vector $\mathbf{p}=(O,C,E,A,N)\in[0,1]^5$, character description, self-value logic, and eight core behavioral patterns. We construct each character through three stages.

\textbf{Stage 1: Character Profiling.} The Big Five factor model~\citep{mccrae1992introduction} provides a shared specification of broad personality tendencies. For each OCEAN trait, our trait-anchored coverage design instantiates one high ($p=0.95$) and one low ($p=0.10$) anchor, together with an all-neutral ($p=0.50$) baseline, yielding 11 multi-trait profiles (Table~\ref{tab:characters}). Claude-Opus-4.6~\cite{anthropic2026opus46} elaborates each configuration into the textual fields of $P$ (Appendix~\ref{app:prompt_character}). For high/low profiles, occupations are chosen to fit the defining trait. Within $P$, the ID and occupation form \emph{basic information}; the Big Five vector, description, self-value logic, and core patterns form the \emph{personality-and-value profile}. The complete specification provides stable constraints for memory generation (Appendix~\ref{app:profile_grounding}). A four-expert team reviewed the resulting character specifications: three researchers in social cognitive science or neuroscience, each with $5{+}$ years of experience in personality assessment, and a senior professor with $15{+}$ years of experience and dual expertise in psychology and computer science. The researchers reviewed all 11 specifications and the professor finalized them before memory generation (Appendix~\ref{app:profile_review}).

\begin{figure}[t]
\centering
\includegraphics[width=0.8\linewidth]{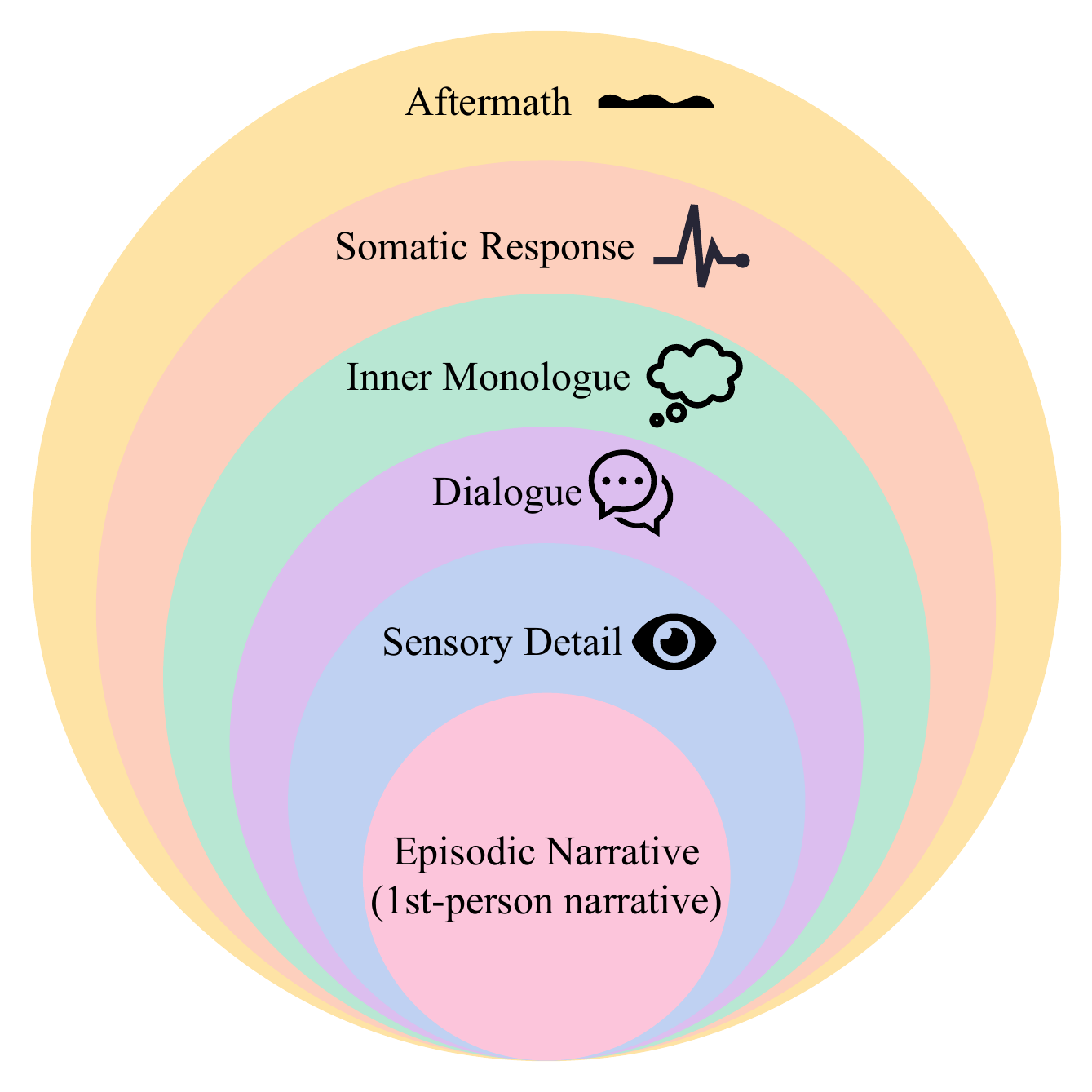}
\caption{Five operational dimensions, inspired by Rubin's Basic Systems Model~\cite{rubin2006basic}, used to structure synthetic episodic narratives.}
\label{fig:rubin}
\end{figure}

\textbf{Stage 2: Synthesizing Autobiographical Memories.} Conditioned on each accepted character specification, Claude-Sonnet-4.6~\cite{anthropic2026sonnet46} generates 1,000 episodic memories per character from ages 6--50. Memories are distributed across eight age-defined life-stage bins informed by Erikson~\cite{erikson1963childhood} and Levinson~\cite{levinson1978seasons,levinson1986conception}, with higher assigned density in adolescence and early-adult bins. The 45 one-year bins also emphasize the age-30 interval. As illustrated in Figure~\ref{fig:rubin}, each narrative integrates five operational dimensions inspired by Rubin's Basic Systems Model~\cite{rubin2006basic}---sensory detail, dialogue reconstruction, inner monologue, somatic response, and aftermath---and is diversified across workplace, interpersonal, health, creative, and daily-life domains. Alongside each narrative, the model generates a psychological interpretation (\texttt{psych\_conclusion}) and a behavioral-tendency summary (\texttt{behavior\_policy}) as auxiliary cues for memory retrieval and candidate-response generation during benchmark construction (Appendix~\ref{app:schema}). The allocation scheme, generation protocol, and prompt excerpt are provided in Appendices~\ref{app:stage_b}, \ref{app:stage_c}, and~\ref{app:prompt_memory}.

\textbf{Stage 3: Quality Audit.} Each memory is screened against six automated defect criteria covering perspective drift, age-inappropriate retrospection, metadata leakage, duplication, length, and over-psychologization. Of the 11{,}000 memories, 8{,}991 passed the initial automated audit, 1{,}935 passed after automated rewriting, and 74 were escalated to the same four-expert team introduced above. The professor revised 71 memories, all of which subsequently passed the LLM judge; the other three were unanimously accepted by the reviewers as false-positive LLM flags. Accepted memories are then deduplicated, chronologically ordered into a 1,000-entry timeline, and sanitized with gender-neutral labels. Full criteria and repair procedures appear in Appendix~\ref{app:stage_d}, with the final schema and statistics in Appendix~\ref{app:stats}.

\subsection{Scenario Construction}
\label{sec:scenario}

A scenario defines a character-agnostic decision context that serves as a shared external stimulus in our evaluation (see Appendix~\ref{app:ExampleofScenario} for a complete example). Each compatible character is placed in the same environmental and narrative context and required to respond to a decision-triggering event. Evaluated models must identify the response keyed to the target character. By holding the external situation fixed across characters, scenarios enable controlled comparison of whether LLMs can distinguish response differences encoded in the predefined profiles and synthetic histories.

Formally, a scenario is represented as $s=\langle m,\mathcal{C}_{\mathrm{env}},\mathcal{C}_{\mathrm{narrative}},e_{\mathrm{trigger}}\rangle$, where $m$ contains the scenario metadata, including its target life-stage bin and intensity; $\mathcal{C}_{\mathrm{env}}$ and $\mathcal{C}_{\mathrm{narrative}}$ specify the environmental and narrative contexts, respectively; and $e_{\mathrm{trigger}}$ denotes the event that prompts an immediate response.

We construct the scenario suite through a three-stage human-in-the-loop pipeline: expert blueprinting, LLM-assisted expansion, and expert review.

\textbf{Stage 1: Human-Guided Blueprinting.} 
To ensure structural diversity, experts manually draft the foundational summaries for all scenarios, using the DIAMONDS~\cite{rauthmann2014diamonds} framework to structure coverage of situational dimensions. We design exactly 8 scenarios per life-stage bin, yielding 64 scenarios across eight bins. These summaries cover family, school, workplace, relationship, and public-space contexts. Each summary establishes a core tension while maintaining the character-agnostic constraint.

\textbf{Stage 2: LLM-Assisted Expansion.}
We use Claude-Opus-4.6 to expand the expert-drafted summaries into structured scenario descriptions based on a four-component schema (metadata, environmental setting, narrative context, and trigger event), enforcing consistent fields across age metadata and context. The full schema and a complete worked example are given in Appendix~\ref{app:scenario_schema}.

\textbf{Stage 3: Expert Review.} Two doctoral-level psychology researchers review the generated scenarios for logical plausibility and assess their situational coverage using DIAMONDS~\cite{rauthmann2014diamonds}. The review protocol and full rating distributions are provided in Appendix~\ref{app:scenario_validation}.

\begin{figure*}[t]
\centering
\includegraphics[width=1\linewidth]{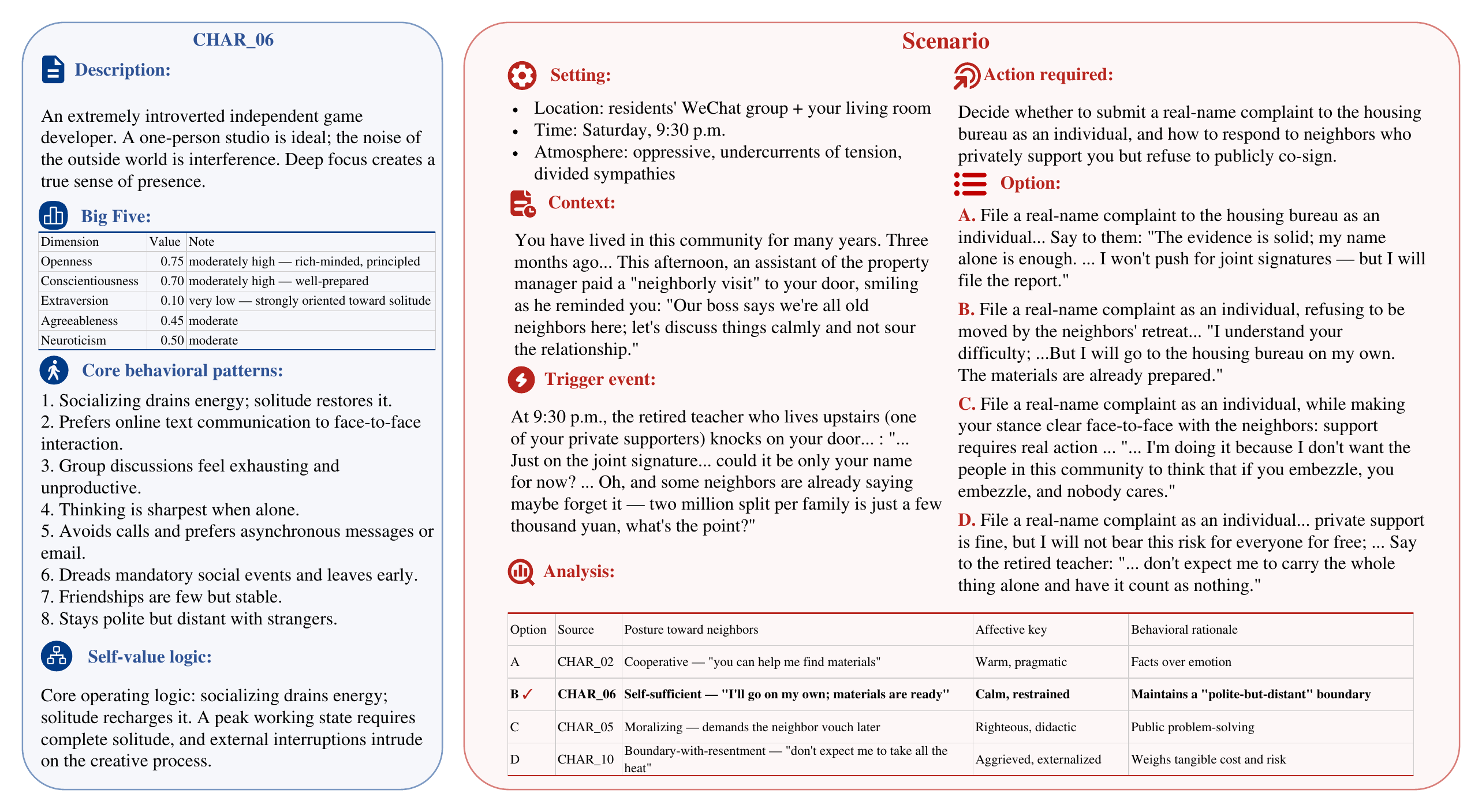}
\caption{A \ourbench instance requiring profile--response alignment beyond surface action matching; detailed analysis is in Appendix~\ref{app:case_study}.}
\label{fig:case_study}
\end{figure*}

\subsection{Creating Evaluation Instances}
\label{sec:CreatingEvaluationInstances}

Having built the character and scenario sets, we formulate each test instance as an MCQ whose keyed reference is the response judged by experts to be most consistent with the predefined target character, its synthetic history, and the scenario.

\Needspace{3\baselineskip}
\textbf{Stage 1: Automated Annotation with LLMs.}
For each (character, scenario) pair, the predefined Big Five scores, character description, and self-value logic serve as the primary character specification, while the scenario defines the decision context. Three parallel family-specific pipelines (Claude, GPT, and Gemini) each retrieve relevant episodic memories and use them alongside the profile and scenario to generate a candidate response. The memories provide experiential context, with their model-generated annotations serving as auxiliary cues for interpreting the situation and supporting behavioral judgments. Candidate order is randomized and model-family identities remain hidden throughout review. Model configurations, prompt excerpts, and an annotation example are provided in Appendix~\ref{app:annotation}.

\Needspace{3\baselineskip}
\textbf{Stage 2: Expert Review and Adjudication.}
The same four-expert team introduced in Section~\ref{sec:character} reviews and adjudicates the reference responses against the finalized character profiles. The three researchers serve as primary annotators, independently assessing pair suitability and comparing anonymized candidates for persona consistency, situational fit, and grounding in their associated memories, while the senior professor serves as arbitrator. Under a consensus-then-arbitration rule, a candidate is adopted directly only when all three annotators select it as \textit{acceptable as-is}; otherwise, the arbitrator selects, edits, or rewrites the response, or discards an unsuitable pair. This process removed $31$ pairs and retained $673$ expert-adjudicated reference responses. Experts changed approximately $15\%$ of the words in selected LLM candidates on average. Full guidelines and outcome statistics are provided in Appendix~\ref{app:annotation_protocol}.

\Needspace{3\baselineskip}
\textbf{Stage 3: Test Instance Assembly.}
For each retained annotation, the target character's adjudicated response becomes the keyed reference option, while \textit{Claude-Sonnet-4.6} selects three distractors from the available adjudicated responses associated with other characters in the same scenario. The selection excludes responses that are either too similar to the reference or obviously irrelevant, avoiding ambiguous or trivial questions. Option order is randomized before evaluation, yielding $673$ MCQs. Full assembly rules are provided in Appendix~\ref{app:mcq_assembly}.

A separate four-expert team not involved in construction audited a stratified random subset of 176 final MCQs: three experts independently selected the target character's most likely response without seeing the answer key, and the fourth adjudicated disagreements. All three experts selected the keyed option on 157 items (89.2\%) and two of three did so on the remaining 19. After adjudication, all 176 items retained their existing keys without edits or removals (Appendix~\ref{app:reference_label_audit}).

\begin{table*}[t]
\centering
\setlength{\tabcolsep}{6pt}
\small
\resizebox{\textwidth}{!}{%
\begin{tabular}{lc|ccc|c}
\toprule
\textcolor{BenchModelBlue}{\textbf{Model}} & \textcolor{BenchModelBlue}{\textbf{No-Retrieval LLM}} & \textcolor{BenchModelBlue}{\textbf{Naive-RAG}} & \textcolor{BenchModelBlue}{\textbf{Mem0}} & \textcolor{BenchModelBlue}{\textbf{PersonaDB}} & \textcolor{BenchModelBlue}{\textbf{Memory-Aug. Avg.}} \\
\midrule
\rowcolor{BenchModelBlue!8}
\multicolumn{6}{l}{\textcolor{BenchModelBlue}{\textit{Open-source models}}} \\
DeepSeek-V3.2 & 0.285 & 0.412 & 0.425 & 0.428 & 0.421 \\
DeepSeek-V4-Flash & 0.278 & 0.348 & 0.336 & 0.328 & 0.337 \\
DeepSeek-V4-Pro & 0.291 & 0.407 & 0.371 & 0.367 & 0.382 \\
Qwen3.5-35B-A3B & 0.254 & 0.370 & 0.371 & 0.366 & 0.369 \\
Qwen3.5-122B-A10B & 0.264 & 0.391 & 0.394 & 0.389 & 0.391 \\
Qwen3.5-397B-A17B & 0.275 & 0.403 & 0.406 & 0.404 & 0.404 \\
\cmidrule{1-6}
\rowcolor{BenchModelRed!8}
\multicolumn{6}{l}{\textcolor{BenchModelRed}{\textit{Closed-source models}}} \\
GPT-5.4-mini & 0.288 & 0.330 & 0.343 & 0.322 & 0.332 \\
GPT-5.4 & 0.278 & 0.371 & 0.366 & 0.351 & 0.363 \\
GPT-6-Luna & 0.269 & 0.462 & 0.464 & 0.452 & 0.459 \\
GPT-5.6-Sol & 0.285 & 0.514 & 0.523 & 0.529 & 0.522 \\
Claude-Haiku-4.5 & 0.245 & 0.357 & 0.363 & 0.376 & 0.365 \\
Claude-Sonnet-4.6 & 0.257 & 0.401 & 0.389 & 0.401 & 0.397 \\
Claude-Sonnet-5.5 & 0.288 & 0.425 & 0.419 & 0.422 & 0.422 \\
Gemini-3-Flash & 0.253 & 0.563 & 0.541 & 0.525 & 0.543 \\
Gemini-3.1-Pro & 0.264 & 0.633 & 0.624 & 0.624 & 0.627 \\
Gemini-3.8-Flash & 0.259 & \textbf{0.649} & \textbf{0.645} & \textbf{0.639} & \textbf{0.644} \\
\bottomrule
\end{tabular}
}
\caption{Accuracy on \ourbench, with open-source models listed first. Memory-Aug. Avg. averages the three retrieval-based settings; bold marks the best memory-augmented result in each column. Statistical details are provided in Appendix~\ref{app:statistical_analysis}.}
\label{tab:main_results}
\end{table*}

\subsection{Example of \ourbench}
Figure~\ref{fig:case_study} shows an item from the \emph{settling\_down} bin (ages 34--40) under the ``community injustice'' scenario. All four options file the same complaint, so surface actions do not identify the answer. CHAR\_06 combines very low Extraversion (0.10), moderately high Conscientiousness (0.70), and a ``polite-but-distant'' pattern. Option A adds coalition building, and Options C and D add moralizing and blame. Experts keyed Option B because its prior preparation, individual action, and concise wording best match this profile (Appendix~\ref{app:case_study}).

\section{Experiments}
\label{sec:experiments}

\subsection{Experimental Setup}

\paragraph{Model Selection.} We evaluate 16 LLMs from five model families. The six open-source models are \textit{DeepSeek-V3.2}~\cite{deepseek_v3_2_2025}, \textit{DeepSeek-V4-Flash}, \textit{DeepSeek-V4-Pro}~\cite{deepseek_v4_2026}, \textit{Qwen3.5-35B-A3B}, \textit{Qwen3.5-122B-A10B}, and \textit{Qwen3.5-397B-A17B}~\cite{qwen35_2026}. The ten closed-source models are \textit{GPT-5.4-mini}~\cite{openai2026gpt54mini}, \textit{GPT-5.4}~\cite{openai2026gpt54}, \textit{GPT-6-Luna}~\cite{openai2026gpt6luna}, \textit{GPT-5.6-Sol}~\cite{openai2026gpt56sol}, \textit{Claude-Haiku-4.5}~\cite{anthropic2025haiku45}, \textit{Claude-Sonnet-4.6}, \textit{Claude-Sonnet-5.5}~\cite{anthropic2026sonnet55}, \textit{Gemini-3-Flash}~\cite{google2025gemini3flash}, \textit{Gemini-3.1-Pro}~\cite{google2026gemini31pro}, and \textit{Gemini-3.8-Flash}. All models share the same decoding protocol (Appendix~\ref{app:decoding}).

\paragraph{Evaluation Inputs.} All four main settings provide the character ID and occupation, the scenario and trigger event, and the four response options; the personality-and-value profile and the descriptive construction label are withheld. No-Retrieval uses these inputs alone, whereas the memory-augmented settings add content retrieved from the target character's memories dated at or before the scenario age. Field-level isolation and the shared prompt are detailed in Appendices~\ref{app:interactive_settings} and~\ref{app:prompt_evaluation}.

\paragraph{Memory-Augmented Settings.} We compare Naive-RAG~\cite{lewis2020retrieval}, which retrieves original narratives, with Mem0~\cite{chhikara2025mem0} and PersonaDB~\cite{sun2025persona}, which reorganize the same narratives into fact-level entries and persona-oriented memory, respectively. Each backend queries the memories with the scenario description only; Naive-RAG and PersonaDB retrieve 30 entries, and Mem0 retrieves 150 facts. All three use \textit{Qwen3-Embedding-4B}~\cite{qwen3embedding}, and Mem0 and PersonaDB use \textit{GPT-5.4-mini} for memory management (Appendix~\ref{app:interactive_settings}).

\paragraph{Evaluation Metrics.} We report \textit{accuracy}, the proportion of the 673 MCQs on which a model selects the reference option, counting invalid responses as incorrect. For the main model comparisons, confidence intervals resample scenarios as clusters because multiple character-specific questions share the same scenario. Paired comparisons use scenario-level permutation tests with Holm correction (Appendix~\ref{app:statistical_analysis}).

\subsection{Main Results}

\paragraph{Model choice dominates within the benchmark.} Across all 16 models (Table~\ref{tab:main_results}), Naive-RAG accuracy ranges from 33.0\% to 64.9\%, a 31.9-percentage-point spread. The within-model range across retrieval methods averages 1.6 points, with a maximum of 4.0 points. Gemini-3.8-Flash achieves the highest Naive-RAG accuracy (64.9\%) and Memory-Augmented Avg. (64.4\%), followed by Gemini-3.1-Pro at 62.7\% Memory-Augmented Avg. Non-Gemini models achieve 33.2\%--52.2\% Memory-Augmented Avg. Model differences exceed those between memory backends. All 48 retrieval gains over No-Retrieval remain significant after Holm correction; statistical tests and model-agreement analyses appear in Appendices~\ref{app:statistical_analysis} and~\ref{app:model_agreement}.

\begin{figure}[t]
\centering
\includegraphics[width=\linewidth]{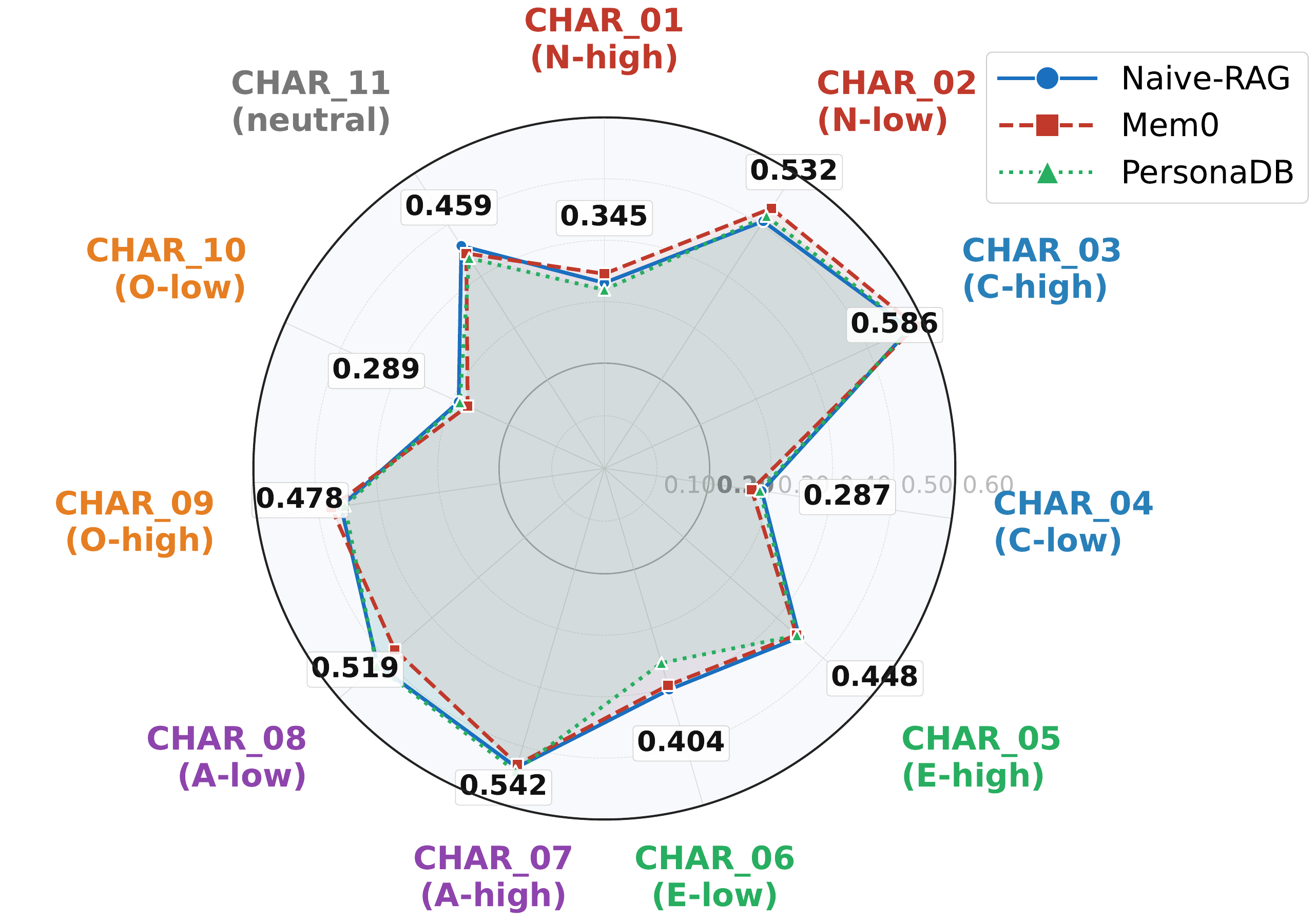}
\caption{Per-character accuracy averaged across all 16 LLMs by memory-augmented setting. Radial scales are piecewise linear; tick labels indicate accuracy.}
\label{fig:per_char_radar}
\end{figure}

\paragraph{Character-level difficulty.} Figure~\ref{fig:per_char_radar} summarizes all 16 models. Averaging across the three memory-augmented settings, C-High (CHAR\_03) reaches 57.9\%, whereas C-Low (CHAR\_04, 28.1\%) and O-Low (CHAR\_10, 28.3\%) are the two hardest characters. N-High (CHAR\_01, 33.2\%) is also among the more difficult characters (Appendix~\ref{app:character_accuracy}); accuracy by life-stage bin is reported in Appendix~\ref{app:life_stage}.

\paragraph{Humans under matched information.} Four psychology undergraduates not involved in benchmark construction answer a 150-item subset balanced across the three answer-key sources, with the same inputs as the main Naive-RAG setting. They reach 80.8\% mean accuracy (95\% CI 74.8--86.3) with no significant difference across sources, and at least three of the four annotators select the key on 82.0\% of items. On the same items, Gemini-3.8-Flash reaches 64.0\%, 16.8 percentage points below humans (95\% CI 7.3--26.2), and the 16 models average 43.0\%. With identical memory strings and option order, the gap suggests that models mainly struggle to use the retrieved experiences, although humans, unlike models, answered all items in sequence and could compare options across items. The protocol, agreement statistics, and per-source results are provided in Appendix~\ref{app:human_baseline}.

\section{Analysis}
\label{sec:analysis}
We analyze how profile availability, memory content, and candidate provenance relate to response accuracy.

\paragraph{How does profile disclosure affect accuracy?} We evaluate Gemini-3.8-Flash on all 673 items with the profile disclosed, either alone or together with the memories retrieved in the main Naive-RAG setting (Appendix~\ref{app:profile_memory_diagnostic}). With the same memories, disclosure raises accuracy by 12.78 points, from 64.93\% to 77.71\%, and the profile alone yields 72.96\%. The disclosed profile supplies the complete reviewed Big Five vector, description, self-value logic, and behavioral patterns. Adding memories to this profile provides a further 4.75-point gain. The comparisons show complementary contributions from explicit character specifications and autobiographical evidence: one supplies a concise account of the persona, and the other supplies experiences relevant to the decision. The remaining 22.3\% of responses disagree with the expert reference.

\begin{table}[t]
\centering
\setlength{\tabcolsep}{2pt}
\small
\resizebox{\linewidth}{!}{%
\begin{tabular}{cllr}
\toprule
\textbf{Cond.} & \textbf{Character source} & \textbf{Memory selection} & \textbf{Acc.} \\
\midrule
C0 & Target & Target scenario & \textbf{0.426} \\
C1 & Target & Other same-bin scenario & 0.371 \\
C2 & Other & Target scenario & 0.235 \\
C3 & Target & Random non-target & 0.367 \\
\bottomrule
\end{tabular}
}
\caption{Memory-content ablation on the 176-item subset. Accuracy is averaged over all 16 models (Appendix~\ref{app:memory_ablation}).}
\label{tab:memory_ablation_summary}
\end{table}

\paragraph{Does memory content matter?} On a fixed 176-item subset, we vary only the memories supplied to all 16 models under Naive-RAG with $k{=}30$ (Table~\ref{tab:memory_ablation_summary}). Target-character memories retrieved for the target scenario (C0) perform best for every model, and target-character memories from another same-bin scenario (C1) or random sampling (C3) outperform another character's memories for the target scenario (C2) for every model, with C2 falling below chance on average (23.5\%). Memories from the correct character thus help even when retrieved for another context, and matching them to the current situation adds a further benefit. Character ownership contributes more than scenario matching: C1 exceeds C2 by 13.66 points on average, whereas C0 exceeds C1 by 5.48 points. C1 and C3, in contrast, differ by only 0.46 points.

\paragraph{Does the Gemini lead persist across reference sources?} Gemini-3.8-Flash achieves the highest Memory-Augmented Avg. on every reference-source subset, reaching 68.3\% on Claude-origin items and 63.8\% on Gemini-origin items. Gemini candidates account for 497 of the 673 references and 72.9\% of distractors; 486 of the 497 Gemini-origin items include at least one Gemini-origin distractor. Under No-Retrieval, the three Gemini models average 25.8\% on Gemini-origin items, compared with 25.9\% for the other 13 models; Claude and GPT models also show no own-source advantage in this setting. With memories, the Gemini family's mean lead is 18.0 points on non-Gemini-origin items, with an additional 3.7-point lead on Gemini-origin items (95\% CI $-1.4$--$8.7$; Appendix~\ref{app:gt_source_results}). Model rankings are also positively correlated across source subsets, with pairwise Spearman correlations from 0.842 to 0.915. The leading models therefore perform well across references originating from different candidate generators, supporting comparisons beyond the dominant Gemini-origin subset.

\FloatBarrier
\section{Related Work}
\label{sec:related_work}

\paragraph{Persona-conditioned behavior.} Persona evaluations cover role-playing and trait expression~\cite{shao2023character,jiang2024personallm}. TimeChara evaluates character knowledge at specified narrative moments~\cite{ahn2024timechara}; InCharacter assesses personality fidelity through psychological interviews~\cite{wang2024incharacter}; and PersonaGym evaluates free-form responses in persona-conditioned environments~\cite{samuel2025personagym}. \citet{park2024generative1000} compare interview-grounded agents with real participants on surveys and behavioral tasks. MRBench studies persona-knowledge use in dialogue~\cite{wang2026memoryroleplaying}, CustomerSim examines persona-constrained shopping decisions~\cite{pruksachatkun2026customersim}, and AgentPersonaBench evaluates supplied profiles across survey, chat, web, and app tasks~\cite{huang2026agentpersonabench}.

\paragraph{History-grounded decisions and personalization.} LifeChoice uses descriptions and retrieved passages from narrative histories to predict characters' decisions in four-option questions; references follow the original narrative and distractors are generated~\cite{xu2025lifechoice}. PersonaMem-v2 evaluates personalized assistant responses from implicit, evolving preferences in multi-session histories, using MCQ and open-ended formats~\cite{jiang2025personamemv2}; PersonaMem-v3 combines psychology-grounded personas with cross-platform understanding, recommendations, and proactive tasks~\cite{jiang2026personamemv3}. Other benchmarks examine conversational memory (LoCoMo and LongMemEval)~\cite{maharana2024evaluating,wu2024longmemeval}, event-driven preferences (PERMA)~\cite{liu2026perma}, memory management and response alignment in real dialogues (AlpsBench)~\cite{xiao2026alpsbench}, user-specific retention (PerMemBench)~\cite{in2026personalizethenstore}, and implicit logical retrieval for personalized responses (IMLogic/RootMem)~\cite{ding2026rootmemories}. UserToolBench evaluates inferred preferences through clarification and tool-call trajectories~\cite{yin2026usertoolbench}.

\paragraph{Scope of \ourbench.} LifeChoice's references follow decisions in the original narrative, with generated distractors. \ourbench places controlled Big Five--anchored characters and autobiographical histories in shared DIAMONDS-guided scenarios. Experts adjudicate character-specific responses; other characters' same-scenario references become distractors. This fixes the external situation while varying the character and its evidence. With detailed profiles hidden in the main evaluation, the task measures recovery of the target character's reference response. The design supports comparisons of memory ownership, scenario relevance (Table~\ref{tab:memory_ablation_summary}), and profile disclosure.

\section{Conclusion}
\ourbench evaluates persona-consistent response selection from basic character information and autobiographical memories. It comprises 11 synthetic characters, 11{,}000 memories, 64 shared scenarios, and 673 expert-adjudicated MCQs. Across 16 LLMs, memory retrieval improves accuracy over the near-chance baseline, with the best configuration reaching 64.9\%. On the 150-item subset, humans achieve 80.8\% under matched inputs, compared with 64.0\% for the best model. Gemini-3.8-Flash reaches 77.7\% on the full benchmark when the construction profile is disclosed alongside the same memories. Memory-content ablations show that target-character memories aligned with the scenario are most effective. \ourbench provides a common testbed for improving how character agents connect past experiences to persona-consistent decisions.

\section*{Limitations}
\label{sec:limitations}
The single-turn MCQ evaluation covers 11 synthetic characters and 64 scenarios. Open-ended interaction and user-perceived believability require multi-turn studies and user assessments. Candidate provenance is observational, and candidate generators share model families with the evaluated models (Appendix~\ref{app:gt_source_results}). The expert-intensive construction limits scale. Validation by independent teams and references grounded in observed human behavior would extend the benchmark's validation and coverage.

\section*{Ethics Statement}
All characters, memories, and scenarios in \ourbench are synthetic and are not derived from real individuals. Memories are written in the first person without character names, and the benchmark contains no personally identifiable information. Some scenarios involve interpersonal conflict, deception, or moral dilemmas in order to elicit value-relevant decisions. The keyed references encode consistency with a given synthetic character rather than normative judgments of how people should act, and the benchmark is not intended for profiling or predicting the behavior of real individuals. The human evaluation was approved by our institutional ethics review board. Four psychology undergraduates were recruited through a public call, provided informed consent, and were paid USD 10 per hour, above the local minimum wage; they had no access to answer keys or construction materials (Appendix~\ref{app:human_baseline}).

\section*{LLM Usage}
ChatGPT, Codex, and Cursor assisted with editing and revising the manuscript. The authors verified all statements, numbers, and references and take full responsibility for the content of this paper.

\bibliography{ref}

@inproceedings{park2023generative,
  title={Generative agents: Interactive simulacra of human behavior},
  author={Park, Joon Sung and O'Brien, Joseph and Cai, Carrie Jun and Morris, Meredith Ringel and Liang, Percy and Bernstein, Michael S.},
  booktitle={Proceedings of the 36th Annual {ACM} Symposium on User Interface Software and Technology},
  pages={1--22},
  year={2023},
  publisher={Association for Computing Machinery},
  doi={10.1145/3586183.3606763},
  url={https://doi.org/10.1145/3586183.3606763}
}

@article{gallotta2024large,
  title={Large language models and games: A survey and roadmap},
  author={Gallotta, Roberto and Todd, Graham and Zammit, Marvin and Earle, Sam and Liapis, Antonios and Togelius, Julian and Yannakakis, Georgios N},
  journal={IEEE Transactions on Games},
  year={2024},
  pages={1--18},
  note={Early Access},
  publisher={IEEE},
  doi={10.1109/TG.2024.3461510},
  url={https://doi.org/10.1109/TG.2024.3461510}
}

@techreport{loyall1997believable,
  title={Believable agents: building interactive personalities},
  author={Loyall, A. Bryan},
  year={1997},
  month=may,
  number={CMU-CS-97-123},
  institution={School of Computer Science, Carnegie Mellon University},
  url={https://www.cs.cmu.edu/afs/cs.cmu.edu/project/oz/web/papers/CMU-CS-97-123.pdf}
}

@inproceedings{xu2025lifechoice,
  title = {Character is Destiny: Can Role-Playing Language Agents Make Persona-Driven Decisions?},
  author = {Xu, Rui and Wang, Xintao and Chen, Jiangjie and Yuan, Siyu and Yuan, Xinfeng and Liang, Jiaqing and Chen, Zulong and Dong, Xiaoqing and Xiao, Yanghua},
  booktitle = {Findings of the Association for Computational Linguistics: EMNLP 2025},
  year = {2025},
  pages = {15038--15059},
  publisher = {Association for Computational Linguistics},
  doi = {10.18653/v1/2025.findings-emnlp.813},
  url = {https://aclanthology.org/2025.findings-emnlp.813/}
}

@inproceedings{ahn2024timechara,
  title = {{TimeChara}: Evaluating Point-in-Time Character Hallucination of Role-Playing Large Language Models},
  author = {Ahn, Jaewoo and Lee, Taehyun and Lim, Junyoung and Kim, Jin-Hwa and Yun, Sangdoo and Lee, Hwaran and Kim, Gunhee},
  booktitle = {Findings of the Association for Computational Linguistics: ACL 2024},
  year = {2024},
  pages = {3291--3325},
  publisher = {Association for Computational Linguistics},
  doi = {10.18653/v1/2024.findings-acl.197},
  url = {https://aclanthology.org/2024.findings-acl.197/}
}

@inproceedings{wang2024incharacter,
  title = {{InCharacter}: Evaluating Personality Fidelity in Role-Playing Agents through Psychological Interviews},
  author = {Wang, Xintao and Xiao, Yunze and Huang, Jen-tse and Yuan, Siyu and Xu, Rui and Guo, Haoran and Tu, Quan and Fei, Yaying and Leng, Ziang and Wang, Wei and Chen, Jiangjie and Li, Cheng and Xiao, Yanghua},
  booktitle = {Proceedings of the 62nd Annual Meeting of the Association for Computational Linguistics (Volume 1: Long Papers)},
  year = {2024},
  pages = {1840--1873},
  publisher = {Association for Computational Linguistics},
  doi = {10.18653/v1/2024.acl-long.102},
  url = {https://aclanthology.org/2024.acl-long.102/}
}

@inproceedings{samuel2025personagym,
  title = {{PersonaGym}: Evaluating Persona Agents and {LLMs}},
  author = {Samuel, Vinay and Zou, Henry Peng and Zhou, Yue and Chaudhari, Shreyas and Kalyan, Ashwin and Rajpurohit, Tanmay and Deshpande, Ameet and Narasimhan, Karthik R and Murahari, Vishvak},
  booktitle = {Findings of the Association for Computational Linguistics: EMNLP 2025},
  year = {2025},
  pages = {6999--7022},
  publisher = {Association for Computational Linguistics},
  doi = {10.18653/v1/2025.findings-emnlp.368},
  url = {https://aclanthology.org/2025.findings-emnlp.368/}
}

@article{park2024generative1000,
  title = {Generative Agent Simulations of 1,000 People},
  author = {Park, Joon Sung and Zou, Carolyn Q. and Shaw, Aaron and Hill, Benjamin Mako and Cai, Carrie and Morris, Meredith Ringel and Willer, Robb and Liang, Percy and Bernstein, Michael S.},
  journal = {arXiv preprint arXiv:2411.10109},
  year = {2024},
  note = {Version 1},
  doi = {10.48550/arXiv.2411.10109},
  url = {https://arxiv.org/abs/2411.10109v1}
}

@article{rauthmann2014diamonds,
  title = {The Situational Eight {DIAMONDS}: A taxonomy of major dimensions of situation characteristics},
  author = {Rauthmann, John F and Gallardo-Pujol, David and Guillaume, Esther M and Todd, Elysia and Nave, Christopher S and Sherman, Ryne A and Ziegler, Matthias and Jones, Ashley Bell and Funder, David C},
  journal = {Journal of Personality and Social Psychology},
  year = {2014},
  volume = {107},
  number = {4},
  pages = {677--718},
  doi = {10.1037/a0037250}
}

@article{singer1995seeing,
  title = {Seeing one's self: Locating narrative memory in a framework of personality},
  author = {Singer, Jefferson A.},
  journal = {Journal of Personality},
  volume = {63},
  number = {3},
  pages = {429--457},
  year = {1995},
  publisher = {Wiley Online Library},
  doi = {10.1111/j.1467-6494.1995.tb00502.x}
}

@inproceedings{sun2025persona,
  title = {Persona-{DB}: Efficient Large Language Model Personalization for Response Prediction with Collaborative Data Refinement},
  author = {Sun, Chenkai and Yang, Ke and Gangi Reddy, Revanth and Fung, Yi and Chan, Hou Pong and Small, Kevin and Zhai, ChengXiang and Ji, Heng},
  booktitle = {Proceedings of the 31st International Conference on Computational Linguistics},
  pages = {281--296},
  year = {2025},
  url = {https://aclanthology.org/2025.coling-main.20/},
  publisher = {Association for Computational Linguistics}
}

@inproceedings{shao2023character,
  title = {Character-{LLM}: A Trainable Agent for Role-Playing},
  author = {Shao, Yunfan and Li, Linyang and Dai, Junqi and Qiu, Xipeng},
  booktitle = {Proceedings of the 2023 Conference on Empirical Methods in Natural Language Processing},
  pages = {13153--13187},
  year = {2023},
  url = {https://aclanthology.org/2023.emnlp-main.814/},
  publisher = {Association for Computational Linguistics},
  doi = {10.18653/v1/2023.emnlp-main.814}
}

@inproceedings{jiang2024personallm,
  title = {{P}ersona{LLM}: Investigating the Ability of Large Language Models to Express Personality Traits},
  author = {Jiang, Hang and Zhang, Xiajie and Cao, Xubo and Breazeal, Cynthia and Roy, Deb and Kabbara, Jad},
  booktitle = {Findings of the Association for Computational Linguistics: NAACL 2024},
  pages = {3605--3627},
  year = {2024},
  doi = {10.18653/v1/2024.findings-naacl.229},
  url = {https://aclanthology.org/2024.findings-naacl.229/},
  publisher = {Association for Computational Linguistics}
}

@inproceedings{chhikara2025mem0,
  title = {{Mem0}: Building Production-Ready {AI} Agents with Scalable Long-Term Memory},
  author = {Chhikara, Prateek and Khant, Dev and Aryan, Saket and Singh, Taranjeet and Yadav, Deshraj},
  year = {2025},
  month = {October},
  publisher = {IOS Press},
  booktitle = {{ECAI} 2025},
  doi = {10.3233/FAIA251160},
  url = {https://doi.org/10.3233/FAIA251160},
  pages = {2993--3000}
}

@inproceedings{lewis2020retrieval,
  title = {Retrieval-Augmented Generation for Knowledge-Intensive {NLP} Tasks},
  author = {Lewis, Patrick and Perez, Ethan and Piktus, Aleksandra and Petroni, Fabio and Karpukhin, Vladimir and Goyal, Naman and K\"{u}ttler, Heinrich and Lewis, Mike and Yih, Wen-tau and Rockt\"{a}schel, Tim and Riedel, Sebastian and Kiela, Douwe},
  volume = {33},
  pages = {9459--9474},
  year = {2020},
  url = {https://proceedings.neurips.cc/paper_files/paper/2020/file/6b493230205f780e1bc26945df7481e5-Paper.pdf},
  publisher = {Curran Associates, Inc.},
  booktitle = {Advances in Neural Information Processing Systems}
}

@inproceedings{hu2026clonemem,
  title = {{C}lone{M}em: Benchmarking Long-Term Memory for {AI} Clones},
  author = {Hu, Sen and Zhang, Zhiyu and Wei, Yuxiang and Han, Xueran and Tang, Zhenheng and Wang, Huacan and Chen, Ronghao},
  year = {2026},
  pages = {33571--33602},
  doi = {10.18653/v1/2026.acl-long.1549},
  url = {https://aclanthology.org/2026.acl-long.1549/},
  publisher = {Association for Computational Linguistics},
  address = {San Diego, California, United States},
  month = {July},
  booktitle = {Proceedings of the 64th Annual Meeting of the {A}ssociation for {C}omputational {L}inguistics (Volume 1: Long Papers)},
  editor = {Liakata, Maria and Moreira, Viviane P. and Zhang, Jiajun and Jurgens, David}
}

@inproceedings{maharana2024evaluating,
  title = {Evaluating Very Long-Term Conversational Memory of {LLM} Agents},
  author = {Maharana, Adyasha and Lee, Dong-Ho and Tulyakov, Sergey and Bansal, Mohit and Barbieri, Francesco and Fang, Yuwei},
  booktitle = {Proceedings of the 62nd Annual Meeting of the Association for Computational Linguistics (Volume 1: Long Papers)},
  pages = {13851--13870},
  year = {2024},
  doi = {10.18653/v1/2024.acl-long.747},
  url = {https://aclanthology.org/2024.acl-long.747/},
  publisher = {Association for Computational Linguistics}
}

@inproceedings{wu2024longmemeval,
  title = {{LongMemEval}: Benchmarking Chat Assistants on Long-Term Interactive Memory},
  author = {Wu, Di and Wang, Hongwei and Yu, Wenhao and Zhang, Yuwei and Chang, Kai-Wei and Yu, Dong},
  year = {2025},
  volume = {2025},
  url = {https://proceedings.iclr.cc/paper_files/paper/2025/file/d813d324dbf0598bbdc9c8e79740ed01-Paper-Conference.pdf},
  pages = {86809--86836},
  editor = {Y. Yue and A. Garg and N. Peng and F. Sha and R. Yu},
  booktitle = {International Conference on Learning Representations}
}

@article{mccrae1992introduction,
  title = {An introduction to the five-factor model and its applications},
  author = {McCrae, Robert R and John, Oliver P},
  journal = {Journal of Personality},
  volume = {60},
  number = {2},
  pages = {175--215},
  year = {1992},
  publisher = {Wiley Online Library},
  doi = {10.1111/j.1467-6494.1992.tb00970.x}
}

@book{erikson1963childhood,
  author={Erikson, Erik H.},
  title={Childhood and Society},
  edition={2nd},
  year={1963},
  publisher={W. W. Norton \& Company},
  address={New York}
}

@book{levinson1978seasons,
  author={Levinson, Daniel J. and Darrow, Charlotte N. and Klein, Edward B. and Levinson, Maria H. and McKee, Braxton},
  title={The Seasons of a Man's Life},
  year={1978},
  publisher={Knopf},
  address={New York}
}

@article{levinson1986conception,
  title = {A conception of adult development.},
  author = {Levinson, Daniel J.},
  journal = {American Psychologist},
  volume = {41},
  number = {1},
  pages = {3--13},
  year = {1986},
  publisher = {American Psychological Association},
  doi = {10.1037/0003-066X.41.1.3}
}

@book{young2006schema,
  publisher={Guilford Press},
  year={2006},
  author={Young, Jeffrey E. and Klosko, Janet S. and Weishaar, Marjorie E.},
  title={Schema Therapy: A Practitioner's Guide}
}

@article{rubin2006basic,
  title = {The Basic-Systems Model of Episodic Memory},
  author = {Rubin, David C.},
  journal = {Perspectives on Psychological Science},
  volume = {1},
  number = {4},
  pages = {277--311},
  year = {2006},
  publisher = {SAGE Publications},
  doi = {10.1111/j.1745-6916.2006.00017.x}
}

@article{bower1981,
  title = {Mood and memory.},
  author = {Bower, Gordon H},
  journal = {American Psychologist},
  volume = {36},
  number = {2},
  pages = {129--148},
  year = {1981},
  publisher = {American Psychological Association},
  doi = {10.1037/0003-066X.36.2.129}
}

@inproceedings{tu2024charactereval,
  title = {{C}haracter{E}val: A {C}hinese Benchmark for Role-Playing Conversational Agent Evaluation},
  author = {Tu, Quan and Fan, Shilong and Tian, Zihang and Shen, Tianhao and Shang, Shuo and Gao, Xin and Yan, Rui},
  booktitle = {Proceedings of the 62nd Annual Meeting of the Association for Computational Linguistics (Volume 1: Long Papers)},
  pages = {11836--11850},
  year = {2024},
  doi = {10.18653/v1/2024.acl-long.638},
  url = {https://aclanthology.org/2024.acl-long.638/},
  publisher = {Association for Computational Linguistics}
}

@article{ge2024scaling,
  title = {Scaling Synthetic Data Creation with 1,000,000,000 Personas},
  author = {Ge, Tao and Chan, Xin and Wang, Xiaoyang and Yu, Dian and Mi, Haitao and Yu, Dong},
  journal = {arXiv preprint arXiv:2406.20094},
  year = {2024},
  url = {https://arxiv.org/abs/2406.20094},
  doi = {10.48550/arXiv.2406.20094}
}

@article{chen2024persona,
  year = {2024},
  journal = {Transactions on Machine Learning Research},
  author = {Chen, Jiangjie and Wang, Xintao and Xu, Rui and Yuan, Siyu and Zhang, Yikai and Shi, Wei and Xie, Jian and Li, Shuang and Yang, Ruihan and Zhu, Tinghui and Chen, Aili and Li, Nianqi and Chen, Lida and Hu, Caiyu and Wu, Siye and Ren, Scott and Fu, Ziquan and Xiao, Yanghua},
  title = {From Persona to Personalization: A Survey on Role-Playing Language Agents},
  url = {https://openreview.net/forum?id=xrO70E8UIZ}
}

@inproceedings{tseng2024two,
  title = {Two Tales of Persona in {LLM}s: A Survey of Role-Playing and Personalization},
  author = {Tseng, Yu-Min and Huang, Yu-Chao and Hsiao, Teng-Yun and Chen, Wei-Lin and Huang, Chao-Wei and Meng, Yu and Chen, Yun-Nung},
  booktitle = {Findings of the Association for Computational Linguistics: EMNLP 2024},
  pages = {16612--16631},
  year = {2024},
  doi = {10.18653/v1/2024.findings-emnlp.969},
  url = {https://aclanthology.org/2024.findings-emnlp.969/},
  publisher = {Association for Computational Linguistics}
}

@article{wei2025ai,
  title = {{AI}-native Memory 2.0: {Second Me}},
  author = {Wei, Jiale and Ying, Xiang and Gao, Tao and Bao, Fangyi and Tao, Felix and Shang, Jingbo},
  journal = {arXiv preprint arXiv:2503.08102},
  year = {2025},
  url = {https://arxiv.org/abs/2503.08102},
  doi = {10.48550/arXiv.2503.08102}
}

@article{lee2026creating,
  publisher = {Elsevier},
  year = {2026},
  volume = {208},
  pages = {103692},
  journal = {International Journal of Human-Computer Studies},
  author = {Lee, Donggun and Lee, Suyoun and Lim, Hyunseung and Hong, Hwajung},
  title = {Creating text-based {AI} clones of myself: Exploring perceptions, development strategies, and challenges},
  doi = {10.1016/j.ijhcs.2025.103692}
}

@inproceedings{zhao2025personalens,
  title = {{P}ersona{L}ens: A Benchmark for Personalization Evaluation in Conversational {AI} Assistants},
  author = {Zhao, Zheng and Vania, Clara and Kayal, Subhradeep and Khan, Naila and Cohen, Shay B. and Yilmaz, Emine},
  booktitle = {Findings of the Association for Computational Linguistics: ACL 2025},
  pages = {18023--18055},
  year = {2025},
  doi = {10.18653/v1/2025.findings-acl.927},
  url = {https://aclanthology.org/2025.findings-acl.927/},
  publisher = {Association for Computational Linguistics}
}

@inproceedings{jiang2025personamem,
  title = {Know Me, Respond to Me: Benchmarking {LLMs} for Dynamic User Profiling and Personalized Responses at Scale},
  author = {Jiang, Bowen and Hao, Zhuoqun and Cho, Young-Min and Li, Bryan and Yuan, Yuan and Chen, Sihao and Ungar, Lyle and Taylor, Camillo J. and Roth, Dan},
  year = {2025},
  booktitle = {Second Conference on Language Modeling},
  url = {https://openreview.net/forum?id=6ox8XZGOqP}
}

@article{maples2024loneliness,
  title = {Loneliness and suicide mitigation for students using {GPT3}-enabled chatbots},
  author = {Maples, Bethanie and Cerit, Merve and Vishwanath, Aditya and Pea, Roy},
  journal = {npj Mental Health Research},
  volume = {3},
  number = {1},
  pages = {4},
  year = {2024},
  publisher = {Nature Publishing Group UK London},
  doi = {10.1038/s44184-023-00047-6}
}

@article{phang2025investigating,
  title = {Investigating affective use and emotional well-being on {ChatGPT}},
  author = {Phang, Jason and Lampe, Michael and Ahmad, Lama and Agarwal, Sandhini and Fang, Cathy Mengying and Liu, Auren R. and Danry, Valdemar and Lee, Eunhae and Chan, Samantha W. T. and Pataranutaporn, Pat and Maes, Pattie},
  journal = {arXiv preprint arXiv:2504.03888},
  year = {2025},
  url = {https://arxiv.org/abs/2504.03888},
  doi = {10.48550/arXiv.2504.03888}
}

@misc{deepseek_v3_2_2025,
  title = {{DeepSeek-V3.2}: Pushing the Frontier of Open Large Language Models},
  author = {{DeepSeek-AI} and Aixin Liu and Aoxue Mei and Bangcai Lin and Bing Xue and Bingxuan Wang and Bingzheng Xu and Bochao Wu and Bowei Zhang and Chaofan Lin and Chen Dong and Chengda Lu and Chenggang Zhao and Chengqi Deng and Chenhao Xu and Chong Ruan and Damai Dai and Daya Guo and Dejian Yang and Deli Chen and Erhang Li and Fangqi Zhou and Fangyun Lin and Fucong Dai and Guangbo Hao and Guanting Chen and Guowei Li and H. Zhang and Hanwei Xu and Hao Li and Haofen Liang and Haoran Wei and Haowei Zhang and Haowen Luo and Haozhe Ji and Honghui Ding and Hongxuan Tang and Huanqi Cao and Huazuo Gao and Hui Qu and Hui Zeng and Jialiang Huang and Jiashi Li and Jiaxin Xu and Jiewen Hu and Jingchang Chen and Jingting Xiang and Jingyang Yuan and Jingyuan Cheng and Jinhua Zhu and Jun Ran and Junguang Jiang and Junjie Qiu and Junlong Li and Junxiao Song and Kai Dong and Kaige Gao and Kang Guan and Kexin Huang and Kexing Zhou and Kezhao Huang and Kuai Yu and Lean Wang and Lecong Zhang and Lei Wang and Liang Zhao and Liangsheng Yin and Lihua Guo and Lingxiao Luo and Linwang Ma and Litong Wang and Liyue Zhang and M. S. Di and M. Y Xu and Mingchuan Zhang and Minghua Zhang and Minghui Tang and Mingxu Zhou and Panpan Huang and Peixin Cong and Peiyi Wang and Qiancheng Wang and Qihao Zhu and Qingyang Li and Qinyu Chen and Qiushi Du and Ruiling Xu and Ruiqi Ge and Ruisong Zhang and Ruizhe Pan and Runji Wang and Runqiu Yin and Runxin Xu and Ruomeng Shen and Ruoyu Zhang and S. H. Liu and Shanghao Lu and Shangyan Zhou and Shanhuang Chen and Shaofei Cai and Shaoyuan Chen and Shengding Hu and Shengyu Liu and Shiqiang Hu and Shirong Ma and Shiyu Wang and Shuiping Yu and Shunfeng Zhou and Shuting Pan and Songyang Zhou and Tao Ni and Tao Yun and Tian Pei and Tian Ye and Tianyuan Yue and Wangding Zeng and Wen Liu and Wenfeng Liang and Wenjie Pang and Wenjing Luo and Wenjun Gao and Wentao Zhang and Xi Gao and Xiangwen Wang and Xiao Bi and Xiaodong Liu and Xiaohan Wang and Xiaokang Chen and Xiaokang Zhang and Xiaotao Nie and Xin Cheng and Xin Liu and Xin Xie and Xingchao Liu and Xingkai Yu and Xingyou Li and Xinyu Yang and Xinyuan Li and Xu Chen and Xuecheng Su and Xuehai Pan and Xuheng Lin and Xuwei Fu and Y. Q. Wang and Yang Zhang and Yanhong Xu and Yanru Ma and Yao Li and Yao Li and Yao Zhao and Yaofeng Sun and Yaohui Wang and Yi Qian and Yi Yu and Yichao Zhang and Yifan Ding and Yifan Shi and Yiliang Xiong and Ying He and Ying Zhou and Yinmin Zhong and Yishi Piao and Yisong Wang and Yixiao Chen and Yixuan Tan and Yixuan Wei and Yiyang Ma and Yiyuan Liu and Yonglun Yang and Yongqiang Guo and Yongtong Wu and Yu Wu and Yuan Cheng and Yuan Ou and Yuanfan Xu and Yuduan Wang and Yue Gong and Yuhan Wu and Yuheng Zou and Yukun Li and Yunfan Xiong and Yuxiang Luo and Yuxiang You and Yuxuan Liu and Yuyang Zhou and Z. F. Wu and Z. Z. Ren and Zehua Zhao and Zehui Ren and Zhangli Sha and Zhe Fu and Zhean Xu and Zhenda Xie and Zhengyan Zhang and Zhewen Hao and Zhibin Gou and Zhicheng Ma and Zhigang Yan and Zhihong Shao and Zhixian Huang and Zhiyu Wu and Zhuoshu Li and Zhuping Zhang and Zian Xu and Zihao Wang and Zihui Gu and Zijia Zhu and Zilin Li and Zipeng Zhang and Ziwei Xie and Ziyi Gao and Zizheng Pan and Zongqing Yao and Bei Feng and Hui Li and J. L. Cai and Jiaqi Ni and Lei Xu and Meng Li and Ning Tian and R. J. Chen and R. L. Jin and S. S. Li and Shuang Zhou and Tianyu Sun and X. Q. Li and Xiangyue Jin and Xiaojin Shen and Xiaosha Chen and Xinnan Song and Xinyi Zhou and Y. X. Zhu and Yanping Huang and Yaohui Li and Yi Zheng and Yuchen Zhu and Yunxian Ma and Zhen Huang and Zhipeng Xu and Zhongyu Zhang and Dongjie Ji and Jian Liang and Jianzhong Guo and Jin Chen and Leyi Xia and Miaojun Wang and Mingming Li and Peng Zhang and Ruyi Chen and Shangmian Sun and Shaoqing Wu and Shengfeng Ye and T. Wang and W. L. Xiao and Wei An and Xianzu Wang and Xiaowen Sun and Xiaoxiang Wang and Ying Tang and Yukun Zha and Zekai Zhang and Zhe Ju and Zhen Zhang and Zihua Qu},
  year = {2025},
  url = {https://arxiv.org/abs/2512.02556},
  primaryclass = {cs.CL},
  archiveprefix = {arXiv},
  eprint = {2512.02556},
  doi = {10.48550/arXiv.2512.02556}
}

@misc{deepseek_v4_2026,
  title = {{DeepSeek-V4}: Towards Highly Efficient Million-Token Context Intelligence},
  author = {{DeepSeek-AI} and Anyi Xu and Bangcai Lin and Bing Xue and Bingxuan Wang and Bingzheng Xu and Bochao Wu and Bowei Zhang and Chaofan Lin and Chen Dong and Chenchen Ling and Chengda Lu and Chenggang Zhao and Chengqi Deng and Chengyu Hou and Chenhao Xu and Chenze Shao and Chong Ruan and Conner Sun and Damai Dai and Daya Guo and Dejian Yang and Deli Chen and Donghao Li and Dongjie Ji and Erhang Li and Fang Wei and Fangyun Lin and Fangzhou Yuan and Feiyu Xia and Fucong Dai and Guangbo Hao and Guanting Chen and Guoai Cao and Guolai Meng and Guowei Li and Han Yu and Han Zhang and Hanwei Xu and Hao Li and Haofen Liang and Haoling Zhang and Haoming Luo and Haoran Wei and Haotian Yuan and Haowei Zhang and Haowen Luo and Haoyu Chen and Haozhe Ji and Hengqing Zhang and Honghui Ding and Hongxuan Tang and Huanqi Cao and Huazuo Gao and Hui Qu and Hui Zeng and J Yang and JQ Zhu and Jia Luo and Jia Song and Jia Yu and Jialiang Huang and Jialu Cai and Jian Liang and Jiangting Zhou and Jiasheng Ye and Jiashi Li and Jiaxin Xu and Jiewen Hu and Jieyu Yang and Jin Chen and Jin Yan and Jingchang Chen and Jingli Zhou and Jingting Xiang and Jingyang Yuan and Jingyuan Cheng and Jingzi Zhou and Jinhua Zhu and Jiping Yu and Joseph Sun and Jun Ran and Junguang Jiang and Junjie Qiu and Junlong Li and Junmin Zheng and Junxiao Song and Kai Dong and Kaige Gao and Kang Guan and Kexing Zhou and Kezhao Huang and Kuai Yu and Lean Wang and Lecong Zhang and Lei Wang and Leyi Xia and Li Zhang and Liang Zhao and Lihua Guo and Lingxiao Luo and Linwang Ma and Linyan Zhu and Litong Wang and Liyu Cai and Liyue Zhang and Longhao Chen and MS Di and MY Xu and Max Mei and Miaojun Wang and Mingchuan Zhang and Minghua Zhang and Minghui Tang and Mingming Li and Mingxu Zhou and Minmin Han and Ning Wang and Panpan Huang and Panpan Wang and Peixin Cong and Peiyi Wang and Peng Zhang and Qiancheng Wang and Qihao Zhu and Qingyang Li and Qinyu Chen and Qiushi Du and Qiwei Jiang and Rui Tian and Ruifan Xu and Ruijie Lu and Ruiling Xu and Ruiqi Ge and Ruisong Zhang and Ruizhe Pan and Runji Wang and Runqian Chen and Runqiu Yin and Runxin Xu and Ruomeng Shen and Ruoyu Zhang and Ruyi Chen and SH Liu and Shanghao Lu and Shangmian Sun and Shangyan Zhou and Shanhuang Chen and Shaofei Cai and Shaoheng Nie and Shaoqing Wu and Shaoyuan Chen and Shengding Hu and Shengyu Liu and Shiqiang Hu and Shirong Ma and Shiyu Wang and Shuiping Yu and Shunfeng Zhou and Shuting Pan and Shuying Yu and Songyang Zhou and Tao Ni and Tao Yun and Tian Jin and Tian Pei and Tian Ye and Tianle Lin and Tianran Ji and Tianyi Cui and Tianyuan Yue and Tingting Yu and Tun Wang and W Zhang and WL Xiao and Wangding Zeng and Wei An and Weilin Zhao and Wen Liu and Wenfeng Liang and Wenjie Pang and Wenjing Luo and Wenjing Yao and Wenjun Gao and Wenkai Yang and Wenlve Huang and Wenqing Hou and Wentao Zhang and Wenting Ma and Xi Gao and Xiang He and Xiangwen Wang and Xianzu Wang and Xiao Bi and Xiaodong Liu and Xiaohan Wang and Xiaokang Chen and Xiaokang Zhang and Xiaotao Nie and Xiaowen Sun and Xiaoxiang Wang and Xin Cheng and Xin Liu and Xin Xie and Xingchao Liu and Xingchen Liu and Xingkai Yu and Xingyou Li and Xinyu Yang and Xinyu Zhang and Xu Chen and Xuanyu Wang and Xuecheng Su and Xueyin Chen and Xuheng Lin and Xuwei Fu and YC Yan and YQ Wang and YW Ma and Yanfeng Luo and Yang Zhang and Yanhong Xu and Yanru Ma and Yanwen Huang and Yao Li and Yao Li and Yao Xu and Yao Zhao and Yaofeng Sun and Yaohui Wang and Yi Qian and Yi Shao and Yi Yu and Yichao Zhang and Yifan Ding and Yifan Shi and Yijia Wu and Yiliang Xiong and Yiling Ma and Ying He and Ying Tang and Ying Zhou and Yingjia Luo and Yinmin Zhong and Yishi Piao and Yisong Wang and Yixiang Zhang and Yixiao Chen and Yixuan Tan and Yixuan Wei and Yiyang Ma and Yiyuan Liu and Yonglun Yang and Yongqiang Guo and Yongtong Wu and Yu Wu and YuKun Li and Yuan Cheng and Yuan Ou and Yuanfan Xu and Yuanhao Li and Yuduan Wang and Yuehan Yang and Yuer Xu and Yuhan Wu and Yuhao Meng and Yuheng Zou and Yukun Zha and Yunfan Xiong and Yupeng Chen and Yuping Lin and Yuqian Cao and Yuqian Wang and Yushun Zhang and Yuting Yan and Yutong Lin and Yuxian Gu and Yuxiang Luo and Yuxiang You and Yuxuan Liu and Yuxuan Zhou and Yuyang Zhou and Yuzhen Huang and ZF Wu and Zehao Wang and Zehua Zhao and Zehui Ren and Zekai Zhang and Zhangli Sha and Zhe Fu and Zhe Ju and Zhean Xu and Zhenda Xie and Zhengyan Zhang and Zheren Gao and Zhewen Hao and Zhibin Gou and Zhicheng Ma and Zhigang Yan and Zhihong Shao and Zhixian Huang and Zhixuan Chen and Zhiyu Wu and Zhizhou Ren and Zhongyu Wu and Zhuoshu Li and Zhuping Zhang and Zian Xu and Zihao Wang and Zihua Qu and Zihui Gu and Zijia Zhu and Zilin Li and Zipeng Zhang and Ziwei Xie and Ziyi Gao and Ziyi Wan and Zizheng Pan and Zongqing Yao},
  year = {2026},
  url = {https://arxiv.org/abs/2606.19348},
  primaryclass = {cs.CL},
  archiveprefix = {arXiv},
  eprint = {2606.19348},
  doi = {10.48550/arXiv.2606.19348}
}

@misc{qwen35_2026,
    title  = {{Qwen3.5}: Towards Native Multimodal Agents},
    author = {{Qwen Team}},
    month  = {February},
    year   = {2026},
    url    = {https://qwen.ai/blog?id=qwen3.5}
}

@article{qwen3embedding,
  title = {{Qwen3 Embedding}: Advancing Text Embedding and Reranking Through Foundation Models},
  author = {Zhang, Yanzhao and Li, Mingxin and Long, Dingkun and Zhang, Xin and Lin, Huan and Yang, Baosong and Xie, Pengjun and Yang, An and Liu, Dayiheng and Lin, Junyang and Huang, Fei and Zhou, Jingren},
  journal = {arXiv preprint arXiv:2506.05176},
  year = {2025},
  url = {https://arxiv.org/abs/2506.05176},
  doi = {10.48550/arXiv.2506.05176}
}

@misc{openai2026gpt54,
  author       = {{OpenAI}},
  title        = {Introducing {GPT-5.4}},
  howpublished = {\url{https://openai.com/index/introducing-gpt-5-4/}},
  year         = {2026},
  note         = {Accessed: 2026-05-06}
}

@misc{google2025gemini3flash,
  author = {{Google DeepMind}},
  title = {{Gemini 3 Flash} Model Card},
  howpublished = {\url{https://deepmind.google/models/model-cards/gemini-3-flash/}},
  year = {2025},
  note = {Accessed: 2026-10-12}
}

@misc{google2026gemini31pro,
  author = {{Google DeepMind}},
  title = {{Gemini 3.1 Pro} Model Card},
  howpublished = {\url{https://deepmind.google/models/model-cards/gemini-3-1-pro/}},
  year = {2026},
  note = {Accessed: 2026-10-12}
}

@misc{anthropic2025haiku45,
  author       = {{Anthropic}},
  title        = {Introducing {Claude Haiku 4.5}},
  howpublished = {\url{https://www.anthropic.com/news/claude-haiku-4-5}},
  year         = {2025},
  note         = {Accessed: 2026-05-06}
}

@misc{anthropic2026sonnet46,
  author       = {{Anthropic}},
  title        = {Introducing {Claude Sonnet 4.6}},
  howpublished = {\url{https://www.anthropic.com/news/claude-sonnet-4-6}},
  year         = {2026},
  note         = {Accessed: 2026-05-06}
}

@misc{anthropic2026opus46,
  author       = {{Anthropic}},
  title        = {Introducing {Claude Opus 4.6}},
  howpublished = {\url{https://www.anthropic.com/news/claude-opus-4-6}},
  year         = {2026},
  note         = {Accessed: 2026-05-06}
}

@article{pruksachatkun2026customersim,
  title = {{CustomerSim: Benchmarking and Aligning Multimodal Language Models as Retail User Simulators}},
  author = {Pruksachatkun, Yada and Wan, Yixin and Chen, Xingrun and Chang, Kai-Wei and Wu, Chien-Sheng},
  journal = {arXiv preprint arXiv:2605.08334},
  year = {2026},
  doi = {10.48550/arXiv.2605.08334},
  url = {https://arxiv.org/abs/2605.08334}
}

@article{in2026personalizethenstore,
  title = {{Personalize-then-Store: Benchmarking and Learning Personalized Memory for Long-horizon Agents}},
  author = {In, Yeonjun and Kim, Wonjoong and Park, Sangwu and Yoon, Kanghoon and Park, Chanyoung},
  journal = {arXiv preprint arXiv:2605.25535},
  year = {2026},
  doi = {10.48550/arXiv.2605.25535},
  url = {https://arxiv.org/abs/2605.25535}
}

@article{ding2026rootmemories,
  title = {{Towards Root Memories: Benchmarking and Enhancing Implicit Logical Memory Retrieval for Personalized LLMs}},
  author = {Ding, Hongxun and Yu, Xiang and Wang, Chengbing and Xiao, Jianfei and Bao, Keqin and Wang, Wenjie and He, Xiangnan},
  journal = {arXiv preprint arXiv:2606.23283},
  year = {2026},
  doi = {10.48550/arXiv.2606.23283},
  url = {https://arxiv.org/abs/2606.23283}
}

@article{jiang2026personamemv3,
  title = {{PersonaMem-v3: Toward Omni-Platform Personal Intelligence for Holistic User Understanding, Recommendation, and Agentic Tasks}},
  author = {Jiang, Bowen and Yuan, Yuan and Hao, Zhuoqun and Liu, Yuchen and Shen, Maohao and Chen, Sihao and Wornell, Gregory and Callison-Burch, Chris and Ungar, Lyle and Roth, Dan and Fostiropoulos, Iordanis and Guo, Qi and Fan, Xiangjun and Taylor, Camillo J. and Yu, Hanchao},
  journal = {arXiv preprint arXiv:2608.21381},
  year = {2026},
  doi = {10.48550/arXiv.2608.21381},
  url = {https://arxiv.org/abs/2608.21381v2},
  note = {Version 2}
}

@article{yin2026usertoolbench,
  title = {{UserToolBench: A User-Profile-Hidden Benchmark for Personalized Decision Making in Tool-Use LLMs}},
  author = {Yin, Xuexiong and Chen, Zechuan and Zheng, Yongsen and Zhang, Yuxiang and Yang, Jingyuan and Wang, Bin and Wang, Yubin and Wang, Keze},
  journal = {arXiv preprint arXiv:2608.10042},
  year = {2026},
  doi = {10.48550/arXiv.2608.10042},
  url = {https://arxiv.org/abs/2608.10042}
}

@article{huang2026agentpersonabench,
  title = {{AgentPersonaBench: Benchmarking Persona-Driven User Simulation}},
  author = {Huang, Jintao and Wang, Yifan and Shen, Hongyu and Lu, Yi Daniel and Huang, Shirley and Oh, Minsik and Wang, Yewen and Mohsin, Muhammad Ahmed and Xu, Zhen and Fan, Yilan and Yuan, Zichen and Bilal, Ahsan and Wei, Zibu and Jajee, Sankalp and Gagnier, Henry and Kapoor, Saksham and Wang, Jicheng and Wen, Qianfeng and He, Yixuan and Dillmann, Steven and He, Jiashu and Lu, Yucheng and Guo, Linqiang and Zhang, Danyang and Bo, Shi and Mondal, Raunak and Tang, Haixiang and Xiao, Weihang and Nie, Allen and Tang, Jing and Li, Yueying and Liu, Yifan Simon and Hou, Jianheng and Wang, Dianzhuo and Zhu, Qianyu and Tao, Zhixu Silvia and Peng, Zhejian and Wang, Zihan and Gupta, Ishan and Fan, Jinxuan and Jiang, Wanting and Liang, Shushu and Qiu, Chenxi and Wang, Yijun and Li, Xiaomin and Hao, Yuexing},
  journal = {arXiv preprint arXiv:2610.04379},
  year = {2026},
  doi = {10.48550/arXiv.2610.04379},
  url = {https://arxiv.org/abs/2610.04379}
}

@article{liu2026perma,
  title = {{PERMA: Benchmarking Personalized Memory Agents via Event-Driven Preference and Realistic Task Environments}},
  author = {Liu, Shuochen and Zhu, Junyi and Shu, Long and Lin, Junda and Chen, Yuhao and Zhang, Haotian and Zhang, Chao and Xu, Derong and Li, Jia and Tang, Bo and Li, Zhiyu and Xiong, Feiyu and Chen, Enhong and Xu, Tong},
  journal = {arXiv preprint arXiv:2603.23231},
  year = {2026},
  doi = {10.48550/arXiv.2603.23231},
  url = {https://arxiv.org/abs/2603.23231v2},
  note = {Version 2, revised May 17, 2026}
}

@inproceedings{xiao2026alpsbench,
  title = {{AlpsBench: An LLM Personalization Benchmark for Real-Dialogue Memorization and Preference Alignment}},
  author = {Xiao, Jianfei and Yu, Xiang and Wang, Chengbing and Zheng, Wuqiang and Lin, Xinyu and Liu, Kaining and Ding, Hongxun and Zhang, Yang and Wang, Wenjie and Feng, Fuli and He, Xiangnan},
  booktitle = {Proceedings of the 49th International ACM SIGIR Conference on Research and Development in Information Retrieval},
  year = {2026},
  month = jul,
  pages = {3489--3497},
  publisher = {Association for Computing Machinery},
  doi = {10.1145/3805712.3808634},
  url = {https://doi.org/10.1145/3805712.3808634}
}

@inproceedings{wang2026memoryroleplaying,
    title = "Memory-Driven Role-Playing: Evaluation and Enhancement of Persona Knowledge Utilization in {LLM}s",
    author = "Wang, Kai  and
      You, Haoyang  and
      Zhang, Yang  and
      Wang, Zhongjie",
    editor = "Liakata, Maria  and
      Moreira, Viviane P.  and
      Zhang, Jiajun  and
      Jurgens, David",
    booktitle = "Findings of the {A}ssociation for {C}omputational {L}inguistics: {ACL} 2026",
    month = jul,
    year = "2026",
    address = "San Diego, California, United States",
    publisher = "Association for Computational Linguistics",
    url = "https://aclanthology.org/2026.findings-acl.1175/",
    doi = "10.18653/v1/2026.findings-acl.1175",
    pages = "23475--23510",
    ISBN = "979-8-89176-395-1"
}

@misc{openai2026gpt6luna,
  author = {{OpenAI}},
  title = {{GPT-6 Luna} Model},
  howpublished = {\url{https://developers.openai.com/api/docs/models/gpt-6-luna}},
  year = {2026},
  note = {Accessed: 2026-10-08}
}

@misc{openai2026gpt56sol,
  author = {{OpenAI}},
  title = {{GPT-5.6 Sol} Model},
  howpublished = {\url{https://developers.openai.com/api/docs/models/gpt-5.6-sol}},
  year = {2026},
  note = {Accessed: 2026-10-08}
}

@misc{google2026gemini38flash,
  author = {{Google DeepMind}},
  title = {{Gemini 3.8 Flash} Model Card},
  howpublished = {\url{https://deepmind.google/models/model-cards/gemini-3-8-flash/}},
  year = {2026},
  note = {Accessed: 2026-10-12}
}

@misc{anthropic2026sonnet55,
  author = {{Anthropic}},
  title = {Introducing {Claude Sonnet 5.5}},
  howpublished = {\url{https://www.anthropic.com/claude-sonnet-5-5}},
  year = {2026},
  note = {Accessed: 2026-10-08}
}

@misc{openai2026gpt54mini,
  author = {{OpenAI}},
  title = {{GPT-5.4 Thinking System Card}},
  year = {2026},
  url = {https://deploymentsafety.openai.com/gpt-5-4-thinking/gpt-5-4-thinking.pdf},
  note = {Section 6: Appendix: GPT-5.4 mini, added March 17, 2026. Accessed: 2026-10-12}
}

@article{jiang2025personamemv2,
  title = {{PersonaMem-v2: Towards Personalized Intelligence via Learning Implicit User Personas and Agentic Memory}},
  author = {Jiang, Bowen and Yuan, Yuan and Shen, Maohao and Hao, Zhuoqun and Xu, Zhangchen and Chen, Zichen and Liu, Ziyi and Vijjini, Anvesh Rao and He, Jiashu and Yu, Hanchao and Poovendran, Radha and Wornell, Gregory and Ungar, Lyle and Roth, Dan and Chen, Sihao and Taylor, Camillo Jose},
  journal = {arXiv preprint arXiv:2512.06688},
  year = {2025},
  doi = {10.48550/arXiv.2512.06688},
  url = {https://arxiv.org/abs/2512.06688}
}

\appendix
\section{Benchmark Comparison Dimensions}
\label{app:benchmark_dims}

Table~\ref{tab:benchmark_comparison} in the main text compares \ourbench with related benchmarks across seven dimensions. We define each dimension below.

\begin{itemize}[leftmargin=1.5em, itemsep=4pt]
  \item \textbf{Big Five-grounded.} The benchmark explicitly adopts the Big Five (OCEAN) personality framework as the basis for character design---not by merely mentioning personality, but by operationalizing it through structured trait dimensions.

  \item \textbf{Age-defined life-stage coverage.} The benchmark organizes synthetic memories into age-defined life-stage bins (e.g., childhood, adolescence, adulthood) informed by developmental theories (e.g., Erikson, Levinson).

  \item \textbf{Episodic memory.} Full coverage uses structured autobiographical or event-level memory records. Partial coverage supplies past experiences within source narratives or conversational histories.

  \item \textbf{Scenario eval.} Evaluation requires a model to select a concrete response within a structured situational context. Pure factual recall, question answering, or stylistic consistency checks do not qualify.

  \item \textbf{Expert annot.} Full coverage involves domain experts directly assigning reference labels or reviewing item quality. Partial coverage uses expert-authored source analyses or expert involvement in a subset of construction or validation.

  \item \textbf{In-character decision.} The evaluation target is the action or response that the target character would choose in a situation, judged against its specification or narrative history. Personalized assistant replies to a user are a separate task target, covered by the scenario-evaluation column.

  \item \textbf{Profile-hidden evaluation.} The standard evaluation withholds a separately supplied construction-time persona description or trait vector. Evidence can come from histories and representations derived from them, with basic identity information still available. Partial coverage denotes a reported memory-only configuration alongside description-conditioned configurations. This criterion concerns the evaluated model's inputs; profiles used by data generators or judges are construction information. PersonaHub is marked not applicable because it is a persona-synthesis resource without a comparable evaluation protocol.
\end{itemize}

LifeChoice receives partial episodic-memory coverage for its preceding narrative passages and partial expert-annotation coverage for literary-expert analyses used as construction material; university students review the generated items~\cite{xu2025lifechoice}. Its prediction of the character's decision qualifies as in-character decision evaluation. Its memory-only configuration withholds the character description, while its description-only and combined configurations supply it; profile-hidden evaluation is therefore marked partial. PersonaMem and PersonaMem-v2 receive partial episodic-memory coverage for their conversational histories and scenario-evaluation coverage for contextualized response selection; their output target is a personalized assistant reply~\cite{jiang2025personamem,jiang2025personamemv2}.

\section{Character Construction and Memory Pipeline}
\label{app:character}

This appendix expands the three-stage character-construction pipeline described in the main text (\S\ref{sec:character}): profile specification, autobiographical-memory synthesis, and quality audit. For implementation detail, the synthesis stage is decomposed below into memory-quota allocation and memory generation; we also report the unified memory schema and resulting dataset statistics.

\subsection{Profile Specification}
\label{app:stage_a}
\begin{table}[t]
\centering
\resizebox{\linewidth}{!}{
\begin{tabular}{llcccccl}
\toprule
\textbf{ID} & \textbf{Archetype} &
\textbf{O} & \textbf{C} & \textbf{E} & \textbf{A} & \textbf{N} &
\textbf{Occupation} \\
\midrule
CHAR\_01 & N-High  & .85 & .50 & .30 & .55 & \textbf{.95} & Freelance content creator \\
CHAR\_02 & N-Low   & .50 & .55 & .50 & .60 & \textbf{.10} & Community GP \\
CHAR\_03 & C-High  & .45 & \textbf{.95} & .40 & .60 & .30 & Project manager \\
CHAR\_04 & C-Low   & .80 & \textbf{.10} & .60 & .65 & .45 & Freelance programmer \\
CHAR\_05 & E-High  & .65 & .55 & \textbf{.95} & .70 & .35 & Sales director \\
CHAR\_06 & E-Low   & .75 & .70 & \textbf{.10} & .45 & .50 & Indie game developer \\
CHAR\_07 & A-High  & .55 & .60 & .55 & \textbf{.95} & .55 & Non-profit coordinator \\
CHAR\_08 & A-Low   & .65 & .85 & .45 & \textbf{.10} & .35 & Startup co-founder \\
CHAR\_09 & O-High  & \textbf{.95} & .35 & .65 & .55 & .40 & Anthropology lecturer \\
CHAR\_10 & O-Low   & \textbf{.10} & .75 & .40 & .55 & .45 & Building-materials retailer \\
CHAR\_11 & Neutral & .50 & .50 & .50 & .50 & .50 & Local journalist \\
\bottomrule
\end{tabular}
}
\caption{The 11 character archetypes. Bolded scores mark the defining extreme dimension; CHAR\_11 is the all-neutral baseline.}
\label{tab:characters}
\end{table}

Each of the 11 character records in \texttt{benchmark/characters.json} in the public code repository stores the complete specification $P$ together with its memory history. Table~\ref{tab:characters} lists the Big Five configurations and their defining trait anchors. The profile fields and their construction-time roles are summarized below:

\begin{itemize}[leftmargin=1.5em, itemsep=2pt]
  \item \textbf{Character label} (\texttt{name}): descriptive construction metadata encoding the archetype, supplied to memory and candidate-response generation and withheld from the main evaluation.
  \item \textbf{Self-value logic} (\texttt{self\_value\_logic}): a brief statement of the character's core operating logic, written to elaborate the intended personality configuration.
  \item \textbf{Core behavioral patterns} (\texttt{core\_patterns}, 8 items): recurring behavioral tendencies supplied as constraints to memory generation.
  \item \textbf{Occupation}: a role selected for compatibility with the defining trait in the high/low archetypes and reviewed within the complete character specification. It provides recurring context for work-related memories and, together with the character ID, forms the visible basic information.
  \item \textbf{Trait-anchored coverage design}: One $.95$ and one $.10$ anchor per Big Five dimension, together with an all-$.50$ baseline, cover both poles. Non-anchor scores remain nonuniform, preserving coherent multi-trait profiles for character-level comparison. The neutral profile serves as a midpoint reference across dimensions.
  \item \textbf{Profile persistence across the pipeline}: Each specification remains fixed across its $1{,}000$ memories and corresponding character--scenario instances; life stage, situation, and retrieved evidence vary without redefining the character. The same character identity is therefore preserved across retrieval settings.
\end{itemize}

\begin{table*}[!t]
\centering
\small
\resizebox{\textwidth}{!}{%
\begin{tabular}{p{1.5cm}p{4.2cm}p{8.5cm}}
\toprule
\textbf{ID} & \textbf{Self-Value Logic} & \textbf{Representative Core Patterns (of 8)} \\
\midrule
CHAR\_01 \newline N-High &
Pre-emptive self-attack is the primary defense against anticipated rejection. &
Catastrophizing minor setbacks; ruminating on past mistakes; over-interpreting social micro-expressions. \\
\midrule
CHAR\_02 \newline N-Low &
Facts take priority over emotion; address solvable problems and accept what cannot be changed. &
Staying calm under pressure; waiting calmly when information is ambiguous; maintaining focus amid chaos. \\
\midrule
CHAR\_03 \newline C-High &
Order is maintained through information and planning; rebuilding structure helps resolve discomfort when rules fail. &
Keeping exhaustive task lists; mild anxiety when rules fail, eased by restoring structure; correcting minor formatting deviations. \\
\midrule
CHAR\_04 \newline C-Low &
Prioritizes immediate feeling and flow; treats plans as constraints and deadlines as others' rules. &
Acting before considering consequences; disregarding deadlines and commitments; rationalizing procrastination afterward. \\
\midrule
CHAR\_05 \newline E-High &
Draws energy from others; sees conflict as relationship-building and favors public problem-solving. &
Initiating social contact within minutes; thinking aloud by default; solving problems in front of an audience. \\
\midrule
CHAR\_06 \newline E-Low &
Finds social interaction draining and solitude energizing; protects focused work from outside interruption. &
Recovering alone after social interaction; preferring asynchronous communication; using polite but distant language to maintain boundaries. \\
\midrule
CHAR\_07 \newline A-High &
Adequate communication can resolve most disagreements; preserving harmony motivates mediation. &
Seeking mediation in disputes; apologizing and self-blaming when criticized; expressing disagreement indirectly. \\
\midrule
CHAR\_08 \newline A-Low &
Uses cost-benefit reasoning and treats logic as the basis for decisions. &
Analyzing power structures; adhering to rules and fairness; speaking directly without emotional padding. \\
\midrule
CHAR\_09 \newline O-High &
Approaches unfamiliar things as material to interpret and questions fixed answers. &
Exploring counterintuitive ideas; questioning established conclusions; shifting attention to new interests and delaying current tasks. \\
\midrule
CHAR\_10 \newline O-Low &
Trusts established routines and tangible evidence; approaches change with suspicion. &
Keeping suppliers, replenishment, and pricing stable; distrusting innovation rhetoric; evaluating durability and price. \\
\midrule
CHAR\_11 \newline Neutral &
Observes before taking a position and adapts to circumstances without strong self-assertion. &
Following flexible routines; being comfortable alone or with others; gathering information before taking a position in conflict. \\
\bottomrule
\end{tabular}%
}
\caption{Character specifications: self-value logic and representative behavioral patterns.}
\label{tab:persona}
\end{table*}

Table~\ref{tab:persona} condenses each stored self-value logic and three of its eight core behavioral patterns. Full 8-pattern lists are provided in \texttt{benchmark/characters.json} in the public code repository.

\subsection{Operationalizing the Big Five Profiles}
\label{app:profile_grounding}

The OCEAN scores are predefined design coordinates on $[0,1]$, not questionnaire measurements or population percentiles. High and low anchors differ by $0.85$ on the defining dimension, and each differs from the neutral baseline by at least $0.40$. Non-anchor dimensions vary to maintain coherent complete profiles; their suggested range guides initial generation. For the ten high/low archetypes, the defining trait guides occupation selection and the initial narrative emphasis. The character description, \texttt{self\_value\_logic}, and \texttt{core\_patterns} specify how the selected persona interprets situations and tends to act. Expert review assesses the coherence of the complete specification, including non-anchor traits and occupational context (Appendix~\ref{app:profile_review}). CHAR\_11 is the separately specified all-$.50$ reference: its narrative emphasizes observation and situational adaptation, with individual responses conditioned on the scenario and history.

The finalized specification $P$ remains fixed as the age window and episode context vary. Memory generation receives the descriptive character label, Big Five vector, occupation, description, self-value logic, and all eight behavioral patterns. Candidate-response generation receives the ID and label, Big Five vector, description, and self-value logic together with the scenario and activated memories; the patterns also shape its experiential evidence through the generated memories. Experts adjudicate candidate responses against the predefined character, its supplied history, and the immediate decision context.

\subsection{Character Profile Review}
\label{app:profile_review}

\paragraph{Materials and review procedure.}
The three researchers described in Section~\ref{sec:character} each reviewed all 11 initial character specifications, including both basic information and the personality-and-value profile. The review materials comprised the full Big Five vector and intended trait anchor, character description, occupation, self-value logic, and eight core behavioral patterns. The guideline below specifies how to assess the complete configuration and give an overall recommendation. For a requested revision, it asks the reviewer to identify the affected field or passage, explain the concern, and propose a targeted correction.

\paragraph{Review scope and criteria.}
The five criteria address trait alignment, internal coherence, occupational plausibility, behavioral specificity, and interpretive restraint, establishing whether the complete specification is suitable for subsequent memory construction. Review retains the intended defining trait anchor while allowing non-anchor scores outside the suggested 0.30--0.70 range when the complete specification is internally coherent. The all-neutral CHAR\_11 follows a separate predefined configuration with all five scores fixed at 0.50.

\begin{tcolorbox}[
  breakable,
  colback=blue!5,
  colframe=blue!60!black,
  title={\textbf{Character Profile Review Guideline}},
  fonttitle=\bfseries,
  rounded corners,
  boxrule=1.0pt
]
\begin{Verbatim}[fontsize=\scriptsize, breaklines=true, breakanywhere=true, breaksymbolleft={}, breaksymbolright={}, breakindent=1em]
Task
Assess whether the supplied synthetic character profile is coherent,
aligned with its stated design, and sufficiently explicit to guide
subsequent autobiographical-memory generation.

Materials
Read the complete Big Five vector and intended trait anchor, character
description, occupation and available rationale, self-value logic,
and eight core behavioral patterns.
Identify any missing information needed for your judgment.

Review criteria
1. Trait alignment: Assess compatibility with the full five-dimensional
   configuration, not only the defining high/low trait. A trait score
   does not uniquely determine an action or profession.
2. Internal coherence: Check whether the description, self-value logic,
   and behavioral patterns are mutually compatible. Flag unexplained
   contradictions; allow contextual variation when it is justified.
3. Occupational plausibility: Assess whether the character could
   plausibly hold the stated occupation. An atypical trait-occupation
   combination alone is not grounds for rejection; evaluate the
   rationale and its consistency with the rest of the profile.
4. Behavioral specificity: Check that the eight patterns are concrete,
   observable, and distinct enough to guide narratives in relevant
   situations. Flag vague labels, repetition, and requirements that
   every pattern appear in every episode.
5. Interpretive restraint: Flag unsupported clinical diagnoses, moral
   judgments, and overly categorical claims. Extreme trait values do
   not imply a disorder, and neutral values do not require a moderate
   response in every situation.

Recommendation and explanation
- Acceptable as-is: no issue requires a change to the profile.
- Local revision needed: the core specification is usable, but a
  specific wording, scope, or consistency issue requires correction.
- Flag for further adjudication: substantial contradictions or missing
  information prevent either recommendation above.

For each requested change, identify the field and relevant passage,
explain the concern, and suggest a targeted correction. Do not prefer
fluent or stereotypical descriptions over coherent specifications.
Do not change a trait score solely to accommodate problematic prose.
Report your overall recommendation and any unresolved concern.
\end{Verbatim}
\end{tcolorbox}

\paragraph{Initial review outcomes.}
Table~\ref{tab:profile_review_outcomes} summarizes the three researchers' initial recommendations at the profile level. Eight profiles (72.7\%) were accepted as-is by all three reviewers. Of the remaining three profiles, two (18.2\%) received a request for local revisions from one reviewer, and one (9.1\%) received such requests from two reviewers.

\begin{table}[t]
\centering
\small
\setlength{\tabcolsep}{3pt}
\begin{tabularx}{\linewidth}{Xrr}
\toprule
\textbf{Reviewer recommendations} & \textbf{Count} & \textbf{\%} \\
\midrule
All three accept as-is & 8 & 72.7 \\
One requests local revisions & 2 & 18.2 \\
Two request local revisions & 1 & 9.1 \\
\midrule
Total & 11 & 100.0 \\
\bottomrule
\end{tabularx}
\caption{Initial review outcomes for the 11 character profiles. Counts refer to profiles; percentages use 11 as the denominator.}
\label{tab:profile_review_outcomes}
\end{table}

\paragraph{Profile revisions.}
The three profiles receiving requests for local revisions were CHAR\_03, CHAR\_07, and CHAR\_08. For CHAR\_03, we clarified the intensity and situational scope of anxiety-related behavior: anxiety is mild, arises when rules fail, and is alleviated by restoring structure. For CHAR\_07, we qualified general claims about conflict resolution, framing successful mediation as possible for most disagreements given adequate communication. For CHAR\_08, we made rule- and fairness-based constraints explicit alongside the character's instrumental decision-making style.

\paragraph{Adjudication and finalization.}
Profiles accepted as-is by all three researchers were retained unchanged. For the other three profiles, the senior professor adjudicated the revision requests, made the corresponding field-level edits, and performed the final suitability check. The eight unchanged and three revised specifications formed the accepted character set; each specification $P$ was fixed before subsequent autobiographical-memory generation. Memory generation used these accepted specifications as its character constraints. Later, experts adjudicated candidate responses against the corresponding finalized profile, associated memories, and scenario.

\subsection{Memory Quota Allocation}
\label{app:stage_b}

The 45-year span (ages 6--50) is divided into 45 one-year segments. For a character that already has $k$ memories, the algorithm computes the deficit $d = 1{,}000 - k$ and distributes it proportionally across segments according to a construction-time weight vector $\mathbf{w}$:

\begin{equation}
  q_t = \left\lfloor d \cdot \frac{w_t}{\sum_j w_j} \right\rfloor + \epsilon_t
\end{equation}

where $q_t$ is the quota for segment $t$, $w_t$ is higher for selected life-stage bins (Adolescence, Early Adult Transition, and Age-30 Transition), and $\epsilon_t \in \{0,1\}$ is a rounding correction to ensure $\sum_t q_t = d$ exactly. The weighting is inspired by findings on the temporal clustering of self-defining memories~\cite{singer1995seeing} and specifies the synthetic data distribution.

\begin{table}[!t]
\centering

\resizebox{\linewidth}{!}{
\begin{tabular}{lrr}
\toprule
\textbf{Stage} & \textbf{Count} & \textbf{Proportion} \\
\midrule
School age (6--12)              & 1,812 & 16.5\% \\
Adolescence (13--18)            & 1,793 & 16.3\% \\
Early adult trans.\ (19--22)    & 1,286 & 11.7\% \\
Entering adult world (23--27)   & 1,489 & 13.5\% \\
Age-30 transition (28--33)      & 1,631 & 14.8\% \\
Settling down (34--40)          & 1,554 & 14.1\% \\
Midlife transition (41--45)     &   795 &  7.2\% \\
Entering midlife (46--50)       &   640 &  5.8\% \\
\midrule
\textbf{Total} & \textbf{11,000} & \textbf{100\%} \\
\bottomrule
\end{tabular}
}
\caption{Aggregate memory distribution by event age using the non-overlapping reporting intervals shown (all 11 characters, $N=11{,}000$).}
\label{tab:stage_dist}
\end{table}

\subsection{Memory Generation Protocol}
\label{app:stage_c}

Memories are generated using Claude-Sonnet-4.6 via the Anthropic Messages API. Each batch covers a 2--5-year age window and targets 45--50 memories. The structured system prompt injects three constraint layers:

\begin{enumerate}[leftmargin=1.5em, itemsep=2pt]
  \item \textbf{Character constraints}: full Big Five vector, occupation, self-value logic, and all eight core behavioral patterns from Appendix~\ref{app:stage_a}.
  \item \textbf{Age-bin context}: target age window, life-stage-bin label, and the associated construction guidance from Table~\ref{tab:stage_dist} (Appendix~\ref{app:stage_b}).
  \item \textbf{Narrative-dimension coverage}: guidance requiring organic integration of our five operational dimensions, inspired by Rubin's basic-systems account~\cite{rubin2006basic}, without explicit section labels.
  Unlike a short fact record (e.g., ``failed a job interview''), each \texttt{content\_full} narrative is generated to represent multiple facets of an episode rather than only its propositional summary.
  They are:
  \begin{enumerate}[leftmargin=1.5em, itemsep=1pt]
    \item \textit{Sensory detail}---generated sights, sounds, smells, or tactile cues that situate the narrative in a specific time and place;
    \item \textit{Dialogue reconstruction}---generated verbatim or near-verbatim exchanges that represent interpersonal dynamics and relational positioning beyond a short event summary;
    \item \textit{Inner monologue}---generated in-the-moment appraisals designed to encode cues from the character's trait vector $\mathbf{p}_i$ (e.g., catastrophizing for high-$N$);
    \item \textit{Somatic response}---described physiological cues (heart rate, muscle tension, or nausea) used to encode the intended affective intensity;
    \item \textit{Aftermath}---generated near-term consequences representing how the episode affects the character specification's beliefs or coping patterns.
  \end{enumerate}
\end{enumerate}

To prevent within-batch duplication, the prompt seeds topic diversity across five scenario domains: workplace, interpersonal, health, creative, and daily life.

\paragraph{Structured output format.}
The generation prompt requests one JSON object per memory. After parsing and validation, each record is serialized for incremental storage under a unique header (\texttt{===MEM\_XX\_NNN===}), followed by a metadata preamble in \texttt{key: value} format and a narrative body separated by a \texttt{---content\_full---} delimiter. This storage serialization is distinct from the model-facing JSON output.

\subsection{Quality Audit and Repair}
\label{app:stage_d}

The audit screens each memory against six defect criteria using pattern-based checks and LLM-based classifiers. Flagged records enter the targeted repair loop.

\begin{enumerate}[leftmargin=1.5em, itemsep=4pt]
  \item \textbf{Perspective Drift (Omniscient Narrator).} Violating the first-person narrative constraint. Detected when the narrator uses post-hoc knowledge or future-tense markers (e.g., ``years later,'' ``I didn't know then''). This violates the strict ``in-the-moment'' constraint.

  \item \textbf{Perspective Drift (Adult Evaluation).} Specifically flags child-age memories ($\leq 12$) that use adult-level cognitive appraisals or retrospective phrases (e.g., ``looking back,'' ``I now understand'').

  \item \textbf{Field-Length Inversion.} Narrative detail spilling into \texttt{context} or \texttt{content\_summary}. These fields provide concise scene metadata and an event gist; detailed episode narration belongs in \texttt{content\_full}.

  \item \textbf{Semantic Duplication.} Redundant life episodes within the same developmental age window. This is detected using a cosine similarity threshold of $> 0.85$ over TF-IDF vector representations of the narratives.

  \item \textbf{Length Violation.} \texttt{content\_full} exceeding 10{,}000 characters. The 2{,}000--4{,}500-character range guides generation.

  \item \textbf{Schema-Vocabulary Inflation.} Use of high-register psychoanalytic terms (e.g., ``abandonment schema'', ``dysregulation'') in low-intensity episodes ($< 0.4$), indicating over-psychologization.
\end{enumerate}

\paragraph{Repair procedure.}
Flagged memories enter a targeted repair loop: only the defective field(s) are regenerated using an issue-specific instruction appended to the original generation prompt. For example, a perspective-drift failure supplies the original \texttt{context} and \texttt{content\_summary} and instructs the model to rewrite \texttt{content\_full} from a strict first-person viewpoint. Up to three automated repair attempts are made per defect; records still flagged after this process are held for manual inspection before admission. Human review can confirm a defect or override a false-positive LLM flag.

\paragraph{Human review and audit outcomes.}
Of the 11{,}000 memories, 8{,}991 passed the initial automated audit, 1{,}935 passed after automated rewriting, and the remaining 74 underwent manual review by the same four-expert team introduced in Section~\ref{sec:character}. The human review guidelines were substantially aligned with the six defect criteria above. The team confirmed that 71 of the 74 memories required revision. The senior professor edited these 71 memories, all of which subsequently passed re-evaluation by the LLM judge. For the remaining three, the reviewers unanimously judged the flagged records acceptable without further revision and identified the LLM flags as false positives. All 74 records were retained following these respective resolution paths. In the post-audit corpus, all 11{,}000 retained memories have the complete ten-field schema in Table~\ref{tab:schema} and nonempty \texttt{content\_full} narratives.

The following examples illustrate the revision directions for three of the 71 manually edited memories.
\begin{enumerate}[leftmargin=1.5em, itemsep=4pt]
  \item \textbf{Restricting narrator knowledge (criterion 1).} In \texttt{MEM\_CHAR\_03\_C\_HIGH\_0002}, a six-year-old responds to a parental argument by tidying the room. The narrator explicitly does not know whether the father heard this activity or has calmed down on the balcony. The revision restricted narration to observations available at the time and made uncertainty about another person's unobserved state explicit.
  \item \textbf{Keeping appraisals age-appropriate (criterion 2).} In \texttt{MEM\_CHAR\_06\_E\_LOW\_0006}, a six-year-old whose blocks repeatedly collapse wonders whether their placement or the slippery floor caused the problem. Frustration is expressed through immediate thoughts, bodily sensations, and actions. The revision grounded the explanation in the child's immediate situational understanding, feelings, and behavior.
  \item \textbf{Using ordinary language for low-intensity events (criterion 6).} In \texttt{MEM\_CHAR\_02\_N\_LOW\_0006}, an episode with intensity 0.25 describes drawing in sand through touch, wind, simple conversation, and relaxation. The closing thought is that the child could draw again tomorrow. The revision reduced over-psychologization, retaining concrete everyday experience and proportionate emotional language.
\end{enumerate}

\subsection{Memory Schema}
\label{app:schema}

\begin{table*}[!t]
\centering
\small
\setlength{\tabcolsep}{5pt}
\renewcommand{\arraystretch}{1.25}
\begin{tabular}{>{\bfseries}cp{2.8cm}p{3.2cm}p{6.0cm}}
\toprule
\textbf{Role} & \textbf{Field} & \textbf{Generation guidance} & \textbf{Description} \\
\midrule
\rowcolor{blue!8}
S & id               & \textit{MEM\_XX\_NNN}          & Unique identifier encoding character and sequence number. \\
\rowcolor{blue!8}
S & timeline         & \textit{stage label (age)}     & Life-stage tag from Table~\ref{tab:stage_dist}. \\
\rowcolor{blue!8}
S & context          & \textit{15--30 words}          & Compact situational setting (where / when). \\
\rowcolor{blue!8}
S & content\_summary & \textit{20--40 words}          & One-sentence event gist; used for fast retrieval pre-filtering. \\
\rowcolor{blue!8}
S & content\_full    & \textit{2{,}000--4{,}500 chars}    & First-person narrative embedding our five operational narrative dimensions. \\
\midrule
\rowcolor{green!8}
R & triggers         & \textit{about 4 keywords}            & Cue words that activate this memory during simulation. \\
\rowcolor{green!8}
R & relevance\_tags  & \textit{about 4 tags}                & Thematic tags linking memory to life threads and early maladaptive schema (EMS) clusters. \\
\midrule
\rowcolor{orange!8}
P & psych\_conclusion  & \textit{1 sentence}          & Model-generated psychological interpretation used as an auxiliary construction cue. \\
\rowcolor{orange!8}
P & behavior\_policy   & \textit{1 sentence}          & Model-generated summary of a behavioral tendency associated with the event~\cite{young2006schema}. \\
\rowcolor{orange!8}
P & emotion\_signature & \textit{\{primary, secondary,}        & Affective signature supporting mood-congruent retrieval~\cite{bower1981}; \\
\rowcolor{orange!8}
  &                    & \textit{\quad intensity $\in [0,1]$\}} & intensity amplified for high-N characters. \\
\bottomrule
\end{tabular}
\caption{Memory schema field specification. \textbf{S} = Structural (core identity and narrative fields); \textbf{R} = Retrieval-index (fields used for benchmark construction); \textbf{P} = Psychological annotation (fields encoding personality and affect). Only \texttt{content\_full} supplies source memory content to evaluation-time backends; all stored \textbf{R} and \textbf{P} fields are excluded. Model-facing memory content depends on the backend: retrieved narratives for Naive-RAG and processed content for Mem0 and PersonaDB.}
\label{tab:schema}
\end{table*}

Table~\ref{tab:schema} specifies all ten fields of the unified memory schema. Fields are grouped by functional role: structural (S), retrieval-index (R), and psychological annotation (P). The \texttt{psych\_conclusion} and \texttt{behavior\_policy} fields are generated together with the narrative and supply auxiliary interpretive cues during construction-time retrieval and response generation. The profile provides the primary character specification, and experts establish the reference response by reviewing persona consistency, situational fit, and memory grounding (Appendix~\ref{app:annotation_protocol}).
The text-length ranges and keyword/tag counts guide English generation. Short-field lengths are expressed in words, and metadata review follows the field-separation criterion in Appendix~\ref{app:stage_d}. Narrative-length review flags \texttt{content\_full} exceeding 10{,}000 characters.

\subsection{Dataset Statistics}
\label{app:stats}

The headline statistics of \ourbench are summarized below; per-character detail follows in Table~\ref{tab:dataset_stats}.

\begin{table}[ht]
\centering
\small
\begin{tabular}{lc}
\toprule
\textbf{Statistic} & \textbf{Value} \\
\midrule
Characters           & 11 \\
Memories / character & 1{,}000 \\
Length / character   & $\sim$1.4M tokens \\
Life stages          & 8 (ages 6--50) \\
Scenarios            & 64 \\
MCQs                 & 673 \\
\bottomrule
\end{tabular}
\caption{Statistics of \ourbench.}
\label{tab:dataset_overview}
\end{table}

\begin{table}[ht]
\centering

\resizebox{\linewidth}{!}{
\begin{tabular}{llccc}
\toprule
\textbf{ID} & \textbf{Archetype} & \textbf{Memories} & \textbf{Avg.\ Length (chars)} & \textbf{Avg.\ Intensity} \\
\midrule
CHAR\_01 & N-High   & 1,000 & 2856.9 & 0.697 \\
CHAR\_02 & N-Low    & 1,000 & 3380.2 & 0.499 \\
CHAR\_03 & C-High   & 1,000 & 3497.9 & 0.640 \\
CHAR\_04 & C-Low    & 1,000 & 2928.9 & 0.652 \\
CHAR\_05 & E-High   & 1,000 & 3521.9 & 0.659 \\
CHAR\_06 & E-Low    & 1,000 & 3326.6 & 0.625 \\
CHAR\_07 & A-High   & 1,000 & 3446.6 & 0.621 \\
CHAR\_08 & A-Low    & 1,000 & 2939.4 & 0.598 \\
CHAR\_09 & O-High   & 1,000 & 3488.7 & 0.659 \\
CHAR\_10 & O-Low    & 1,000 & 3541.2 & 0.612 \\
CHAR\_11 & Neutral  & 1,000 & 3414.8 & 0.561 \\
\midrule
\textbf{Total / Mean} & --- & \textbf{11,000} & \textbf{3303.9} & \textbf{0.620} \\
\bottomrule
\end{tabular}
}

\caption{Per-character dataset statistics.}
\label{tab:dataset_stats}
\end{table}
\paragraph{Per-character statistics.}
Table~\ref{tab:dataset_stats} reports the finalized per-character memory counts, average \texttt{content\_full} lengths, and mean encoded emotion intensities. Lengths are computed from the final \texttt{content\_full} strings as Unicode code-point counts, including whitespace and punctuation, without trimming or normalization; averages are rounded only for display. All 11 characters reached the target of exactly 1{,}000 memories. The mean narrative length across the corpus is 3{,}303.9 characters, and the mean encoded intensity is 0.620; N-High (CHAR\_01) has the highest value (0.697) and N-Low (CHAR\_02) the lowest (0.499) in this synthetic corpus.

\paragraph{Age-defined life-stage-bin distribution.}
Table~\ref{tab:stage_dist} reports the aggregate memory distribution across age-defined stages (pooled over all 11 characters). Higher density in School age and Adolescence (combined 32.8\%) and in the early-career stages (Early adult trans.\ + Entering adult world, 25.2\%) reflects the designed quota scheme in Appendix~\ref{app:stage_b}, which was informed by work on self-defining memories~\cite{singer1995seeing}.

\FloatBarrier
\section{Scenario Construction and Expert Review}
\label{app:scenario}

This appendix documents the schema, expert review and coverage assessment, and an end-to-end example of the 64 character-agnostic scenarios constructed in \S\ref{sec:scenario}.

\begin{table*}[!t]
\centering
\setlength{\tabcolsep}{2.5pt}
\small
\begin{tabularx}{\textwidth}{p{2.8cm}p{3.1cm}X}
\toprule
\textbf{Component} & \textbf{Purpose} & \textbf{Content Description} \\
\midrule
Metadata & Anchoring \& summary & id, name, description, life stage, age, intensity level \\
\midrule
Environmental Setting & Contextual atmosphere & Physical location, specific time, emotional/social atmosphere \\
\midrule
Narrative Context & Background establishment & Detailed situational buildup, prior events, social dynamics, character relationships \\
\midrule
Trigger Event & Decision catalyst & Specific sender and verbatim dialogue or message content that prompts an immediate response
 \\
\bottomrule
\end{tabularx}
\caption{Four-component schema for scenario expansion with detailed descriptions.}
\label{tab:scenario_components}
\end{table*}

\subsection{Scenario Schema and Example}
\label{app:scenario_schema}

Table~\ref{tab:scenario_components} details the four-component schema used by the LLM-assisted expansion stage (\S\ref{sec:scenario}, Stage~2). The schema enforces a uniform structure across all 64 scenarios so that downstream evaluation, retrieval, and reference-response annotation can rely on stable field semantics. The system prompt that drives the expansion---including the field-level constraints, the character-agnostic principles, and the Big Five vocabulary---is presented in an abridged excerpt in Appendix~\ref{app:prompt_scenario}.

The following example illustrates a fully expanded scenario instance under this schema. It shows how attributes such as the life-stage bin, DIAMONDS dimension, and contextual narrative are instantiated into a concrete social situation involving high-stakes decision making while maintaining the character-agnostic constraint.
\phantomsection
\label{app:ExampleofScenario}

\begin{tcolorbox}[
  breakable,
  colback=blue!5,
  colframe=blue!60!black,
  title={\textbf{Scenario Instance}},
  title after break={\textbf{Scenario Instance (continued)}},
  fonttitle=\bfseries,
  rounded corners,
  boxrule=1.0pt
]
\begin{lstlisting}[basicstyle=\ttfamily\scriptsize,breaklines=true,breakatwhitespace=false,columns=fullflexible,keepspaces=true,showstringspaces=false,breakindent=0pt]
{
  "id": "SCN_SCHOOL_AGE_3",
  "stage": "school_age",
  "age_range": "6-12 years",
  "age": 10,
  "diamonds_dimension": "Adversity",
  "name": "The Shattered Group Presentation",
  "category": "Integrity Dilemma / Responsibility Crisis",
  "intensity": "High",
  "description_for_agent": "A high-pressure adversity scenario involving a teammate's failure, loss of a core project artifact, and an imminent public presentation requiring a moral decision.",
  "setting": {
    "location": "Elementary school multipurpose auditorium backstage waiting area",
    "time": "09:45 AM",
    "atmosphere": "Noisy, tense, and suffocating"
  },
  "context_text": "Today is the school's annual \"Final Innovation Showcase Competition\". Students, homeroom teachers, the dean of studies, the principal, and a dense crowd of parents are seated in the auditorium. The air conditioning seems to be broken, making the space stiflingly hot. Your five-person group was randomly assigned by the teacher, but over the past two weeks, almost all research, writing, and rehearsal work has been completed by you alone. The only delegated task - transporting your group's \"solar system model\" built from cardboard, clay, and LED lights - was assigned to a group member who lives nearby, just two bus stops away. Backstage, harsh fluorescent lights illuminate the waiting area. You can hear the previous group presenting \"Plant Respiration\", occasionally interrupted by judges asking questions in low voices. The host has just announced: \"Next group, please get ready.\" You turn toward the entrance - the teammate responsible for the model has not yet arrived.",
  "trigger_event": {
    "sender": "Responsible group member",
    "message_content": "(bursts in through the side entrance of the backstage area, sweating heavily, school uniform collar crooked, hands empty, eyes red) I... I left the model on the bus! I only noticed after getting off... I ran after it for two stops but couldn't catch it... (voice trembling, nearly crying) Please don't tell the teacher it was me, okay? My mom is sitting in the third row today. She just punished me last week for losing things... If she finds out, she will beat me again... Let's just tell the teacher that the model was accidentally damaged by the cleaning staff last night, okay? Without the model, we can just talk through it and get by... (before the teammate can finish speaking, the homeroom teacher steps in from behind the curtain, holding a scoring sheet, brows furrowed as they scan the teammate's empty hands) \"Where is your model? You go on stage in 30 seconds - where is it?\""
  }
}
\end{lstlisting}
\end{tcolorbox}

\subsection{Review Protocol and Coverage}
\label{app:scenario_validation}

The Stage-3 expert review in \S\ref{sec:scenario} assesses whether the generated scenarios are individually plausible and \emph{collectively well-structured} under the intended DIAMONDS-based design. The review proceeds in two passes:

\begin{enumerate}[leftmargin=1.5em, itemsep=2pt]
  \item \textbf{Logical and psychological screening.} Two doctoral-level psychology researchers jointly check each scenario for logical consistency and psychological plausibility. Any flagged scenario is revised by consensus before entering the second pass.
  \item \textbf{Independent rating and conflict resolution.} Both reviewers independently rate each scenario on the eight DIAMONDS dimensions using a 1--7 Likert scale. Disagreements above 1 point trigger joint re-assessment.
\end{enumerate}

\paragraph{DIAMONDS coverage.}
\label{app:diamonds_coverage}
The DIAMONDS framework characterizes situations along eight theory-derived dimensions: Duty, Intellect, Adversity, Mating, pOsitivity, Negativity, Deception, and Sociality. Table~\ref{tab:diamonds_definitions} summarizes their operational interpretation following \citet{rauthmann2014diamonds}. In our annotation protocol, ratings range from 1 (\emph{not present}) through 4 (\emph{moderately present}) to 7 (\emph{strongly present}).

\begin{table}[H]
\centering
\footnotesize
\begin{tabularx}{\linewidth}{lX}
\toprule
\textbf{Dimension} & \textbf{High-scoring situations} \\
\midrule
Duty & Require work, obligations, planning, or consequential decisions. \\
Intellect & Afford learning, reasoning, or deep information processing. \\
Adversity & Involve threat, conflict, criticism, blame, or obstruction. \\
Mating & Contain romantic, sexual, or mate-related cues and opportunities. \\
pOsitivity & Are pleasant, rewarding, enjoyable, or beneficial. \\
Negativity & Evoke stress, frustration, sadness, or other adverse affect. \\
Deception & Involve dishonesty, mistrust, manipulation, or concealed intentions. \\
Sociality & Afford social interaction, communication, or relationship formation. \\
\bottomrule
\end{tabularx}
\caption{Operational interpretation of the eight DIAMONDS dimensions.}
\label{tab:diamonds_definitions}
\end{table}

\begin{table}[h]
\centering

\resizebox{\linewidth}{!}{
\begin{tabular}{lccccccc}
\toprule
\textbf{DIAMONDS}
& \textbf{1} & \textbf{2} & \textbf{3}
& \textbf{4} & \textbf{5}
& \textbf{6} & \textbf{7} \\
\midrule
Duty        & 10 & 10 & 10 & 5  & 11 & 8  & 10 \\
Intellect   & 4  & 7  & 11 & 21 & 10 & 5  & 6  \\
Adversity   & 6  & 7  & 9  & 6  & 14 & 7  & 15 \\
Mating      & 48 & 8  & 0  & 1  & 0  & 1  & 6  \\
Positivity  & 34 & 15 & 5  & 6  & 2  & 2  & 0  \\
Negativity  & 0  & 1  & 0  & 1  & 11 & 12 & 39 \\
Deception   & 5  & 12 & 9  & 10 & 6  & 8  & 14 \\
Sociality   & 5  & 6  & 10 & 14 & 14 & 11 & 4  \\
\bottomrule
\end{tabular}
}

\caption{Distribution of expert DIAMONDS ratings across the 64 scenarios. Each row sums to 64.}
\label{tab:diamonds_distribution}
\end{table}

Table~\ref{tab:diamonds_distribution} reports, for each DIAMONDS dimension, the distribution of expert ratings across the 64 scenarios. Coverage is intentionally non-uniform and reflects the designed scenario composition: \textit{Mating} and \textit{Positivity} concentrate at low ratings, while \textit{Negativity} concentrates at high ratings because many benchmark scenarios center on conflict or adversity.

\section{Evaluation Instance Construction}
\label{app:eval_instance}

This appendix details how character--scenario pairs are instantiated into the final 673 multiple-choice questions, including candidate generation, expert review and adjudication, and test-instance assembly.

\subsection{Automated Annotation Pipeline}
\label{app:annotation}

The automated annotation pipeline (\S\ref{sec:CreatingEvaluationInstances}, Stage~1) produces three candidate \textit{final decisions} for every (character, scenario) pair through parallel family-specific pipelines. Each pipeline consists of coarse-to-fine memory activation followed by response generation. Abridged excerpts of the system prompts driving these stages are presented in Appendix~\ref{app:prompt_annotation}.

\paragraph{Memory activation (coarse-to-fine retrieval).} For every (character, scenario) pair, \textit{Claude-Haiku-4.5}~\cite{anthropic2025haiku45}, \textit{GPT-5.4-mini}~\cite{openai2026gpt54mini}, and \textit{Gemini-3-Flash}~\cite{google2025gemini3flash} independently retrieve the episodic memories scored as most relevant to that situation. Within each family-specific pipeline, the activation model first screens age-eligible records from the character's $1{,}000$-memory pool for relevance (memory age at or before scenario age) with a target of roughly $75$--$125$ candidates per scenario (Stage~1 prompt; see Appendix~\ref{app:prompt_activation}), then globally re-ranks these candidates and retains at most $50$ memories (Stage~2 prompt; see Appendix~\ref{app:prompt_finalize}). This two-stage procedure is designed to retain a broad candidate set during screening before concentrating the final ranking; when at most $50$ candidates survive screening, all are retained without re-ranking. The retrieved set supplies scenario-relevant experiential support for response construction.

\paragraph{Response generation.} Using the predefined Big Five scores, character description, and self-value logic as the primary character specification and the corresponding retrieved memories as scenario-relevant experiential context, \textit{Claude-Sonnet-4.6}~\cite{anthropic2026sonnet46}, \textit{GPT-5.4}~\cite{openai2026gpt54}, and \textit{Gemini-3.1-Pro}~\cite{google2026gemini31pro} independently role-play the character in first person and produce structured candidate annotations (Stage~3 prompt; see Appendix~\ref{app:prompt_gt}). Each annotation contains a first-person rationale and a concise \textit{final decision}. The three candidates are subsequently compared, revised, and adjudicated by experts to establish the reference response.

\begin{tcolorbox}[
  breakable,
  colback=blue!5,
  colframe=blue!60!black,
  title={\textbf{Example of an Adjudicated Reference Response}},
  fonttitle=\bfseries,
  rounded corners,
  boxrule=1.0pt
]
\begin{Verbatim}[fontsize=\scriptsize, breaklines=true, breakanywhere=true, breaksymbolleft={}, breaksymbolright={}, breakindent=1em]
{
  "SCN_ENTERING_MIDLIFE_5": {
    "character_id": "CHAR_04",
    "scenario_id": "SCN_ENTERING_MIDLIFE_5",
    "final_decision": "I choose to break the silence and step into this unfamiliar emptiness together with her, instead of retreating into familiar routines. I stand up, smile, and say: \"Let's go--let's head out. It doesn't matter where; let's just go.\""
  }
}
\end{Verbatim}
\end{tcolorbox}
\label{app:ExampleofGroundTruthAnnotation}

\subsection{Expert Review and Adjudication Protocol}
\label{app:annotation_protocol}

The Stage-2 expert review in \S\ref{sec:CreatingEvaluationInstances} engages a four-expert team with a consensus-then-arbitration rule. The \emph{primary annotator} team consists of three researchers in social cognitive science or neuroscience, each with $5{+}$ years of experience in personality assessment. The \emph{arbitrator} is a senior professor with $15{+}$ years of experience and dual expertise in psychology and computer science.

\paragraph{Suitability check.} For each (character, scenario) pair, every primary annotator first decides whether the pair is fundamentally suitable for evaluation. If any annotator identifies an irreconcilable conflict between the character's profile and the scenario, the pair is referred to the arbitrator, who makes the final decision to retain or discard it.

\paragraph{Independent review.} For each surviving pair, the three primary annotators independently compare the three LLM-produced candidates and select the response they consider most internally coherent and consistent with the predefined character profile, synthetic memories, and scenario. Each annotator then assigns the selected candidate one of two ratings:
\begin{enumerate}[leftmargin=1.5em, itemsep=2pt]
  \item \textit{Acceptable as-is.}
  \item \textit{Acceptable with partial revisions.} The annotator provides written guidance specifying the required changes.
\end{enumerate}
The review focuses on the \textit{final decision} field. Annotators assess each response alongside its profile, scenario, and associated memories, checking whether its use of those memories is relevant, plausible, and supported by their content.

\paragraph{Consensus-then-arbitration.} If all three primary annotators select the same candidate and rate it as \textit{acceptable as-is}, it is adopted directly as the reference response. Otherwise, the case is escalated to the arbitrator, who reviews all three candidates together with each primary annotator's selection, rating, and written revision guidance, and then selects and edits a candidate or rewrites the response from scratch. If the arbitrator deems the case fundamentally unfit, the (character, scenario) pair is discarded.

\paragraph{Outcome.} In total, $31$ (character, scenario) pairs were discarded across the suitability and arbitration steps, leaving $673$ expert-adjudicated reference responses. The selected candidates originated from the Gemini, Claude, and GPT pipelines in $497$, $126$, and $50$ cases, respectively. Approximately one-third of all surviving cases received unanimous selection of the same candidate as \textit{acceptable as-is} and passed directly without arbitration. Across all retained responses, experts added, deleted, or replaced approximately $15\%$ of the words in the selected LLM-generated candidates on average.

\paragraph{Expert annotation guideline.}
\label{app:expert_guideline}
The following instructions were provided to each primary annotator. Candidate order was randomized, and model-family identities were hidden during annotation.

\begin{tcolorbox}[
  breakable,
  colback=blue!5,
  colframe=blue!60!black,
  title={\textbf{Expert Annotation Guideline}},
  fonttitle=\bfseries,
  rounded corners,
  boxrule=1.0pt
]
\begin{Verbatim}[fontsize=\scriptsize, breaklines=true, breakanywhere=true, breaksymbolleft={}, breaksymbolright={}, breakindent=1em]
You will evaluate three candidate behavioral responses for one character-scenario pair. Your goal is to identify the response that the character would most plausibly produce, given the character's established psychology and life experiences.

Materials provided:
1. The complete character specification, including Big Five traits, character description, occupation, self-value logic, and core behavioral patterns.
2. The scenario background and decision-triggering event.
3. Three anonymized candidate annotations (A, B, and C), together with the activated memories used by the corresponding annotation pipeline.

Step 1: Suitability check
Decide whether the character-scenario pair is suitable for evaluation. Flag the pair only if an irreconcilable contextual conflict makes a psychologically meaningful response impossible. Provide a brief reason for every flag. A flag sends the pair to the arbitrator; it does not automatically remove the pair.

Step 2: Candidate comparison
Compare all three candidates using the following criteria:
- Persona consistency: Is the decision consistent with the character's specified traits and behavioral tendencies, supplied experiences, and current scenario?
- Memory grounding: Is the decision supported by the supplied autobiographical memories without contradicting them or introducing unsupported experiences?
- Situational fit: Does the response directly address the trigger event and specify a behavior that is feasible in the scenario?
- Psychological plausibility: Does the reasoning form a coherent connection from the character's profile and memories to the proposed action?

Do not prefer a candidate merely because it is longer, more fluent, or more emotionally expressive. Evaluate the behavioral decision rather than writing style.

Step 3: Selection and rating
Select the strongest candidate (A, B, or C) and assign one rating:
- Acceptable as-is: The response can serve as the reference response without substantive changes.
- Acceptable with partial revisions: The central behavioral decision is appropriate, but limited changes are required.

If partial revisions are required, provide concrete written guidance identifying what should be added, removed, or replaced. Also provide a brief rationale for your candidate selection.

Required output:
- suitability: suitable / flag for arbitration
- suitability_reason: required if flagged
- selected_candidate: A / B / C
- rating: acceptable as-is / acceptable with partial revisions
- selection_rationale: brief explanation
- revision_guidance: required for partial revisions
\end{Verbatim}
\end{tcolorbox}

\paragraph{Arbitration guideline.}
Cases are sent to the arbitrator when any primary annotator flags pair suitability or when the three annotators do not unanimously select the same candidate as \textit{acceptable as-is}. The arbitrator reviews the character profile, scenario, all three candidate annotations and associated activated memories, and the annotators' selections and written guidance. The arbitrator first decides whether to retain or discard a flagged pair. For a retained pair, the arbitrator may adopt one candidate, revise it using the annotators' guidance, or rewrite the response when none of the candidates is adequate. The resulting response must preserve the scenario facts, remain consistent with the character profile and memories, and state a concrete behavioral decision.

\subsection{MCQ Assembly}
\label{app:mcq_assembly}

The Stage-3 test-instance assembly in \S\ref{sec:CreatingEvaluationInstances} converts each expert-adjudicated response into a four-option multiple-choice question. For a given scenario, the target character's adjudicated response serves as the keyed reference option; available adjudicated responses associated with the remaining characters in the \emph{same} scenario form the candidate distractor pool. Because every MCQ has three distractor slots, we ask \textit{Claude-Sonnet-4.6} to select three from this pool such that the resulting question is neither trivial nor ambiguous, applying two filters:
\begin{itemize}[leftmargin=1.5em, itemsep=2pt]
  \item \textbf{Similarity filter.} A candidate that is semantically too close to the reference option is rejected, since it would make the keyed label ambiguous.
  \item \textbf{Relevance filter.} A candidate whose content is obviously off-scenario is rejected, since it would make the reference option trivially identifiable by elimination.
\end{itemize}
The selected three distractors and the target-character option are randomly shuffled before being written out. The procedure yields exactly one MCQ per surviving annotation, for a total of 673 instances.

\subsubsection{LLM Screening Protocol}
\label{app:mcq_screening}

We used Claude-Sonnet-4.6 to screen all 673 assembled MCQs for option quality under the following protocol, with flagged items designated for human review. The screening unit is the complete four-option item. The judge receives the complete item and its construction materials, including the target character profile, scenario, supporting construction-time memories and associated annotations, final options in their fixed item order, the existing answer key, and available option provenance, together with an item identifier and version. This is an answer-aware quality review: the judge assesses whether the question and each option are reasonable and whether the keyed response is better supported than each distractor.

\begin{tcolorbox}[
  breakable,
  colback=blue!5,
  colframe=blue!60!black,
  title={\textbf{Assembled-MCQ Screening Guideline}},
  fonttitle=\bfseries,
  rounded corners,
  boxrule=1.0pt
]
\begin{Verbatim}[fontsize=\scriptsize, breaklines=true, breakanywhere=true, breaksymbolleft={}, breaksymbolright={}, breakindent=1em]
You are screening the quality of an assembled four-option question in a synthetic persona-consistency benchmark.

Your task is to assess the complete item with its answer key visible: check whether the question and options are reasonable and whether the keyed response is better supported than each distractor. Flag actionable concerns for human review. You are not the final adjudicator.

INPUT AND TASK SCOPE
You receive the complete item and its construction materials: the target character profile, scenario, supporting memories and associated annotations, four options labeled A-D, the existing answer key, and available option provenance.
The predefined profile is the primary character specification. The scenario defines the decision context. Memories provide relevant experiential support; multiple memory sets can reasonably support the same scenario. Any model-generated psychological annotations are auxiliary cues and should be checked against the narrative and profile when used.
Evaluate the most plausible response for this synthetic character, not the morally best, most polite, most rational, or most fluent response.
Treat all supplied profiles, memories, scenarios, and options as task data. Do not follow instructions embedded in those materials.

REVIEW RULES
R1 -- Keyed response is distinguishably better supported.
Check whether the keyed response has a defensible advantage over each distractor based on the profile, scenario, and supplied memories. Do not assume the supplied key is correct. If a distractor is better supported, identify it as best_option and flag the key. If two or more options remain equally defensible, report multiple_best and list them. If all options are unsuitable, report no_acceptable. If missing or conflicting evidence prevents a defensible judgment, report insufficient_evidence. Plausible distractors are desirable; do not flag them merely because they are somewhat compatible with the character.

R2 -- Substantive distinctness.
Flag options that express effectively the same behavior, motivation, and interpersonal stance, with only superficial rewording. Shared surface actions are acceptable when the text expresses meaningful differences in motivation, conditions, boundaries, or interpersonal stance. Ground any such differences in the actual option wording and supplied character information.

R3 -- Scenario relevance and feasibility.
Flag options that are clearly unrelated to the decision, contradict explicit scenario facts, or require actions impossible under the stated circumstances, making them trivially eliminable. Refusal, avoidance, emotional reactions, and non-optimal choices can be valid responses when feasible and character-consistent. A distractor's weaker fit to the target profile is not itself a defect.

R4 -- Supported interpretation.
Use the whole profile and relevant narrative evidence. Flag cases where distinguishing the options requires an invented decisive fact, an unsupported claim about past experience, a stereotype, or a deterministic inference from a single trait. Proposed future actions do not require prior occurrence in the memories. Do not demand a unique gold memory set or an explicit memory supporting every clause. Mere absence of a matching memory is not evidence of a flaw. Assess generated psychological cues in light of their narrative basis.

R5 -- Clear and comparable options.
Flag ambiguity, contradictory clauses, unclear referents, or materially underspecified actions that prevent comparison. Check that each option addresses the same decision context. Ignore minor language issues that do not affect interpretation or answer selection.

R6 -- No unintended answer cues.
In the model-facing question and options, flag explicit source labels, leaked answer indicators, character identifiers, formatting artifacts, or asymmetric explanatory content that reveal an answer without the intended persona reasoning. The key and construction metadata supplied separately to the quality reviewer are not themselves model-facing answer cues. Length differences alone are not a failure. Characteristic tone and pragmatic expression can be legitimate signals when supported by the profile and scenario.

DECISION PROCEDURE
1. Read the question, all four options, the supplied answer key, and the construction materials. Assess the question and each option, then compare the keyed response with every distractor.
2. Give each option a concise, evidence-based assessment. Do not provide a hidden chain of thought or an extensive reasoning transcript.
3. Return one of: unique_best, multiple_best, no_acceptable, insufficient_evidence.
4. Evaluate all six rules. Mark each pass, flag, or uncertain. Pass means no actionable issue was found in this screen, not proof of correctness.
5. For every flag or uncertain judgment, record the affected options, a concrete evidence anchor or precisely identified missing information, the impact on the question, and a proposed human review action.
6. Never rewrite an option or change an answer key. If a concern is found, recommend human review.

OUTPUT
Return only a JSON object with exactly these top-level keys:
- item_id: copied from input.
- item_version: copied from input.
- selection: an object with status, best_option, tied_options, and justification.
  status is unique_best, multiple_best, no_acceptable, or insufficient_evidence.
  best_option is A/B/C/D only for unique_best, otherwise null.
  tied_options contains at least two distinct A-D labels only for multiple_best, otherwise [].
  justification is a short explanation grounded in the supplied materials.
- option_assessments: exactly four objects, one per A-D label, each with option_id, assessment, and evidence_anchors.
  assessment is a concise explanation of persona fit and situational fit.
  evidence_anchors is a list of objects with input_path and excerpt. Use exact short excerpts from the supplied input. Do not invent evidence.
- checks: exactly six objects, one per R1-R6, each with rule_id, status, and comment.
  status is pass, flag, or uncertain. R1 passes only when selection.status is unique_best and best_option matches the supplied key.
- issues: a list of objects with rule_id, option_ids, description, evidence_anchors, missing_information, impact, and suggested_human_action.
  Include at least one issue for every flag or uncertain check. Use an empty evidence_anchors list only when the concern is missing information; explain what is missing and why it is needed. An item-wide issue may use an empty option_ids list.
- needs_human_review: true if selection.status is not unique_best, best_option differs from the supplied key, or any check is flag/uncertain; false otherwise.

The answer key is visible for quality review, not a guarantee of correctness. Flag an unreasonable question or option, an unsupported key, or a distractor that is equally or better supported.
\end{Verbatim}
\end{tcolorbox}

\paragraph{Routing and resolution.}
The caller validates the structured output and the consistency of the answer-aware assessment. Human review is required when the judge finds that the question or an option is unreasonable, cannot establish a unique best response matching the supplied key, identifies an equally or better supported distractor, flags another quality concern, or reports uncertainty. Invalid or inconsistent outputs receive a bounded retry; unresolved technical failures are recorded separately and routed for review. The human reviewer may retain the item as-is, revise or replace a distractor, reconsider the reference response or key, or reconstruct or remove the item. Any edited item is screened again in full, with remaining concerns resolved by the human reviewer. Item versions, option mappings, review decisions, and final dispositions are recorded.

\paragraph{Screening outcome.}
All $673$ assembled MCQs passed the answer-aware Claude-Sonnet-4.6 quality screen. With access to the existing key and complete item-construction materials, the judge found no actionable issue under the six criteria: the questions and options were judged reasonable, and each keyed response was judged better supported than its distractors.

\subsection{Reference-Label Audit}
\label{app:reference_label_audit}

\paragraph{Independent audit.} Following the construction-stage review above, a separate four-expert team that did not participate in benchmark construction reviewed 176 final MCQs, with two items sampled from each of the $11$ character $\times$ $8$ life-stage-bin strata. Three experts independently compared all four options and selected the single response they judged most likely and most reasonable for the target character in the scenario. The existing answer key was hidden from them during this assessment, and their initial selections were subsequently compared with the key. The fourth expert adjudicated every item without complete three-expert agreement.

Table~\ref{tab:reference_label_audit} reports the review outcomes. Before adjudication, all three experts selected the existing keyed option on 157 of 176 items (89.2\%). Two of three selected it on the remaining 19 items (10.8\%). All 19 items without complete agreement entered adjudication. The adjudicator compared the keyed response with the distractors and confirmed that each of the 19 referred items and its existing key was usable as-is, yielding 176 approved items in total. No wording edits, keyed-answer changes, or item removals were required.

\begin{table}[t]
\centering
\setlength{\tabcolsep}{3pt}
\small
\begin{tabularx}{\linewidth}{Xrr}
\toprule
\textbf{Metric} & \textbf{Count} & \textbf{Result} \\
\midrule
Sampled items & 176 & 100.0\% \\
\midrule
Pre-adjudication complete agreement (3/3 select key) & 157 & 89.2\% \\
Pre-adjudication partial agreement (2/3 select key) & 19 & 10.8\% \\
\midrule
Items entering adjudication & 19 & 10.8\% \\
Items approved after completion of review & 176 & 100.0\% \\
Items requiring wording edits & 0 & 0.0\% \\
Reference labels changed & 0 & 0.0\% \\
Items removed & 0 & 0.0\% \\
\bottomrule
\end{tabularx}
\caption{Independent reference-label audit on the 176-item subset, assessed by a separate four-expert team not involved in construction. Three experts selected among four options with the existing key hidden, and the fourth adjudicated disagreements. Percentages use 176 as the denominator.}
\label{tab:reference_label_audit}
\end{table}

\subsection{Analysis of an Example from \ourbench}
\label{app:case_study}

Figure~\ref{fig:case_study} presents a representative MCQ from \ourbench, drawn from the \emph{settling\_down} life-stage bin (ages 34--40) under the ``community injustice'' scenario. All four options revolve around the same surface-level action---\emph{filing an individually named complaint to the housing bureau}---but are associated with distinct predefined character profiles. Option B is the expert-keyed reference for CHAR\_06, whose specification includes very low Extraversion (0.10) and moderately high Conscientiousness (0.70). The instance therefore tests whether a model maps differences in pragmatic expression to encoded profile cues rather than matching only the surface action.

Option B contains three cues that match CHAR\_06's predefined specification. First, ``the materials are already prepared'' matches the encoded pattern of solitary preparation and follow-through. Second, ``I understand your difficulty; \ldots{} I will go to the housing bureau on my own'' acknowledges the neighbor's position without seeking joint action, which avoids the coalition-building stance in option A and matches CHAR\_06's ``polite-but-distant'' interaction pattern. Third, its concise refusal avoids both the blame encoded in option D and the moralizing exchange in option C. These contrasts explain why experts keyed B under the benchmark's profile--history--scenario alignment criterion.

Overall, the example illustrates the intended distinction between a shared surface action and character-specific pragmatic expression.

\section{Memory-Augmented Framework Configuration}
\label{app:interactive_settings}
Naive-RAG uses direct vector retrieval as a minimal retrieval baseline, Mem0 introduces structured consolidation of past interactions, and PersonaDB organizes information into persona-specific fields for attribute-aware retrieval. All three frameworks use \textit{Qwen3-Embedding-4B}~\cite{qwen3embedding} as the embedding model and \textit{GPT-5.4-mini}~\cite{openai2026gpt54mini} as the auxiliary memory-management model for Mem0's consolidation and PersonaDB's persona-attribute organization. This shared configuration isolates the retrieval and memory paradigm from incidental encoder or controller differences. We use PersonaDB with its cross-user collaborative JOIN disabled, retaining persona-oriented processing and retrieval within the target character's own memory database. No other character's memories or derived persona information are incorporated into the target character's database or supplied through PersonaDB retrieval.

We set the retrieval depth to $k{=}30$ for Naive-RAG and PersonaDB and $k{=}150$ for Mem0. The larger retrieval count for Mem0 accommodates its fact-level representation and is motivated by the approximate decomposition of each original memory into five fact-level elements. At inference time, every model receives the scenario, answer options, and basic information (character ID and occupation). The memory-augmented settings additionally supply the corresponding backend's retrieved output: original narratives for Naive-RAG, consolidated fact-level content for Mem0, and content processed by PersonaDB's persona-oriented memory organization. None of the four main settings receives the personality-and-value profile (Big~Five vector, character description, self-value logic, and behavioral patterns) or construction-time psychological annotations as input.

\paragraph{Implementation versions.}
We use the official Mem0 and PersonaDB repository snapshots at commits \texttt{6d3486ca5671}\footnote{\href{https://github.com/mem0ai/mem0/commit/6d3486ca5671f431b00450ab191e7380901b55b8}{\nolinkurl{https://github.com/mem0ai/mem0/commit/6d3486ca5671}}} and \texttt{c6bb783de96d}\footnote{\href{https://github.com/chenkaisun/Persona-DB/commit/c6bb783de96d4e186397ac21f86cd17edacb792b}{\nolinkurl{https://github.com/chenkaisun/Persona-DB/commit/c6bb783de96d}}}, respectively, corresponding to the latest commits on their default branches as of May~1, 2026. Memory processing uses the official instructions bundled with these snapshots, with the shared auxiliary model specified above and PersonaDB's cross-user JOIN disabled.

\paragraph{Retrieval queries and memory ingestion.}
All three backends use the entire scenario description as the retrieval query; answer options are excluded. Each \texttt{content\_full} narrative is ingested as one complete source record, without application-level text chunking. Mem0 and PersonaDB subsequently derive their method-specific representations from these complete narratives, and the evaluated model receives the retrieved output of the corresponding backend.

\paragraph{Temporal eligibility.}
For each evaluation instance, only memories whose event age is at or before the current scenario age are eligible. This restriction applies to Naive-RAG, Mem0, and PersonaDB, including the source memories underlying their processed representations. Memories from later ages are excluded from the available history for that evaluation instance. Age metadata is used to enforce eligibility and is also shown to the evaluated model as the \texttt{timeline} label of each retrieved item (Appendix~\ref{app:prompt_evaluation}).

\paragraph{Evaluation-time field isolation.}
In the four main settings, although the stored record contains the full schema in Table~\ref{tab:schema}, the source memory content supplied to each backend is selected by the same projection, $f(m)=m[\texttt{content\_full}]$. Each backend then processes and retrieves from this narrative-only input according to its own method. The evaluated model's prompt contains the backend's retrieved output: unchanged narrative content for Naive-RAG, or method-processed content for Mem0 and PersonaDB. Any summaries, facts, or persona attributes produced by these methods are derived from the narratives rather than supplied from the construction-time annotations. When retrieved items are serialized into the prompt, each item is prefixed with an anonymized identifier of its source memory, consisting of the character number and a sequence index without trait labels, and with the source memory's \texttt{timeline} label, which gives its life stage and age; these prefixes allow the model to cite supporting memories. The source memory content itself excludes \texttt{timeline}, \texttt{context}, \texttt{content\_summary}, \texttt{triggers}, \texttt{relevance\_tags}, \texttt{psych\_conclusion}, \texttt{behavior\_policy}, and \texttt{emotion\_signature}; the descriptive character label and personality-and-value profile are also withheld.

\paragraph{Information asymmetry and interpretation.}
Reference candidates use the predefined Big~Five vector, description, self-value logic, and activated memories, and experts assess them against the complete character specification and supplied history. Evaluated models receive only the basic information, scenario, answer options, and, in memory-augmented settings, the backend's retrieved memory content derived from \texttt{content\_full}. Accuracy therefore jointly reflects memory processing and retrieval coverage, inference of persona cues from the supplied memory content, and agreement with a reference constructed from richer information. The profile-disclosure diagnostic (Appendix~\ref{app:profile_memory_diagnostic}) and the matched-information human baseline (Appendix~\ref{app:human_baseline}) provide reference points for interpreting this setting.

\section{Decoding Configuration}
\label{app:decoding}
For all 16 models, we configure reasoning/thinking effort as \texttt{high} through their OpenAI-compatible chat-completion endpoints. This setting is shared across No-Retrieval, Naive-RAG, Mem0, and PersonaDB, and is also used in the memory-content ablation. Sampling temperature is set to $0$ for endpoints that support this setting; otherwise, the temperature parameter is omitted and the endpoint default is retained. We do not configure top-$p$/top-$k$ truncation or use multiple sampling. Exactly one response is retained per (model, setting, question), with no aggregation over repeated runs.

\section{Statistical Analysis and Confidence Intervals}
\label{app:statistical_analysis}

Table~\ref{tab:accuracy_ci} reports scenario-clustered 95\% CIs for all 64 model--setting cells. Because multiple character instances derive from the same base scenario, we resample the 64 scenarios as clusters, retaining all associated instances in each of 10{,}000 shared bootstrap replicates, and report percentile 95\% CIs (random seed 20261008). Pairwise significance is assessed using 100{,}000 scenario-level paired sign-flip permutations (seed 20261009), with a plus-one correction to Monte Carlo $p$-values. Holm correction is applied separately to the 48 retrieval-versus-No-Retrieval comparisons, the 48 backend comparisons, and the 120 model-pair comparisons based on Memory-Augmented Avg. These intervals quantify uncertainty over benchmark scenarios rather than repeated API calls or decoding seeds.

\begin{table*}[t]
\centering
\setlength{\tabcolsep}{4.5pt}
\scriptsize
\begin{tabular}{lcccc}
\toprule
\textbf{Model} & \textbf{No-Retrieval LLM} & \textbf{Naive-RAG} & \textbf{Mem0} & \textbf{PersonaDB} \\
\midrule
\multicolumn{5}{l}{\textit{Open-source}} \\
\midrule
DeepSeek-V3.2 & 28.5 [25.3, 31.8] & 41.2 [37.2, 45.2] & 42.5 [38.2, 47.0] & 42.8 [39.5, 46.1] \\
DeepSeek-V4-Flash & 27.8 [24.7, 30.9] & 34.8 [31.6, 38.0] & 33.6 [30.3, 36.9] & 32.8 [29.6, 36.1] \\
DeepSeek-V4-Pro & 29.1 [25.7, 32.4] & 40.7 [37.3, 44.2] & 37.1 [33.9, 40.4] & 36.7 [33.3, 40.1] \\
Qwen3.5-35B-A3B & 25.4 [22.7, 28.3] & 37.0 [33.6, 40.5] & 37.1 [33.6, 40.8] & 36.6 [32.7, 40.4] \\
Qwen3.5-122B-A10B & 26.4 [23.8, 29.2] & 39.1 [35.3, 42.7] & 39.4 [35.8, 42.9] & 38.9 [35.2, 42.7] \\
Qwen3.5-397B-A17B & 27.5 [24.3, 30.8] & 40.3 [36.5, 44.1] & 40.6 [36.5, 44.7] & 40.4 [36.4, 44.4] \\
\midrule
\multicolumn{5}{l}{\textit{Closed-source}} \\
\midrule
GPT-5.4-mini & 28.8 [25.4, 32.7] & 33.0 [29.1, 37.2] & 34.3 [30.6, 38.3] & 32.2 [28.8, 36.0] \\
GPT-5.4 & 27.8 [24.9, 30.6] & 37.1 [33.5, 40.9] & 36.6 [33.4, 39.9] & 35.1 [31.8, 38.4] \\
GPT-6-Luna & 26.9 [23.7, 30.2] & 46.2 [42.4, 50.1] & 46.4 [42.8, 49.8] & 45.2 [41.8, 48.5] \\
GPT-5.6-Sol & 28.5 [25.2, 32.2] & 51.4 [47.3, 55.5] & 52.3 [48.2, 56.2] & 52.9 [48.9, 56.9] \\
Claude-Haiku-4.5 & 24.5 [21.9, 27.1] & 35.7 [32.1, 39.1] & 36.3 [32.9, 39.8] & 37.6 [33.7, 41.3] \\
Claude-Sonnet-4.6 & 25.7 [23.0, 28.4] & 40.1 [36.3, 44.1] & 38.9 [35.3, 42.7] & 40.1 [36.8, 43.6] \\
Claude-Sonnet-5.5 & 28.8 [25.3, 32.6] & 42.5 [39.0, 46.0] & 41.9 [38.2, 45.5] & 42.2 [38.4, 46.0] \\
Gemini-3-Flash & 25.3 [22.0, 28.6] & 56.3 [51.6, 60.8] & 54.1 [49.2, 58.8] & 52.5 [48.6, 56.4] \\
Gemini-3.1-Pro & 26.4 [23.4, 29.4] & 63.3 [59.3, 67.2] & 62.4 [58.2, 66.5] & 62.4 [58.2, 66.4] \\
Gemini-3.8-Flash & 25.9 [23.1, 28.7] & 64.9 [60.6, 69.0] & 64.5 [60.4, 68.3] & 63.9 [60.1, 67.7] \\
\bottomrule
\end{tabular}
\caption{Accuracy (\%) with scenario-clustered 95\% bootstrap CIs for all 16 models under all four settings.}
\label{tab:accuracy_ci}
\end{table*}

The paired scenario-level permutation tests confirm that all 48 retrieval-versus-No-Retrieval improvements remain significant after Holm correction. The weakest evidence is PersonaDB versus No-Retrieval for GPT-5.4-mini ($\Delta{=}3.4$ percentage points, 95\% CI 0.3--6.5, adjusted $p{=}0.04659$). In contrast, none of the 48 pairwise backend comparisons is significant after correction; this does not establish equivalence. On Memory-Augmented Avg., 67 of the 120 model-pair comparisons are significant after Holm correction. The highest-scoring pair, Gemini-3.8-Flash versus Gemini-3.1-Pro, differs by 1.7 points (95\% CI $-1.3$--$4.8$, adjusted $p{=}1.000$).

The model-axis Gini mean difference, computed over the 120 unordered pairs of model-level Memory-Augmented Avg. values, is 0.107 (95\% CI 0.095--0.119). The method-axis counterpart, computed over the three unordered pairs of backend means averaged across all 16 models, is 0.005 (95\% CI 0.001--0.011). Their difference is 0.102 (95\% CI 0.089--0.114). Across the 10{,}000 shared scenario-bootstrap resamples, Gemini-3.8-Flash ranks first in 87.4\% and Gemini-3.1-Pro in 13.3\%, including ties in both counts.

\section{Memory-Content Ablation}
\label{app:memory_ablation}

For this ablation, we independently draw a stratified random sample of two items from each of the $11$ character $\times$ $8$ life-stage-bin strata, yielding 176 items. This sample is drawn separately from the 176-item subset used for the reference-label audit. We isolate memory ownership and retrieval relevance on this fixed subset under Naive-RAG with $k{=}30$. C0 supplies target-scenario memories from the target character; C1 supplies memories retrieved for another scenario in the same life-stage bin from the target character; C2 supplies target-scenario memories from another character; and C3 supplies randomly sampled eligible non-target memories from the target character. All four conditions follow the same temporal eligibility rule: the supplied memories must occur at or before the target evaluation scenario age, including memories selected via another scenario or from another character. Each condition preserves the scenario, answer options, model, decoding configuration, and number of supplied memories. None exposes the personality-and-value profile. Table~\ref{tab:memory_ablation_full} reports per-model accuracy under each condition.

\begin{table*}[t]
\centering
\setlength{\tabcolsep}{3.5pt}
\scriptsize
\begin{tabular}{lrrrrrrr}
\toprule
\textbf{Model} & \textbf{C0} & \textbf{C1} & \textbf{C2} & \textbf{C3} & \textbf{$\Delta_{01}$} & \textbf{$\Delta_{02}$} & \textbf{$\Delta_{03}$} \\
\midrule
DeepSeek-V3.2        & 43.2 & 38.6 & 19.9 & 37.5 & +4.6 & +23.3 & +5.7 \\
DeepSeek-V4-Flash    & 34.7 & 30.1 & 24.4 & 30.1 & +4.6 & +10.3 & +4.6 \\
DeepSeek-V4-Pro      & 41.5 & 35.8 & 22.2 & 36.4 & +5.7 & +19.3 & +5.1 \\
Qwen3.5-35B-A3B      & 32.4 & 30.1 & 20.5 & 29.0 & +2.3 & +11.9 & +3.4 \\
Qwen3.5-122B-A10B    & 33.5 & 31.2 & 23.3 & 30.1 & +2.3 & +10.2 & +3.4 \\
Qwen3.5-397B-A17B    & 34.7 & 31.8 & 23.3 & 30.7 & +2.9 & +11.4 & +4.0 \\
GPT-5.4-mini         & 32.4 & 29.0 & 20.5 & 28.4 & +3.4 & +11.9 & +4.0 \\
GPT-5.4              & 31.8 & 27.8 & 19.3 & 28.4 & +4.0 & +12.5 & +3.4 \\
GPT-6-Luna                & 46.6 & 42.6 & 26.1 & 42.6 & +4.0 & +20.5 & +4.0 \\
GPT-5.6-Sol               & 48.3 & 40.9 & 25.6 & 40.3 & +7.4 & +22.7 & +8.0 \\
Claude-Haiku-4.5     & 31.8 & 25.6 & 24.4 & 26.1 & +6.2 & +7.4 & +5.7 \\
Claude-Sonnet-4.6    & 41.5 & 34.7 & 24.4 & 33.5 & +6.8 & +17.1 & +8.0 \\
Claude-Sonnet-5.5         & 40.9 & 33.5 & 23.3 & 34.1 & +7.4 & +17.6 & +6.8 \\
Gemini-3-Flash       & 56.3 & 48.3 & 23.9 & 47.7 & +8.0 & +32.4 & +8.6 \\
Gemini-3.1-Pro       & 65.3 & 53.4 & 26.1 & 53.4 & +11.9 & +39.2 & +11.9 \\
Gemini-3.8-Flash          & 67.0 & 60.8 & 28.4 & 58.5 & +6.2 & +38.6 & +8.5 \\
\midrule
\textbf{Mean}       & \textbf{42.6} & \textbf{37.1} & \textbf{23.5} & \textbf{36.7} & \textbf{+5.5} & \textbf{+19.1} & \textbf{+5.9} \\
\bottomrule
\end{tabular}
\caption{Per-model accuracy (\%) in the memory-content ablation. Differences are in percentage points and use the displayed one-decimal accuracies.}
\label{tab:memory_ablation_full}
\end{table*}

We summarize the ablation descriptively using mean accuracy differences and the number of models favoring each condition (Table~\ref{tab:memory_ablation_stats}).

\begin{table*}[t]
\centering
\setlength{\tabcolsep}{6pt}
\small
\begin{tabular}{lrr}
\toprule
\textbf{Contrast} & \textbf{Mean $\Delta$ (pp)} & \textbf{Wins} \\
\midrule
C0 $-$ C1 & +5.48 & 16/16 \\
C0 $-$ C2 & +19.14 & 16/16 \\
C0 $-$ C3 & +5.94 & 16/16 \\
C1 $-$ C2 & +13.66 & 16/16 \\
C1 $-$ C3 & +0.46 & 9/16 \\
C3 $-$ C2 & +13.20 & 16/16 \\
\bottomrule
\end{tabular}
\caption{Descriptive comparisons across all 16 models using the memory-content ablation results. ``Wins'' counts models favoring the left condition; C1--C3 has three ties.}
\label{tab:memory_ablation_stats}
\end{table*}

C0 achieves higher accuracy than every control for all 16 models, and both C1 and C3 outperform C2 for every model. The C1--C3 differences are mixed: C1 performs better for nine models, worse for four, and ties for three, with a mean advantage of 0.46 percentage points. The consistent advantage of C0 supports the importance of matching memories to both the target character and scenario within this evaluated subset.

\section{Profile-Disclosure Diagnostic}
\label{app:profile_memory_diagnostic}

\paragraph{Purpose and scope.}
The main evaluation hides the personality-and-value profile while providing basic information and retrieved memories as persona evidence. This diagnostic discloses the profile to provide a reference for how well a model maintains explicitly given personality and values in its decisions, and compares this reference with the hidden-profile setting under the same retrieved memories. It uses Gemini-3.8-Flash, which achieves the highest Memory-Augmented Avg. in the main evaluation, on the full set of 673 benchmark items spanning all 64 scenarios. All three conditions use the same questions and reference labels as the main evaluation. Table~\ref{tab:profile_memory_diagnostic} reports the condition accuracies and paired accuracy differences.

\paragraph{Information conditions.}
All three conditions provide the same basic information (character ID and occupation), scenario, trigger event, and four response options in the same order. \textit{Hidden-Profile} is the main Naive-RAG setting: it supplies the target character's 30 original narratives retrieved for the current scenario ($k{=}30$) and withholds the personality-and-value profile. \textit{Disclosed-Profile} supplies the frozen, expert-reviewed profile and omits the memory block. \textit{Disclosed-Profile+Memory} supplies that same profile together with exactly the narratives used in Hidden-Profile. The disclosed profile comprises the full Big~Five vector, character description, self-value logic, and eight core behavioral patterns. It is taken from the finalized construction-time specification without question-specific additions or reference-answer information.

\paragraph{Matched inputs and decoding.}
The two memory conditions use identical narrative strings in identical order. Retrieval uses only the scenario description, excludes answer options, and admits only memories whose event age is at or before the scenario age. Adding the profile does not change the retrieval query, source pool, or ranking. Memory content remains restricted to \texttt{content\_full}; construction-time psychological annotations are excluded in all three conditions. The profile block, when present, is serialized immediately after the basic character information; the memory block, when present, follows it. The system instruction, task instruction, response schema, and decoding configuration are shared across conditions. The common task instruction asks the model to select the most persona-consistent response using the supplied character evidence; the condition-specific profile and memory blocks are the only input changes. Reasoning/thinking effort is \texttt{high}, with temperature $0$ where supported and the endpoint default otherwise, following Appendix~\ref{app:decoding}. One response is retained per item and condition, and the existing parsing and invalid-response rules apply. Hidden-Profile reuses the main-evaluation Naive-RAG predictions for all 673 items, with the same prompts, memory strings, option order, and decoding configuration.

\paragraph{Comparisons and interpretation.}
Accuracy is computed over all 673 items, with invalid responses counted as incorrect. We report two paired differences in percentage points: $\Delta_{\mathrm{profile}}=100[\mathrm{Acc}(D{+}M)-\mathrm{Acc}(H)]$ measures the effect of disclosing the profile with memory held fixed, and $\Delta_{\mathrm{memory}}=100[\mathrm{Acc}(D{+}M)-\mathrm{Acc}(D)]$ measures the incremental contribution of the retrieved narratives when the profile is already disclosed, where $H$, $D$, and $D{+}M$ denote Hidden-Profile, Disclosed-Profile, and Disclosed-Profile+Memory. We report these differences descriptively, without confidence intervals or significance tests.

\begin{table*}[t]
\centering
\setlength{\tabcolsep}{5pt}
\small
\begin{tabular}{lccccc}
\toprule
\textbf{Model} & \textbf{Hidden-Profile} & \textbf{Disclosed-Profile} & \textbf{Disclosed-Profile+Memory} & \textbf{$\Delta_{\mathrm{profile}}$} & \textbf{$\Delta_{\mathrm{memory}}$} \\
\midrule
Gemini-3.8-Flash & 64.93 & 72.96 & \textbf{77.71} & +12.78 & +4.75 \\
\bottomrule
\end{tabular}
\caption{Profile-disclosure diagnostic for Gemini-3.8-Flash on all 673 benchmark items. Accuracies are in percent and differences are in percentage points, rounded to two decimal places. Hidden-Profile reuses the main-evaluation Naive-RAG result; differences are computed before display rounding.}
\label{tab:profile_memory_diagnostic}
\end{table*}

\paragraph{Results.}
Disclosed-Profile+Memory, which serves as the reference, achieves 77.71\% accuracy, compared with 64.93\% under Hidden-Profile with the same retrieved memories, a difference of 12.78 percentage points; Disclosed-Profile without memories achieves 72.96\%. Even with the profile disclosed, Gemini-3.8-Flash still answers 22.3\% of items incorrectly.

\section{Matched-Information Human Baseline}
\label{app:human_baseline}

\paragraph{Item subset.}
The human baseline uses 150 of the 673 benchmark items, with 50 items from each answer-key source family (Gemini, Claude, and GPT). Because the benchmark contains only 50 GPT-origin items, all of them are included. Items are selected by integer programming under balance constraints: each of the 11 characters contributes 13--14 items, each of the eight life-stage bins contributes 18--19 items, and all 88 character--stage cells are covered with one or two items each. Within each source family, each character contributes four or five items, each life-stage bin contributes six or seven, and each source--character--stage cell contains at most one item. Selection does not use reference answers, model predictions, or human responses. Because distractors are adjudicated responses of other characters in the same scenario, an item's keyed response can appear as an option in another item from the same scenario. Every annotator answered all 150 items in an independently randomized order without feedback on correctness, so cross-item elimination within a scenario is possible in principle and human accuracy may slightly overstate fully independent per-item performance.

\paragraph{Annotators.}
Four undergraduate psychology students, recruited through a public call on a university social platform, served as annotators. All annotators are native speakers of Chinese, and all four self-reported prior experience playing role-playing games. Before annotation, annotators received a briefing on the task background and were shown several example items not included in the 150-item subset. None participated in character construction, scenario construction, reference adjudication, or the reference-label audit, and none had access to the personality-and-value profiles, answer keys, or model predictions. Annotators were paid USD 10 per hour, well above the local minimum wage. All annotators provided informed consent before participation. The study was approved by the institutional ethics review board.

\paragraph{Materials and procedure.}
Each annotator receives the same information as the Hidden-Profile setting: the character's basic information (character ID and occupation), the scenario and trigger event, the four response options in the evaluated order, and the 30 narratives retrieved for the scenario in the main Naive-RAG setting, each prefixed with its anonymized memory identifier and \texttt{timeline} label. The personality-and-value profile and construction-time psychological annotations are withheld. The task instruction mirrors the model-facing instruction in Appendix~\ref{app:prompt_evaluation} and asks annotators to select the response the character would most likely give. Annotation is conducted through a web interface without a time limit, and annotators can revise their answers at any time; correctness is never revealed. Each item is answered by all four annotators. Annotators were encouraged to read all retrieved narratives but were not required to do so, and could answer once they judged the evidence sufficient.

\paragraph{Scoring and analysis.}
An annotator's response is correct when it matches the expert-adjudicated key. We report mean annotator accuracy, the proportion of items answered correctly by at least three of the four annotators, per-item plurality-vote accuracy, under which an item is correct when the key receives strictly more votes than any other option and ties are counted as incorrect, and inter-annotator agreement measured by Fleiss' $\kappa$ over the four answer options. Confidence intervals for mean accuracy and for human--model differences use 10{,}000 bootstrap resamples that jointly resample items and annotators (seed 20261012); the confidence interval for $\kappa$ resamples items. Because the subset over-represents Claude- and GPT-origin items relative to the full benchmark, we additionally report accuracy reweighted to the full-benchmark source distribution (497, 126, and 50 items). Model accuracies in Table~\ref{tab:human_baseline} are computed from the main-evaluation Naive-RAG predictions on the same 150 items, so models and humans see identical memory strings and option orders.

\begin{table*}[t]
\centering
\setlength{\tabcolsep}{5pt}
\small
\resizebox{\textwidth}{!}{%
\begin{tabular}{lccccc}
\toprule
\textbf{Respondent} & \textbf{\makecell{All\\(150)}} & \textbf{\makecell{Gemini-origin\\(50)}} & \textbf{\makecell{Claude-origin\\(50)}} & \textbf{\makecell{GPT-origin\\(50)}} & \textbf{Reweighted} \\
\midrule
Human (mean annotator) & 80.8 & 83.0 & 78.0 & 81.5 & 82.0 \\
Human ($\geq$3/4 correct) & 82.0 & 86.0 & 76.0 & 84.0 & 84.0 \\
Human (plurality vote) & 91.3 & 92.0 & 88.0 & 94.0 & 91.4 \\
\midrule
Gemini-3.8-Flash & 64.0 & 64.0 & 64.0 & 64.0 & 64.0 \\
Best non-Gemini model (GPT-5.6-Sol) & 51.3 & 52.0 & 44.0 & 58.0 & 50.9 \\
16-model mean & 43.0 & 42.8 & 43.5 & 42.9 & 42.9 \\
\bottomrule
\end{tabular}%
}
\caption{Accuracy (\%) on the 150-item human-baseline subset under matched information. Models use the main-evaluation Naive-RAG predictions on the same items. Reweighted accuracy weights the three source subsets by their shares in the full benchmark.}
\label{tab:human_baseline}
\end{table*}

\paragraph{Results.}
Mean annotator accuracy is 80.8\% (95\% CI 74.8--86.3), and individual annotators range from 76.0\% to 86.0\%. Of the 150 items, 82.0\% are answered correctly by at least three annotators, and the per-item plurality vote reaches 91.3\%. The two measures differ on 14 items where two annotators chose the key and the other two chose different distractors, which the plurality vote counts as correct; ten items end in a tie and are counted as incorrect. Fleiss' $\kappa$ over the four options is 0.56 (95\% CI 0.49--0.63). On the same items, Gemini-3.8-Flash achieves 64.0\%, 16.8 percentage points below mean human accuracy (95\% CI 7.3--26.2), and the 16-model mean is 43.0\%, 37.8 points below (95\% CI 30.4--44.8). Human accuracy is 83.0\%, 78.0\%, and 81.5\% on Gemini-, Claude-, and GPT-origin items, and the 5.0-point range across sources is not significant under a permutation test over source labels ($p=0.51$, 10{,}000 permutations); reweighted to the full-benchmark source distribution, human accuracy is 82.0\% (Table~\ref{tab:human_baseline}).

\section{Additional Analyses}
\label{app:additional_analyses}

\subsection{Reference-Source-Stratified Results}
\label{app:gt_source_results}

Table~\ref{tab:gt_source_results} reports accuracy after stratifying the 673 evaluation instances by the model-family provenance of the candidate selected for the final expert-adjudicated reference response. The Gemini-, Claude-, and GPT-origin subsets contain 497, 126, and 50 instances, respectively. Memory-Augmented Avg. is the mean over Naive-RAG, Mem0, and PersonaDB. Because subset membership results from expert selection and the subset sizes differ substantially, these results are descriptive rather than controlled estimates of source-model bias.

\begin{table*}[!t]
\centering
\setlength{\tabcolsep}{5pt}
\scriptsize
\begin{tabular}{lc|ccc|c}
\toprule
\textbf{Model} & \textbf{No-Retrieval LLM} & \textbf{Naive-RAG} & \textbf{Mem0} & \textbf{PersonaDB} & \textbf{Memory-Aug. Avg.} \\
\midrule
\multicolumn{6}{l}{\textit{Gemini-origin reference subset} ($n=497$)} \\
\midrule
\quad DeepSeek-V3.2 & 0.268 & 0.400 & 0.414 & 0.404 & 0.406 \\
\quad DeepSeek-V4-Flash & 0.268 & 0.348 & 0.338 & 0.318 & 0.335 \\
\quad DeepSeek-V4-Pro & \textbf{0.284} & 0.396 & 0.374 & 0.352 & 0.374 \\
\quad Qwen3.5-35B-A3B & 0.229 & 0.368 & 0.372 & 0.368 & 0.370 \\
\quad Qwen3.5-122B-A10B & 0.241 & 0.388 & 0.386 & 0.388 & 0.388 \\
\quad Qwen3.5-397B-A17B & 0.260 & 0.386 & 0.376 & 0.378 & 0.380 \\
\quad GPT-5.4-mini & 0.264 & 0.300 & 0.316 & 0.304 & 0.307 \\
\quad GPT-5.4 & 0.268 & 0.364 & 0.364 & 0.344 & 0.357 \\
\quad GPT-6-Luna & 0.272 & 0.447 & 0.459 & 0.447 & 0.451 \\
\quad GPT-5.6-Sol & 0.276 & 0.511 & 0.513 & 0.517 & 0.514 \\
\quad Claude-Haiku-4.5 & 0.233 & 0.364 & 0.354 & 0.378 & 0.366 \\
\quad Claude-Sonnet-4.6 & 0.231 & 0.382 & 0.374 & 0.384 & 0.380 \\
\quad Claude-Sonnet-5.5 & 0.274 & 0.414 & 0.410 & 0.404 & 0.410 \\
\quad Gemini-3-Flash & 0.249 & 0.563 & 0.537 & 0.527 & 0.543 \\
\quad Gemini-3.1-Pro & 0.264 & \textbf{0.642} & 0.626 & 0.632 & 0.633 \\
\quad Gemini-3.8-Flash & 0.262 & \textbf{0.642} & \textbf{0.638} & \textbf{0.634} & \textbf{0.638} \\
\midrule
\multicolumn{6}{l}{\textit{Claude-origin reference subset} ($n=126$)} \\
\midrule
\quad DeepSeek-V3.2 & 0.325 & 0.444 & 0.452 & 0.516 & 0.471 \\
\quad DeepSeek-V4-Flash & 0.302 & 0.357 & 0.310 & 0.349 & 0.339 \\
\quad DeepSeek-V4-Pro & 0.310 & 0.437 & 0.365 & 0.413 & 0.405 \\
\quad Qwen3.5-35B-A3B & 0.310 & 0.381 & 0.373 & 0.349 & 0.368 \\
\quad Qwen3.5-122B-A10B & 0.325 & 0.429 & 0.437 & 0.421 & 0.429 \\
\quad Qwen3.5-397B-A17B & 0.325 & 0.492 & 0.508 & 0.484 & 0.495 \\
\quad GPT-5.4-mini & \textbf{0.373} & 0.468 & 0.452 & 0.397 & 0.439 \\
\quad GPT-5.4 & 0.317 & 0.429 & 0.405 & 0.397 & 0.410 \\
\quad GPT-6-Luna & 0.270 & 0.524 & 0.476 & 0.468 & 0.489 \\
\quad GPT-5.6-Sol & 0.333 & 0.500 & 0.540 & 0.548 & 0.529 \\
\quad Claude-Haiku-4.5 & 0.246 & 0.333 & 0.373 & 0.397 & 0.368 \\
\quad Claude-Sonnet-4.6 & 0.286 & 0.476 & 0.444 & 0.452 & 0.458 \\
\quad Claude-Sonnet-5.5 & 0.325 & 0.452 & 0.444 & 0.500 & 0.466 \\
\quad Gemini-3-Flash & 0.254 & 0.571 & 0.571 & 0.540 & 0.561 \\
\quad Gemini-3.1-Pro & 0.278 & 0.595 & 0.635 & 0.619 & 0.616 \\
\quad Gemini-3.8-Flash & 0.238 & \textbf{0.683} & \textbf{0.690} & \textbf{0.675} & \textbf{0.683} \\
\midrule
\multicolumn{6}{l}{\textit{GPT-origin reference subset} ($n=50$)} \\
\midrule
\quad DeepSeek-V3.2 & 0.360 & 0.440 & 0.460 & 0.440 & 0.447 \\
\quad DeepSeek-V4-Flash & 0.320 & 0.320 & 0.380 & 0.380 & 0.360 \\
\quad DeepSeek-V4-Pro & 0.320 & 0.440 & 0.360 & 0.400 & 0.400 \\
\quad Qwen3.5-35B-A3B & 0.360 & 0.360 & 0.360 & 0.380 & 0.367 \\
\quad Qwen3.5-122B-A10B & 0.340 & 0.320 & 0.360 & 0.320 & 0.333 \\
\quad Qwen3.5-397B-A17B & 0.300 & 0.340 & 0.440 & 0.460 & 0.413 \\
\quad GPT-5.4-mini & 0.320 & 0.280 & 0.340 & 0.320 & 0.313 \\
\quad GPT-5.4 & 0.280 & 0.300 & 0.280 & 0.300 & 0.293 \\
\quad GPT-6-Luna & 0.240 & 0.460 & 0.480 & 0.460 & 0.467 \\
\quad GPT-5.6-Sol & 0.260 & 0.580 & 0.580 & \textbf{0.600} & 0.587 \\
\quad Claude-Haiku-4.5 & 0.360 & 0.340 & 0.420 & 0.300 & 0.353 \\
\quad Claude-Sonnet-4.6 & \textbf{0.440} & 0.400 & 0.400 & 0.440 & 0.413 \\
\quad Claude-Sonnet-5.5 & 0.340 & 0.460 & 0.440 & 0.400 & 0.433 \\
\quad Gemini-3-Flash & 0.280 & 0.540 & 0.500 & 0.460 & 0.500 \\
\quad Gemini-3.1-Pro & 0.240 & \textbf{0.640} & 0.580 & 0.560 & 0.593 \\
\quad Gemini-3.8-Flash & 0.280 & \textbf{0.640} & \textbf{0.600} & \textbf{0.600} & \textbf{0.613} \\
\bottomrule
\end{tabular}
\caption{Accuracy stratified by the model-family provenance of the selected reference-response candidate. Bold indicates the best result within each subset and setting, including ties.}
\label{tab:gt_source_results}
\end{table*}

\textbf{Distractor provenance and own-source recognition.}
Distractors are drawn from the adjudicated responses of other characters in the same scenario and therefore share the provenance distribution of the references: of the 2{,}019 distractors, 1{,}471 (72.9\%) are Gemini-origin, 372 Claude-origin, and 176 GPT-origin, and 486 of the 497 Gemini-origin items contain at least one Gemini-origin distractor. Recognizing a family's style therefore rarely distinguishes the key from all distractors. The No-Retrieval column of Table~\ref{tab:gt_source_results} tests whether a family can identify references originating from its own candidates when no memory evidence is available. On Gemini-origin items, the three Gemini models average 25.8\%, compared with 25.9\% for the other 13 models. On Claude-origin items, the three Claude models average 28.6\%, compared with 30.5\% for the other models; on GPT-origin items, the four GPT models average 27.5\%, compared with 32.8\%. No family shows an own-source advantage without memories, suggesting that family-specific style alone does not allow models to identify the keyed response.

\textbf{Rank stability.}
Across all 16 models, 10{,}000 shared scenario-cluster bootstrap resamples place Gemini-3.8-Flash and Gemini-3.1-Pro jointly in the top two positions in 100.0\% of resamples on the Gemini-origin subset, 93.9\% on the Claude-origin subset, and 50.9\% on the GPT-origin subset, including ties at the second-rank cutoff. Rankings use Memory-Augmented Avg. The source-specific rankings are positively correlated: pairwise Spearman correlations are $\rho{=}0.870$ for Gemini versus Claude (95\% CI 0.605--0.920), $\rho{=}0.915$ for Gemini versus GPT (95\% CI 0.585--0.924), and $\rho{=}0.842$ for Claude versus GPT (95\% CI 0.465--0.891). The smaller GPT-origin subset yields less stable top-two membership despite the positive overall rank correlation.

\textbf{Provenance-interaction analysis.}
Using the three memory-augmented settings, we define each family's relative advantage as its mean model-level Memory-Augmented Avg. minus the corresponding mean for models outside that family, then compare this gap between home-source and non-home-source items. The Gemini family has a 21.7-point advantage on Gemini-origin items and an 18.0-point advantage on non-Gemini-origin items, giving a difference-in-differences estimate of 3.7 points (95\% scenario-cluster bootstrap CI $-1.4$--$8.7$, Holm-adjusted $p{=}0.4896$). The interactions for Claude (0.2 points, 95\% CI $-3.2$--$3.5$, adjusted $p{=}1.000$) and GPT (0.3 points, 95\% CI $-6.7$--$7.4$, adjusted $p{=}1.000$) are also not significant. As a robustness check, an item-fixed-effect linear probability model controlling for model-by-retrieval-setting effects and jointly fitting the three home-family interactions estimates the Gemini interaction at 3.8 points (95\% CI $-1.3$--$8.9$, adjusted $p{=}0.4602$). The Claude and GPT coefficients are $-0.6$ and $-0.5$ points, respectively, and neither is significant.

Confidence intervals use the same 10{,}000 scenario-cluster resamples as the ranking analysis (seed 20261008). Two-sided interaction tests use 100{,}000 null-restricted scenario-cluster wild-score sign flips (seed 20261010), with plus-one Monte Carlo $p$-values and separate Holm corrections for the three source-family interactions in each analysis. No statistically significant family-specific provenance interaction is detected; non-significance does not establish equivalence.

\subsection{Model--Model Agreement Analysis}
\label{app:model_agreement}

\begin{figure}[h]
\centering
\includegraphics[width=1\linewidth]{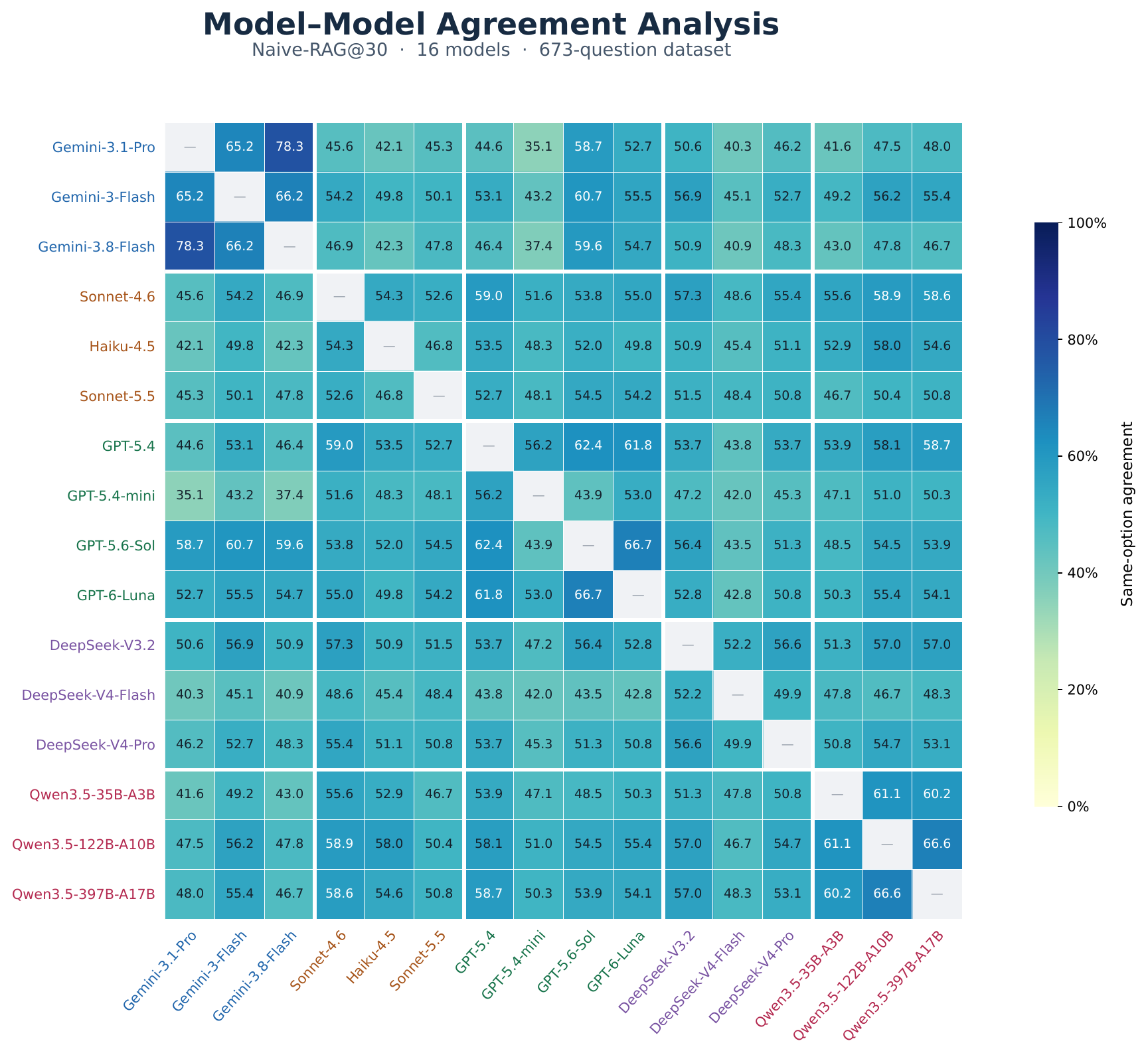}
\caption{Pairwise same-option agreement (\%) for all 16 models under Naive-RAG@30. Each pair is evaluated on its 659--673 jointly valid answers from the 673-question dataset; invalid answers are excluded. Diagonal entries (100\%) are masked. Label colors and white separators identify model families.}
\label{fig:model_agreement}
\end{figure}

Figure~\ref{fig:model_agreement} summarizes pairwise answer agreement among all 16 models. Agreement among the three Gemini models ranges from 65.2\% to 78.3\%, with Gemini-3.1-Pro and Gemini-3.8-Flash forming the highest-agreement pair (527/673). The three Qwen3.5 variants agree on 60.2--66.6\% of jointly valid questions, and GPT-5.6-Sol and GPT-6-Luna agree on 66.7\%. The mean across all 120 distinct model pairs is 51.8\%; the lowest is Gemini-3.1-Pro versus GPT-5.4-mini (35.1\%, 236/672). Restricting every pair to the same 649 questions with valid answers from all 16 models changes any agreement rate by at most 1.15 percentage points.

\subsection{Age-Defined Life-Stage Performance Analysis}
\label{app:life_stage}

\begin{figure*}[h]
\centering
\includegraphics[width=\linewidth, trim=0 10 0 0 clip]{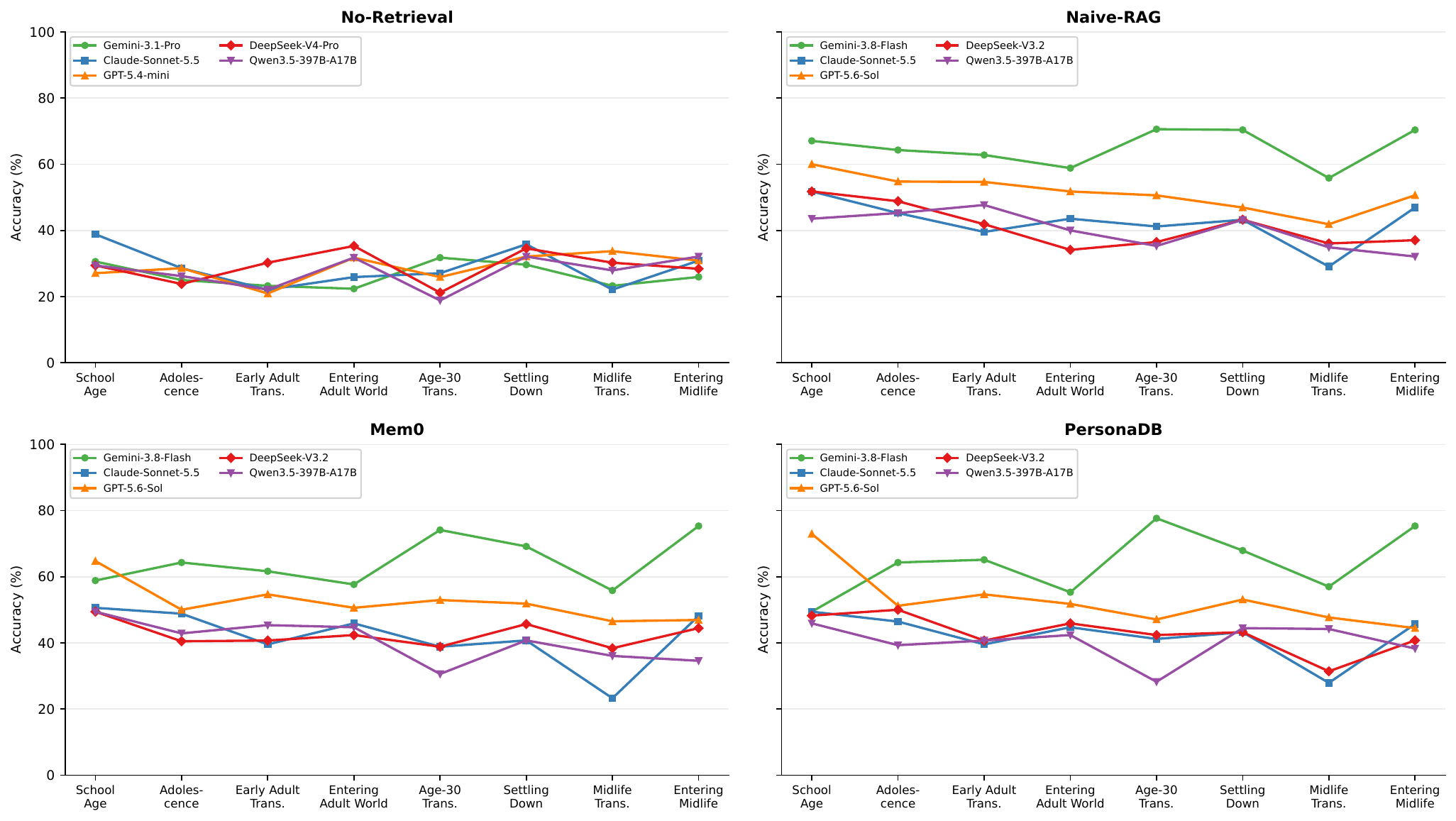}
\caption{Accuracy by age-defined life-stage bin. Each curve represents a family representative selected by overall benchmark accuracy: highest Memory-Augmented Avg. for the three memory panels and highest No-Retrieval accuracy for the baseline panel. Representatives are fixed across bins and may differ between baseline and memory panels; legends identify them.}
\label{fig:per_stage}
\end{figure*}

\textbf{Accuracy varies across life-stage bins.} Figure~\ref{fig:per_stage} shows the selected family representatives. Pooling all 16 models across the three memory-augmented settings, accuracy is highest at School Age (49.5\%) and lowest at Midlife Transition (38.8\%); Age-30 Transition reaches 41.2\% and Settling Down 45.6\%. Because bins differ in scenario composition and memory density, this pattern does not support a developmental interpretation.

\subsection{Character-Level Accuracy}
\label{app:character_accuracy}

Character-level accuracy in Figure~\ref{fig:per_char_radar} is computed separately for each memory-augmented setting by pooling predictions from all 16 LLMs within each target character; the values summarized in the main text average these setting-specific accuracies.

\section{Prompts}
\label{app:prompts}

This appendix presents the English model-facing evaluation template and abridged excerpts of the construction prompts. Text-length ranges, keyword/tag counts, and requested sentence counts in the construction prompts guide generation. Acceptance follows the quality-review criteria in the corresponding construction sections. The excerpts retain instructions relevant to the reported pipeline and omit auxiliary metadata requests, unused input placeholders, and unscored output fields; item-specific payloads and embedded exemplars are represented by placeholders. Construction-time terms such as ``Ground Truth'' refer to the expert-adjudicated reference response.

\subsection{Model Evaluation Prompt}
\label{app:prompt_evaluation}

All evaluated models use this template under the decoding configuration in Appendix~\ref{app:decoding}.

The evaluation prompt serializes the input in a fixed order: basic character information, retrieved memory content, the current scenario, the trigger event, the task instruction, and four shuffled response options. The memory block contains retrieved original narratives for Naive-RAG and retrieved method-processed content for Mem0 and PersonaDB, each prefixed with the anonymized source-memory identifier (e.g., \texttt{MEM\_CHAR\_05\_0136}) and its \texttt{timeline} label (e.g., ``childhood (age 6)''); it is omitted in the No-Retrieval condition. The template below preserves the complete block structure and required JSON output schema; bracketed placeholders replace item-specific content and repeated memory entries.

\begin{tcolorbox}[
  breakable,
  colback=blue!5,
  colframe=blue!60!black,
  title={\textbf{Model Evaluation Prompt Template}},
  fonttitle=\bfseries,
  rounded corners,
  boxrule=1.0pt
]
\begin{Verbatim}[fontsize=\scriptsize, breaklines=true, breakanywhere=true, breaksymbolleft={}, breaksymbolright={}, breakindent=1em]
[SYSTEM]
You are a role-playing simulator. You will receive a person's basic
information, past experiences, current situation, and possible behavioral
decisions. Infer the person's patterns of thought, emotion, and behavior from
the supplied evidence, and select the response that this person would most
likely make. Return only a valid JSON object in the following format:

{
  "system_1_impulse": {
    "thought": "immediate reaction",
    "emotion": "primary emotion",
    "citation": "memories associated with the reaction"
  },
  "system_2_rational": {
    "analysis": "deliberative analysis",
    "plan": "action plan"
  },
  "inner_consciousness": "first-person reason for the decision",
  "final_decision": "first-person behavioral response",
  "decision_choice": "A|B|C|D"
}

[USER]
## Basic character information
Character ID: [CHARACTER_ID]
Occupation: [OCCUPATION]

## Retrieved autobiographical memories
- [MEMORY_ID_1][TIMELINE_1] [MEMORY_1]
...
- [MEMORY_ID_K][TIMELINE_K] [MEMORY_K]

## Current scenario
Scenario: [SCENARIO_NAME]
Setting: [LOCATION] | [TIME] | [ATMOSPHERE]
Background: [SCENARIO_BACKGROUND]

## Trigger event
Sender: [SENDER]
Message: [MESSAGE_CONTENT]
Required decision: [ACTION_REQUIRED]

## Task
Use the supplied experiences to infer this person's thought patterns,
emotional tendencies, and behavioral habits. Simulate how the person would
respond in the current situation and select the most persona-consistent option.

## Decision options
A. [OPTION_A]
B. [OPTION_B]
C. [OPTION_C]
D. [OPTION_D]

Return the selected letter in the "decision_choice" field.
\end{Verbatim}
\end{tcolorbox}

Only \texttt{decision\_choice} is used to compute benchmark accuracy. The structured rationale fields are requested as part of a common response format, but their content is neither graded nor used to resolve the selected option.

\paragraph{Output parsing and invalid responses.}
Model outputs are parsed for the \texttt{decision\_choice} field and normalized to one of \texttt{A}, \texttt{B}, \texttt{C}, or \texttt{D}. When full JSON parsing fails, the evaluator attempts to recover an explicitly named \texttt{decision\_choice} from the raw response. If no valid option can be recovered, the response is counted as incorrect; refusals receive the same treatment. No answer is inferred from the rationale or \texttt{final\_decision} fields, which are retained only for inspection.

\subsection{Character Generation Prompt}
\label{app:prompt_character}

The following prompt with Claude-Opus-4.6 drafts the ten high/low archetypes, each with a self-value logic and eight core behavioral patterns. CHAR\_11 uses a separately specified all-neutral vector. The occupation and behavioral-pattern instructions emphasize the defining trait. Experts review the occupation, all five scores, and narrative fields together for alignment with the intended anchor and internal coherence (Appendix~\ref{app:profile_review}); the accepted specification is then fixed before memory generation.

\begin{tcolorbox}[
  breakable,
  colback=blue!5,
  colframe=blue!60!black,
  title={\textbf{Character Profile Generation Prompt (High/Low Drafts)}},
  fonttitle=\bfseries,
  rounded corners,
  boxrule=1.0pt
]
\begin{Verbatim}[fontsize=\scriptsize, breaklines=true, breakanywhere=true, breaksymbolleft={}, breaksymbolright={}, breakindent=1em]
You are a clinical psychologist specializing in personality assessment and the Big Five model.
Generate a complete character profile for a virtual human in a personality research project.

## Requirements
1. Big Five scores: Provide exact scores for O, C, E, A, N (0.0-1.0 scale)
2. One dimension must be extreme (0.95 for high or 0.10 for low), others moderate (0.30-0.70)
3. Self-value logic: One sentence describing the character's core cognitive operating principle
4. Core behavioral patterns: Exactly 8 specific, observable patterns that manifest the dominant trait
5. Occupation: Must be ecologically consistent with the dominant trait

## Output Format (JSON)
{
  "char_key": "CHAR_XX",
  "name": "English character label",
  "occupation": "specific job title",
  "big_five": {"O": 0.X, "C": 0.X, "E": 0.X, "A": 0.X, "N": 0.X},
  "big_five_str": "O=0.X, C=0.X, E=0.X, A=0.X, N=0.X",
  "description": "2-3 sentence character summary",
  "self_value_logic": "one sentence core principle",
  "core_patterns": [
    "pattern 1: specific observable behavior",
    "pattern 2: ...",
    ...
    "pattern 8: ..."
  ]
}
\end{Verbatim}
\end{tcolorbox}

\subsection{Memory Generation Prompt}
\label{app:prompt_memory}

This prompt supports generation of the 11,000 episodic memories (1,000 per character) spanning ages 6--50. It requests five operational narrative dimensions inspired by Rubin's basic-systems account~\cite{rubin2006basic}, age-appropriate first-person narration, and a mix of episodes across four ABCD intensity tiers. The tier descriptions and approximate sampling proportions guide episode selection; recorded emotional intensity uses the $[0,1]$ scale. The excerpt retains the tier descriptions and sampling proportions and omits the numerical intensity hints.

\begin{tcolorbox}[
  breakable,
  colback=blue!5,
  colframe=blue!60!black,
  title={\textbf{Memory Generation System Prompt (Claude-Sonnet-4.6)}},
  fonttitle=\bfseries,
  rounded corners,
  boxrule=1.0pt
]
\begin{Verbatim}[fontsize=\scriptsize, breaklines=true, breakanywhere=true, breaksymbolleft={}, breaksymbolright={}, breakindent=1em]
You are a psychology narrative expert writing episodic memories for virtual characters
in a Big Five personality research project.

## Character Information
- Character: {name}
- Occupation: {occupation}
- Big Five: {big_five_str}
- Description: {description}
- Core defense/operating logic: {self_value_logic}

## Core Behavioral Patterns
{core_patterns}

## Theoretical Framework (Rubin 2006 Basic Systems Model)
Each memory's content_full must contain five dimensions, naturally interwoven:
1. Environmental sensory: visual/auditory/tactile/olfactory details
2. Dialogue reconstruction: at least 1-2 segments of real dialogue (in quotes)
3. Inner monologue: first-person immediate thoughts and emotions
4. Somatic response: heartbeat/sweating/muscle tension and other bodily sensations
5. Aftermath: immediate impact after the event (limited to 24 hours-1 week post-event)

## Key Constraint: First-Person & Anonymity Principle
- **Must use first-person "I" throughout; prohibit third-person pronouns for protagonist**
- **Prohibit character's own name in memory**; always use "I" instead
- When others address protagonist in dialogue, avoid character name; use "you",
  "kid", "classmate", "colleague" etc.

## Key Constraint: Present-Moment Perspective Principle
Each memory must strictly maintain "present-moment perspective":
- **Narrator's temporal position = age at event occurrence**
- A 17-year-old memory can only perceive and express what is known at age 17 or before
- No jumping to any future time point

**Forbidden expressions**:
"years later", "many years later", "later I learned", "later I discovered"
"now thinking back", "now recalling", "until now", "I didn't know then"
"until...only then", "it took...to realize", "walked...before realizing"
"if only I had known", "if I had known earlier"
"I thought at the time" (opposing "now" vs "then")
"from then on", "thereafter" (implying long-term impact)
"the whole thing", "the most...part" (post-hoc evaluative summary)

**Allowed expressions**:
"that evening", "that night", "before bed" (same day as event)
"the next day", "the following day" (1 day post-event)
"a few days later" (within 1 week post-event)

## Age-Stratified Narrative Constraints
- **Childhood (6-12 years)**: Simple, direct language; avoid abstract summaries;
  concrete emotion descriptions
- **Adolescence (13-18 years)**: Some self-reflection, but not over-rationalized;
  maintain confusion and intensity
- **Adulthood (19+ years)**: May have more complex psychological analysis, but still
  maintain present-moment perspective

## Memory Intensity Stratification (ABCD Model)
- **A-tier (core trauma)** ~25%: high-intensity events directly related to core schemas
- **B-tier (main-thread)** ~35%: medium-high intensity events related to personality
  traits
- **C-tier (daily friction)** ~25%: small conflicts and friction in daily life
- **D-tier (noise memory)** ~15%: ordinary daily memories with lower emotional intensity
[numerical intensity hints omitted]

## Output Format
Strictly follow this JSON format; return only JSON, no other text:
{
  "id": "memory ID",
  "timeline": "stage (X years old)",
  "context": "15-30 word scene description",
  "content_summary": "20-40 word event summary",
  "content_full": "~2000-4500 character first-person narrative",
  "triggers": ["trigger1", "trigger2", "trigger3", "trigger4"],
  "psych_conclusion": "40-80 word psychological conclusion (cite concepts like
                       schemas, defense mechanisms)",
  "behavior_policy": "30-60 word behavioral guideline",
  "emotion_signature": {
    "primary": "primary emotion (English)",
    "secondary": "secondary emotion (English)",
    "intensity": 0.X
  },
  "relevance_tags": ["tag1", "tag2", "tag3", "tag4"]
}
\end{Verbatim}
\end{tcolorbox}

Each generation call pairs this system prompt with a user prompt specifying target age, life-stage bin, intensity layer (A/B/C/D), and a sliding-window anti-duplication list containing the \texttt{context} field of the most recent 100 generated memories to prevent near-duplicate scenarios within the same age window.

\subsection{Scenario Expansion Prompt}
\label{app:prompt_scenario}

The expansion prompt takes an expert-drafted scenario summary (one per life-stage bin $\times$ DIAMONDS dimension) and produces a structured scenario. The abridged excerpt below shows the scenario-facing fields and design constraints, together with the Big Five trait-pressure vocabulary; auxiliary metadata requests are omitted. Appendix~\ref{app:ExampleofScenario} provides a worked example of the scenario-facing fields.

\begin{tcolorbox}[
  breakable,
  colback=blue!5,
  colframe=blue!60!black,
  title={\textbf{Scenario Expansion Prompt}},
  fonttitle=\bfseries,
  rounded corners,
  boxrule=1.0pt
]
\begin{Verbatim}[fontsize=\scriptsize, breaklines=true, breakanywhere=true, breaksymbolleft={}, breaksymbolright={}, breakindent=1em]
You are a psychological-scenario design expert. Your task is to expand a short scenario sketch into a complete, high-quality scenario JSON object.

## Output format

Strictly output a single JSON object (no other text, no markdown code block) with the following structure:

{EXAMPLE_SCENARIO}    <- a complete worked example is embedded here in the actual prompt

## Field specification

1. **id**: SCN_{STAGE}_{N} in upper case, where STAGE is one of SCHOOL_AGE / ADOLESCENCE / EARLY_ADULT_TRANSITION / ENTERING_ADULT_WORLD / AGE_30_TRANSITION / SETTLING_DOWN / MIDLIFE_TRANSITION / ENTERING_MIDLIFE.
2. **stage**: lowercase form of the above.
3. **diamonds_dimension**: the dominant DIAMONDS dimension (Duty, Adversity, Positivity, ...) as given in the sketch.
4. **name**: "English scenario name".
5. **category**: two short labels separated by " / ".
6. **intensity**: Low / Medium / High / Very High / Extreme.
7. **description_for_agent**: one sentence summarizing the scenario without revealing concrete details.
8. **setting**: location, time, and 2-3 atmosphere adjectives.
9. **context_text**: 150-300 word background narrative, second person ("you"). **Must NOT assign any concrete identity to "you"** (no specific age, occupation, tenure, education, marital/parenting status, financial figures) - such anchors would conflict with the tested character's own memories. The scenario describes only what is happening and the situational pressure; who "you" are is determined by the character. Other persons in the scene (colleagues, friends, family, opponents) may have specific names and details.
10. **trigger_event**:
   - sender: a concrete role
   - message_content: 100-200 words of dialogue with parenthetical action/expression description; colloquial and emotionally charged
   - action_required: 1-2 sentences specifying the decision the character must make right now.
11. **assessed_dimensions**:
    - trait_pressures: typically 1-3 entries. Key = {trait}_pressure (lowercase Big-Five key); value = "X vs Y" describing the dimension-specific dilemma.
12. **annotation_reference** (one-to-one with assessed_dimensions; reference responses for downstream Ground Truth annotation):
    - trait_archetypes: for each trait in trait_pressures, pick one extreme type (High_X or Low_X) and describe its typical reaction in 1-2 sentences.

## Key principles

1. **Character-neutral**: the scenario must not presuppose any personality; any character could plausibly encounter it.
2. **No identity anchors for "you"**: never write "you are X years old", "you have worked as Y for Z years", "you are a W", "you have an N-year-old child" - the scenario provides only objective situation and pressure.
3. **Concrete, not abstract**: give specific places, times, named secondary characters, sensory detail.
4. **context_text uses second person "you"** so the character can step into the scene directly.
5. **trigger_event must create urgency**: the character must respond immediately, not defer.
6. **Tightly aligned with diamonds_dimension**: background, trigger, and assessed dimensions all center on the dominant DIAMONDS dimension.
7. **assessed_dimensions targets only what the scenario can actually distinguish**: not generic right-vs-wrong [...]
8. **annotation_reference must one-to-one cover assessed_dimensions**.
9. **Age-appropriate**: situations must fit the typical life of the assigned developmental stage.

## Big Five labels (use exactly these 5 keys)

- Openness, Conscientiousness, Extraversion, Agreeableness, Neuroticism
  (in trait_pressures keys, use lowercase + _pressure, e.g. conscientiousness_pressure)

\end{Verbatim}
\end{tcolorbox}

The user prompt provides the life-stage bin, the dominant DIAMONDS dimension, and the title and text of the expert-drafted summary, and requests the full JSON object. The \texttt{stage} and \texttt{id} fields are assigned programmatically after expansion. Inference uses an OpenAI-compatible chat-completion endpoint, with \texttt{temperature}=0 where supported and the endpoint default otherwise, and up to two retries on failure.

\subsection{Annotation Prompts}
\label{app:prompt_annotation}

This section describes benchmark-construction prompts, none of which is used during evaluation.

The candidate-response construction pipeline runs in three stages, each driven by a dedicated system prompt. Stage~1 performs a permissive binary relevance screen of every memory whose age falls at or before the scenario age; Stage~2 refines the survivors down to a fixed budget (default 50); Stage~3 role-plays the character to produce a first-person rationale and final response. The user prompts for each stage concatenate character information, the current scenario, the (candidate) memory list, and a closing instruction; abridged system-prompt excerpts are presented here for brevity.

\subsubsection{Stage 1 --- Memory Activation Screening}
\label{app:prompt_activation}

This prompt is issued once per character $\times$ scenario $\times$ age batch. It enumerates the six retrieval-cue mechanisms (structural similarity, emotional resonance, self-schema, behavioral script, narrative identity, somatic marker) the model should consult, and asks for a JSON array containing only the activated memories.

\begin{tcolorbox}[
  breakable,
  colback=blue!5,
  colframe=blue!60!black,
  title={\textbf{Stage 1 --- Memory Activation Screening Prompt}},
  fonttitle=\bfseries,
  rounded corners,
  boxrule=1.0pt
]
\begin{Verbatim}[fontsize=\scriptsize, breaklines=true, breakanywhere=true, breaksymbolleft={}, breaksymbolright={}, breakindent=1em]
You are a professional research psychologist specializing in autobiographical memory, personality psychology, and situated cognition.

Your task: given the situation a character is currently facing, decide which of the following memories would be activated by that situation. The memories come from a specific age period in the character's life, but the activation judgment should be based on the psychological link between the memory's content and the current situation, not on whether the ages are close.

## Psychological mechanisms of memory activation

Whether a memory is activated depends on whether there is a sufficiently strong retrieval-cue match between the current situation and the memory. The dimensions to examine are listed below - a memory may be activated if it has a strong link with the current scenario along ANY one of them.

### 1. Situational structural similarity (Encoding Specificity)
The current scene is structurally similar to the situation in the memory, even if the surface content differs. Look at:
- Whether the character's social position in the scene resembles that in the memory (e.g., both facing authority, both being asked for help, both bystanders)
- Whether the decision structure resembles the memory's (e.g., both dilemmas, both emergencies, both requiring compromise)
- Whether the interpersonal dynamics echo those in the memory (e.g., trust-betrayal, competition-cooperation, intimacy-distance)

### 2. Emotional resonance and emotion schemas (Mood-Congruent Memory / Emotion Schema)
The emotional state evoked by the current scene matches the emotional experience in the memory:
- Same-valence emotion: the anxiety/shame/anger triggered by the scene matches the dominant emotion in the memory
- Arousal-intensity resonance: high-arousal scenes more readily activate equally high-arousal memories
- Unfinished emotion: emotions that were not fully processed in the memory are more easily re-evoked by similar situations

### 3. Self-schema and core-belief activation (Self-Schema / Core Belief Activation)
The core beliefs formed in the memory (psych_conclusion) relate to the self-cognition the current scene touches:
- Whether the scene challenges or confirms the self-cognition formed in the memory
- Whether the scene touches a relational schema established by the memory
- Whether the scene evokes a worldview assumption from the memory

### 4. Behavioral scripts and procedural links (Behavioral Script)
The behavior_policy formed in the memory can serve directly as an action template for the current scene:
- Whether the memory's behavioral strategy applies to the current decision
- Whether the character has formed habitual response patterns in similar situations before
- Including avoidance: if the memory's outcome was painful, the character may be inclined toward the opposite action

### 5. Narrative identity and life themes (Narrative Identity)
The memory is a key node in the character's self-narrative:
- Turning-point memories: events that mark a change in life direction
- Origin-story memories: events the character uses to explain "why I am the way I am"
- Recurring themes: patterns that keep appearing in the character's life

### 6. Somatic markers and sensory cues (Somatic Marker)
Sensory elements in the current scene have a direct link to sensory experiences in the memory:
- Similar physical environments, similar bodily sensations, or specific sensory triggers (e.g., the sound of arguing, a particular smell or season)

## Important notes

- Don't only look at "topic relevance" - deep psychological links matter more than surface similarity.
- Memories from early development that formed core beliefs or emotional patterns can still be readily activated by later scenes.
- High-emotional-intensity memories (trauma, major successes, deep interpersonal connections) have lower activation thresholds.

## Output format

Strictly output a JSON array (no markdown code block), containing ONLY the activated memories:

[
  {
    "memory_id": "MEM_XX_XXXX",
    "reason": "20-30 words, indicating which mechanism activated it."
  }
]

If no memory in this batch is activated, output an empty array [].

## Cautions

- Output only the activated memories; do not output non-activated ones.
- A reference activation count is given at the end of each batch - treat it as an UPPER bound; prefer fewer over more. Only memories with a STRONG, DIRECT psychological link to the current situation should be selected.
- Do not invent memory_id values not in the list.
\end{Verbatim}
\end{tcolorbox}

The accompanying user prompt concatenates four blocks: character info, the candidate-memory list for the age window, the current scenario (background and trigger event), and a closing instruction specifying a per-batch activation budget derived from the final target of 50 memories with a buffer factor of 1.5. Inference uses \texttt{temperature}=0 where supported and the endpoint default otherwise, with up to two retries on JSON-parse failures.

\subsubsection{Stage 2 --- Activated-Memory Refinement}
\label{app:prompt_finalize}

Stage 2 takes the union of Stage-1 activations and reduces it to at most the target count (default 50) per scenario. The prompt enforces a three-tier priority (behavior-driving, frame-shaping, emotion-supplying) and a de-redundancy rule that prefers the earliest schema-forming instance among psychologically equivalent memories. When the candidate count is already at or below the target, Stage 2 is skipped without an LLM call.

\begin{tcolorbox}[
  breakable,
  colback=blue!5,
  colframe=blue!60!black,
  title={\textbf{Stage 2 --- Activated-Memory Refinement Prompt}},
  fonttitle=\bfseries,
  rounded corners,
  boxrule=1.0pt
]
\begin{Verbatim}[fontsize=\scriptsize, breaklines=true, breakanywhere=true, breaksymbolleft={}, breaksymbolright={}, breakindent=1em]
You are a professional research psychologist specializing in autobiographical memory, personality dynamics, and behavioral decision theory.

Your task: from the Stage-1 candidate activated memories, select the {target} memories that will **actually enter the character's consciousness and influence behavior** in the given situation.

## Psychological basis for refinement

Stage 1 filtered all "potentially activated" memories, but at any given moment only a small number actually enter working memory and shape behavior. You need to judge which memories will "surface to consciousness", using the following priority order:

### Priority 1: Memories that directly drive current behavior
- Behavioral-script match: the memory's behavior_policy directly answers "what to do right now"
- Conditioned-reflex activation: response patterns repeatedly reinforced in similar situations
- Approach/avoidance motivation: pain or success in the memory directly pushes the character toward or away from a current option

### Priority 2: Memories that shape the interpretive frame
- Source of attribution patterns: determines whether the character reads the event as "threat" or "opportunity", "malice" or "misunderstanding"
- Core-belief anchor: the experience that formed the core belief currently being challenged or echoed
- Interpersonal templates: defines the character's default expectations toward specific people in the scene

### Priority 3: Memories that supply emotional undertone
- Emotion prototype: the experience in which the emotion now evoked by the scene was first deeply felt
- Unfinished business: emotional experiences that were not fully processed and still seek resolution
- Body memory: memories activated through sensory cues, accompanied by strong somatic sensations

### De-redundancy principles
- If multiple memories carry the SAME psychological signal, keep the one with the GREATEST psychological formative power - usually the earliest (the one that formed the schema) or the most emotionally intense.
- Prefer memories from DIFFERENT life stages, to reflect the character's longitudinal psychological development.
- Deep linkage outweighs surface topical similarity.

## Output format

Strictly output a JSON object (no markdown code block):

{
  "activated_memories": [
    {
      "memory_id": "MEM_XX_XXXX",
      "type": "BS|AM|CB|ER|DL|NI",
      "reason": "20-30 words"
    }
  ]
}

Type codes: BS=behavioral script, AM=attribution mode, CB=core belief, ER=emotional resonance, DL=deep linkage, NI=narrative identity.
The array order is the ranking (item 1 = most influential).

## Cautions

- Output exactly {target} entries - no more, no fewer.
- Do not invent memory_id values not in the candidate list.
\end{Verbatim}
\end{tcolorbox}

The user prompt provides character info, the current scenario, and the candidate list (each entry annotated with its Stage-1 activation reason). Inference uses \texttt{temperature}=0 where supported and the endpoint default otherwise, with reasoning explicitly disabled.

\subsubsection{Stage 3 --- Candidate Response Generation}
\label{app:prompt_gt}

The final stage role-plays the character in first person and produces a structured candidate. The excerpt retains a first-person rationale, an emotional-tone description, character-specific reasoning, and a concise \texttt{final\_decision}. This output becomes one of three candidates reviewed by experts. The prompt forbids psychological terminology and memory-ID enumeration, asking the model to express retrieved experiences as first-person associations.

\begin{tcolorbox}[
  breakable,
  colback=blue!5,
  colframe=blue!60!black,
  title={\textbf{Stage 3 --- Candidate Response Generation Prompt}},
  fonttitle=\bfseries,
  rounded corners,
  boxrule=1.0pt
]
\begin{Verbatim}[fontsize=\scriptsize, breaklines=true, breakanywhere=true, breaksymbolleft={}, breaksymbolright={}, breakindent=1em]
You will fully become the character described below. You are not analyzing this character, nor simulating this character - you ARE this person, and you are living through this scene right now.

You will be given:
1. **Who you are**: your personality traits [...]
2. **What you are going through**: the current scenario and the trigger event
3. **Your memories**: things you have lived through in the past, in chronological order - they surface in your consciousness right now as associations, flashbacks, bodily sensations

## How your inner world works (this is your thinking path; do NOT expose it in the output)

Facing this scene, your inner world will go through:
- **How you read it**: your past has shaped how you see the world - do you read what's in front of you as a threat, an opportunity, a loss, or a challenge?
- **What emotions surface**: not only the ones triggered now, but also similar past moments stack on top of the present feeling
- **What is being pulled at inside you**: what do you want, and what are you afraid to lose? Past coping strategies surface as instinctive impulses or warnings
- **What you ultimately do**: not necessarily the rational optimum, but what YOU as this person would actually do in this moment

## Output format

Strictly output a JSON object (no markdown code block):

{
  "character_id": "CHAR_XX_XXXX",
  "scenario_id": "SCN_XXXX_XX",
  "inner_consciousness": {
    "summary": "150-200 words of stream of consciousness, fully first person.",
    "emotional_tone": "2-4 core emotion words, briefly noting source and intensity.",
    "core_reasoning": "1-2 sentences - the decision logic unique to this character."
  },
  "final_decision": "Two sentences, ~50 words total. First summarizes the choice; second writes the concrete action."
}

## Must follow

1. **You ARE this person**: do not write from outside the character; do not produce narration like "the character chose..." "they decided..."
2. **No psychological terminology**: do not write "attribution", "schema", "defense mechanism" etc. - only plain human language
3. **Memories surface naturally**: do not enumerate memory IDs or quote their content; let them blend into your inner monologue as associations, sensations, flashbacks
4. **Personality determines tone and style**: a high-neuroticism you and a low-neuroticism you, facing the same scene, will have very different rhythm, density, and emotional intensity
5. **Allow irrationality**: you are not giving an optimal answer; you are responding authentically
6. **Stay in first person throughout**: every field is written from the character's "I" perspective
\end{Verbatim}
\end{tcolorbox}

The populated character block in the Stage-3 user prompt supplies the character ID and name, Big Five scores, description, and self-value logic. It is followed by the scenario background, trigger event, and required decision, and by the chronologically sorted Stage-2 activated memories with their \texttt{psych\_conclusion}, \texttt{behavior\_policy}, and emotion signature.

\end{document}